\documentclass[twoside,11pt]{article}

\usepackage{jmlr2e}
\usepackage[utf8]{inputenc} 
\usepackage[T1]{fontenc}    
\usepackage{url}            
\usepackage{booktabs}       
\usepackage{amsfonts}       
\usepackage{nicefrac}       
\usepackage{microtype}      
\usepackage[page,header]{appendix}
\usepackage{titletoc}
\usepackage{float}
\usepackage[english]{babel}
\usepackage{mathtools}
\usepackage{verbatim}
\usepackage{hyperref}
\usepackage{bm}
\usepackage{tabu}
\usepackage{enumitem}
\usepackage[dvipsnames]{xcolor}
\usepackage{mathrsfs}
\usepackage{enumitem}
\usepackage{amsfonts}
\usepackage{graphicx}
\usepackage{parskip}
\usepackage{color}
\usepackage{tikz}
\usepackage[linesnumbered,ruled]{algorithm2e}
\usepackage{amssymb}
\usepackage{bbm}

\newcommand*\at[2]{\left.#1\right|_{#2}}

\newcommand*\Z[0]{\mathbb{Z}}

\newcommand*\ddt[0]{\frac{d}{d t}}

\newcommand*\tr[0]{\text{tr}}

\newcommand*\lin[1]{\bm{\left\langle} #1 \bm{\right\rangle}}

\newcommand*\E[1]{\mathbb{E}\left[{#1}\right]}

\newcommand*\F[0]{\mathcal{F}}

\newcommand*\N[0]{\mathcal{N}}

\newcommand\numberthis{\addtocounter{equation}{1}\tag{\theequation}}

\newcommand*\lrbb[1]{\left\{#1\right\}}
\newcommand*\lrp[1]{{\left(#1\right)}}
\newcommand*\lrn[1]{{\left\|#1\right\|}}
\newcommand*\lrabs[1]{{\left|#1\right|}}

\newcommand*\cvec[2]{\begin{bmatrix} #1\\#2\end{bmatrix}}

\newcommand*\Ts[0]{{T_{sync}}}

\newcommand*\ind[1]{{\mathbbm{1}\lrbb{#1}}}

\renewcommand*{\Re}{\mathbb{R}}

\newcommand*{\nablat}{\Delta}
\newcommand*{\twocase}[4]{\left\{\begin{array}{ll}
        #1 & #2\\
        #3 & #4
        \end{array}\right.}
        
\newcommand*{\threecase}[6]{\left\{\begin{array}{ll}
        #1, & \text{for } #2\\
        #3, & \text{for } #4\\
        #5, & \text{for } #6
        \end{array}\right.}

\usepackage{graphicx}
\usepackage{subcaption}

\newcommand{\Cf}{\alpha_f}

\renewcommand{\c}{c_{\kappa}}
\renewcommand{\L}{\mathcal{L}}
\renewcommand{\l}{{\ell}}
\newcommand{\eqrefmike}[1]{Eq.~\eqref{#1}}
\newcommand{\Rf}{\mathcal{R}_f}
\renewcommand{\epsilon}{\varepsilon}

\newcommand{\kd}{k\nu}

\newcommand{\Cd}{C}
\newcommand{\q}{{q}}
\newcommand{\Cm}{{C_{m}}}
\newcommand{\Co}{{C_{o}}}
\newcommand{\step}[2]{{\left\lfloor \frac{#1}{#2} \right\rfloor}}
\newcommand*\lv{\lVert}
\newcommand*\rv{\rVert}
\newcommand{\Ito}{\text{It\^o}}

\jmlrheading{}{2019}{}{}{}{}{Xiang Cheng, Niladri S. Chatterji, Yasin Abbasi-Yadkori, Peter L. Bartlett and Michael I. Jordan}

\ShortHeadings{Convergence Rates for Langevin Monte Carlo in the Nonconvex Setting}{Cheng, Chatterji, Abbasi-Yadkori, Bartlett and Jordan}
\firstpageno{1}

\begin{document}
    \title{Convergence Rates for Langevin Monte Carlo in the Nonconvex Setting}
    \author{\name Xiang Cheng \email x.cheng@berkeley.edu \\
        \addr Division of Computer Science\\
        University of California Berkeley, CA, USA
        \AND
        \name Niladri S. Chatterji \email chatterji@berkeley.edu \\
        \addr  Department of Physics\\
        University of California Berkeley, CA, USA
        \AND
        \name Yasin Abbasi-Yadkori \email yasin.abbasi@gmail.com \\
        \addr  VinAI\\
        Hanoi, Vietnam
        \AND
        \name Peter L. Bartlett \email peter@berkeley.edu \\
        \addr Division of Computer Science and Department of Statistics\\
        University of California Berkeley, CA, USA
        \AND
        \name Michael I.\ Jordan \email jordan@cs.berkeley.edu \\
        \addr Division of Computer Science and Department of Statistics\\
        University of California Berkeley, CA, USA}
    \editor{}
    \maketitle
    
    \begin{abstract}
        We study the problem of sampling from a distribution $p^*(x) \propto \exp\lrp{-U(x)}$, where the function $U$ is $L$-smooth everywhere and $m$-strongly convex outside a ball of radius $R$,
        but potentially nonconvex inside this ball.
        We study both overdamped and underdamped Langevin
        MCMC and establish upper bounds on the number of steps required to
        obtain a sample from a distribution that is within $\epsilon$ of
        $p^*$ in $1$-Wasserstein distance. For the
        first-order method (overdamped Langevin MCMC), the iteration complexity is
        $\widetilde{\mathcal{O}}\lrp{e^{cLR^2}d/\epsilon^2}$, where $d$
        is the dimension of the underlying space. For the second-order method
        (underdamped Langevin MCMC), the iteration complexity is
        $\widetilde{\mathcal{O}}\lrp{e^{cLR^2}\sqrt{d}/\epsilon}$ for
        an explicit positive constant $c$. Surprisingly,
        the iteration complexity for both these algorithms is only polynomial in the dimension $d$ and
        the target accuracy $\epsilon$. It is exponential, however, in the problem
        parameter $LR^2$, which is a measure of non-log-concavity of the target
        distribution.
    \end{abstract}
    
    
    \begin{section}{Introduction}
        We study the problem of sampling from a target distribution of the following form:
        \[
        p^*(x) \propto \exp\left(-U(x)\right),
        \]
        where $x \in \mathbb{R}^d$, and the \emph{potential function} $U:
        \mathbb{R}^d \mapsto \mathbb{R}$ is $L$-smooth everywhere and $m$-strongly
        convex outside a ball of radius $R$ (see detailed assumptions in Section \ref{sec:assumptions}).

        Our focus is on theoretical rates of convergence of sampling algorithms, including analysis
        of the dependence of these rates on the dimension $d$. Much of the theory of convergence
        of sampling---for example, sampling based on Markov chain Monte Carlo (MCMC) algorithms---has
        focused on asymptotic convergence, and has stopped short of providing a detailed study of dimension
        dependence.  In the allied field of optimization algorithms, a significant new literature has
        emerged in recent years on nonasymptotic rates, including tight characterizations of dimension
        dependence.  The optimization literature, however, generally stops short of the kinds of inferential
        and decision-theoretic computations that are addressed by sampling, in domains such as Bayesian
        statistics~\citep{robert2013monte}, bandit algorithms~\citep{cesa2006prediction} and adversarial online learning \citep{Bubeck-notes-2011,
            abbasi2013online}.
        
        In both optimization and sampling, while the classical theory focused on convex problems, recent
        work focuses on the more broadly useful setting of nonconvex problems.  While general nonconvex
        problems are infeasible, it is possible to make reasonable assumptions that allow theory to proceed
        while still making contact with practice.

        We will consider the class of MCMC algorithms that have access to the gradients of the potential,
        $\nabla U(\cdot)$. A particular algorithm of this kind that has received significant recent
        attention from theoreticians is the \emph{overdamped Langevin MCMC algorithm} \citep{parisi1981correlation,roberts1996exponential}.
        The underlying \emph{first-order} stochastic differential equation (henceforth SDE) is given by:
        \begin{align}
        \label{e:exactlangevindiffusion}
        dx_t = -\nabla U(x_t) dt + \sqrt{2} dB_t,
        \end{align}
        where $B_t$ represents a standard Brownian motion in $\mathbb{R}^d$.
        Overdamped Langevin MCMC (Algorithm \ref{olmcmc}) 
        is a discretization of this SDE. It is possible to show that under mild
        assumptions on $U$, the invariant distribution of the overdamped
        Langevin diffusion is given by $p^*(x)$. 
        
        The \emph{second-order} generalization of the overdamped Langevin
        diffusion is \emph{underdamped Langevin diffusion}, which can be represented
        by the following SDE:
        \begin{align*}
        \numberthis \label{e:exactunderdampedlangevindiffusiongeneral}
        d x_t &= u_t dt \,,\\
        d u_t &= -\lambda_1 u_t - \lambda_2 \nabla U(x_t) dt +\sqrt{2\lambda_1 \lambda_2} dB_t,
        \end{align*}
        where $\lambda_1,\lambda_2>0$ are free parameters. This SDE can also be discretized
        appropriately to yield a corresponding MCMC algorithm (Algorithm \ref{ulmcmc}).
        Second-order methods such as underdamped Langevin MCMC are particularly interesting
        as it has been previously observed both empirically \citep{neal} and
        theoretically \citep{cheng2017underdamped,mangoubi2017rapid} that these methods
        can be faster to converge than the classical first-order methods.
        
        In this work, we show that it is possible to sample from $p^*$ in time polynomial
        in the dimension $d$ and the target accuracy $\epsilon$ (as measured in $1$-Wasserstein
        distance).  We also show that the convergence depends exponentially on the product
        $LR^2$. Intuitively, $LR^2$ is a measure of the nonconvexity of $U$.
        Our results establish rigorously that as long as the problem is not ``too badly
        nonconvex,'' sampling is provably tractable.
        
        Our main results are presented in Theorem \ref{t:overdamped}
        and Theorem \ref{t:maintheoremunderdamped0}, and can be summarized informally as follows:
        
        \begin{theorem}[informal]
            \label{t:informal}
            Given a potential $U$ that is $L$-smooth everywhere
            and strongly-convex outside a ball of radius $R$, we can output a
            sample from a distribution which is $\epsilon$-close to
            $p^*(x)\propto\exp\left(-U(x)\right)$ in $W_1$ distance by running
            $\widetilde{\mathcal{O}}\left(e^{cLR^2}d / \epsilon^2\right)$
            steps of overdamped Langevin MCMC (Algorithm~\ref{olmcmc}), or
            $\widetilde{\mathcal{O}}\left(
            e^{cLR^2}\sqrt{d} /\epsilon \right)$ steps of underdamped Langevin MCMC
            (Algorithm~\ref{ulmcmc}). Here, $c$ is an explicit positive constant.
        \end{theorem}
        
        For the case of strongly convex $U$, it has been shown by \cite{cheng2017underdamped}
        that the iteration complexity of Algorithm~\ref{ulmcmc} is
        $\widetilde{\mathcal{O}}(\sqrt{d}/\epsilon)$, improving quadratically upon the best known
        iteration complexity of $\widetilde{\mathcal{O}}(d/\epsilon^2)$ for Algorithm~\ref{olmcmc}~\citep{durmus}. We will find this quadratic speed-up in $d$ and $\epsilon$ 
        in our setting as well (see Theorem \ref{t:overdamped}
        versus Theorem \ref{t:maintheoremunderdamped0}).
        \paragraph{Related work:}
            A convergence rate for overdamped Langevin diffusion, under
            assumptions \ref{ass:smoothness} -- \ref{ass:strongconvexity} (see Section~\ref{sec:assumptions}) has been
            established by \cite{reflectioneberle}, but the continuous-time
            diffusion studied in that paper is not implementable algorithmically.
            In a more algorithmic line of work, \cite{dalalyan} bounded the discretization error of
            overdamped Langevin MCMC, and provided the
            first nonasymptotic convergence rate of overdamped Langevin MCMC under log-concavity assumptions. This was
            followed by a sequence of papers in the strongly log-concave 
            setting~\citep[see, e.g.,][]{durmus,cheng2017convergence,
                dalalyan2017user,dwivedi2018log}.

            Our result for overdamped Langevin MCMC
            is in line with this existing work; indeed, we combine the continuous-time convergence rate
            of \citet{reflectioneberle} with a variant of the discretization error analysis by
            \citet{durmus}. The final number of timesteps needed is $\widetilde{\mathcal{O}}(e^{cLR^2} d/\epsilon^2)$,
            which is expected, as the rate of \cite{reflectioneberle} is $\mathcal{O}(e^{-cLR^2})$
            (for the continuous-time process) and the iteration complexity established by 
            \cite{durmus} is $\widetilde{\mathcal{O}}(d/\epsilon^2)$.
            
            
            On the other hand, convergence of underdamped Langevin MCMC under (strongly) log-concave assumptions was first established by \citet{cheng2017underdamped}. Also very relevant to our results is the work of \citet{eberle2017couplings},
            who demonstrated a contraction property of the continuous-time process stated in Eq.~\eqref{e:exactunderdampedlangevindiffusiongeneral}. That result deals,
            however, with a much larger class of potential functions, and accordingly the distance to the invariant distribution scales exponentially
            with dimension $d$. Our analysis
            yields a more favorable result by combining ideas from both \citet{eberle2017couplings} and \citet{cheng2017underdamped},
            under new assumptions; see Section \ref{sec:underdampedsection} for a full discussion.
            
            Also noteworthy is the fact that the problem of sampling from non-log-concave distributions has
            been studied by \cite{raginsky2017non}, but under weaker assumptions, with a worst-case
            convergence rate that is exponential in $d$. In \cite{xu2018global}, this technique is used to study the application of Stochastic Gradient Langevin Diffusion (and its variance-reduced version) to nonconvex optimization. Similarly, \citet{durmus2017nonasymptotic} analyze the overdamped Langevin MCMC algorithm under the assumption that $U$ is superlinear outside a ball. This is more general than our assumption of ``strong convexity outside a ball''; in this setting, the authors prove a rate that is exponential in dimension. On the other hand, \cite{ge2017beyond}
            established a $poly(d,1/\epsilon)$ convergence rate for sampling from a distribution
            that is close to a mixture of Gaussians, where the mixture components have the same variance
            (which is subsumed by our assumptions).
            
            Finally, there is a large class of sampling algorithms known as
            Hamiltonian Monte Carlo (HMC), which involve Hamiltonian dynamics in
            some form. We refer to \citet{ma} for a survey of the results in this
            area. Among these, the variant studied in this paper (Algorithm
            \ref{ulmcmc}), based on  the discretization of the SDE in
            Eq.~\eqref{e:exactunderdampedlangevindiffusiongeneral}, has a natural physical
            interpretation as the evolution of a particle's dynamics under a viscous force
            field. This model was first studied by
            \citet{kramers1940brownian} in the context of chemical reactions. The
            continuous-time process has been studied
            extensively~\citep{herau2002isotropic,dric2009hypocoercivity,eberle2017couplings,
                gorham2016measuring,baudoin2016wasserstein,bolley2010trend,
                calogero2012exponential, dolbeault2015hypocoercivity,
                mischler2014exponential}. Four recent papers---\citet{mangoubi2017rapid},
            \citet{lee2017convergence}, \citet{mangoubi2018dimensionally} and  \cite{deligiannidis2018randomized}---study the convergence rate of (variants of) HMC under
            log-concavity assumptions. In \cite{eberle2019quantitative}, the authors study the convergence of HMC on general metric state spaces. \citet{couplingseberle} study the convergence of HMC under assumptions similar to ours, and prove a convergence rate that  depends on $e^{cLR^2}$ for some constant $c$. We remark that the algorithm studied in this case is different from the underdamped Langevin MCMC algorithm, because of the incorporation of an accept-reject step.
            
    \end{section}
    
    \begin{section}{Notation, definitions and assumptions}
        In this section, we present the basic definitions, notational conventions and assumptions used throughout the paper.
        For $q \in \mathbb{N}$ we let $\lv v \rv_q$ denote the $q$-norm of a vector $v \in \mathbb{R}^d$. Throughout the paper
        we use $B_t$ to denote standard Brownian motion~\citep[see, e.g.,][]{brownian}. 
        
        \begin{subsection}{Assumptions on the potential $U$}
            \label{sec:assumptions}
            We make the following assumptions on the \emph{potential function} $U$:
            \begin{enumerate}[label=(A{\arabic*})]
                \item \label{ass:smoothness}The function $U$ is continuously-differentiable on $\mathbb{R}^d$ and has
                Lipschitz-continuous gradients; that is, there exists a positive constant $L>0$ such that for all $x,y\in \mathbb{R}^d$,
                \begin{align*}
                \lVert \nabla U(x) - \nabla U(y) \rVert_2 \le L \lVert x-y\rVert_2.
                \end{align*}
                \item \label{ass:globalminima}The function has a stationary point at zero:
                \begin{align*}
                \nabla U(0) = 0.
                \end{align*}
                \item \label{ass:strongconvexity}The function is strongly convex outside of a ball; that is, there exist
                constants $m,R>0$ such that for all $x,y \in \mathbb{R}^d$ with $\lVert x-y \rVert_2 > R$, we have:
                \begin{align*}
                \langle \nabla U(x) - \nabla U(y),x-y \rangle \ge m \lVert x-y\rVert_2^2.
                \end{align*}
            \end{enumerate}
            Finally we define the condition number as $\kappa : = L/m$. Observe that Assumption \ref{ass:globalminima}
            is imposed without loss of generality, because we can always find a stationary point in polynomial time and shift the coordinate
            system so that this stationary point of $U$ is at zero. These conditions are similar to the assumptions
            made by \citet{reflectioneberle}. Note that crucially Assumption \ref{ass:strongconvexity} is \emph{strictly stronger}
            than the assumption made in recent papers by \citet{durmus2017nonasymptotic}, \citet{raginsky2017non} and \citet{zhang2017hitting}. To see this observe that these papers only require Assumption \ref{ass:strongconvexity} to hold for a
            fixed $y=0$, while we require this condition to hold for all $y\in \mathbb{R}^d$.  One can also think of the
            difference between these two conditions as being analogous to the difference between strong convexity
            (outside a ball) and one-point strong convexity (outside a ball). 
        \end{subsection}
        
        \begin{subsection}{Coupling and Wasserstein distance}
            Denote by $\mathcal{B}(\mathbb{R}^d)$ the Borel $\sigma$-field of $\mathbb{R}^d$. Given probability measures
            $\mu$ and $\nu$ on $(\mathbb{R}^d,\mathcal{B}(\mathbb{R}^d))$, we define a \emph{transference plan} $\zeta$
            between $\mu$ and $\nu$ as a probability measure on $(\mathbb{R}^d \times \mathbb{R}^d,\mathcal{B}(\mathbb{R}^d\times
            \mathbb{R}^d))$ such that for all sets $A \in \mathcal{B}(\mathbb{R}^d)$, $\zeta(A\times \mathbb{R}^d) = \mu(A)$
            and $\zeta( \mathbb{R}^d \times A) = \nu(A)$. We denote by $\Gamma(\mu,\nu)$ the set of all transference plans.
            A pair of random variables $(X,Y)$ is called a \emph{coupling} if there exists a $\zeta \in \Gamma(\mu,\nu)$
            such that $(X,Y)$ are distributed according to $\zeta$. (With some abuse of notation, we will also refer to
            $\zeta$ as the coupling.)
            
        \end{subsection}
        
    \end{section}
    
    \section{Overdamped Langevin diffusion} 
    \label{sec:overdampedcont}
    
      In this section, we study \emph{overdamped Langevin diffusion}, given by the following stochastic differential equation (SDE):
    \begin{align} \label{e:exactoverdampedlangevindiffusion}
    dy_t &= -\nabla U(y_t) dt + \sqrt{2} dB_t.
    \end{align}
    It can be readily verified that the
    invariant distribution of the  SDE is $p^*(y)\propto e^{-U(y) }$, which ensures that the marginal along $y$ is the distribution that we are interested in.
    Based on \eqrefmike{e:exactoverdampedlangevindiffusion}, we define the discretized overdamped Langevin diffusion as
    \begin{align} \label{e:discreteoverdampedlangevindiffusion}
    dx_t &= -\nabla U\lrp{x_{\step{t}{\delta}}} dt + \sqrt{2} dB_t,
    \end{align}
    where $\delta$ is the step-size of the discretization and $\lfloor \cdot \rfloor$ denotes the floor function. 
    
    Our first result, stated as Theorem \ref{t:overdamped}, 
    establishes the rate at which the distribution of the solution of \eqrefmike{e:discreteoverdampedlangevindiffusion} converges to $p^*$.
    The SDE in \eqrefmike{e:discreteoverdampedlangevindiffusion} is implementable as Algorithm~\ref{olmcmc}.
    \begin{algorithm}[th]
        \caption{Overdamped Langevin MCMC} \label{olmcmc}
        \SetKwInOut{Input}{Input~}
        \SetKwInOut{Output}{Output}
        \Input{Step-size $\delta<1$, number of iterations $n$, initial point $x_0 = x^{(0)}$, and gradient oracle $\nabla U(\cdot)$.}
        \For {$i=0,1,\ldots,n-1$} 
        {
            Sample $x_{(i+1)\delta}\sim \mathcal{N}\lrp{x_{i\delta} - \delta \nabla U(x_{i\delta}), 2\delta I_{d\times d}}$
        }
    \end{algorithm} 
    It can be verified that $x_{i\delta}$ in Algorithm~\ref{olmcmc} and the solution to the SDE in Eq.~\eqref{e:discreteoverdampedlangevindiffusion} at time $t=i\delta$ have the same distribution. 
    The following theorem establishes a convergence rate for Algorithm~\ref{olmcmc}.
    
    \begin{theorem}\label{t:overdamped}
        Assume that $m \ge \frac{\exp\lrp{-LR^2/2}}{R^2}$, and let $0<\epsilon \leq \frac{d R^2}{\sqrt{d/m + R^2}}$ be the desired accuracy. Also let the initial point $x^{(0)}$ be such that $\lv x^{(0)}\rv_2 \le R$. Then if the step size scales as:
        \begin{align*}
        \delta = \frac{\epsilon^2 \exp\lrp{-LR^2}}{2^{10}R^2 d},
        \end{align*}
        and number of iterations scales as:
        \begin{align*}
        n = \widetilde{\Omega}\left(  \exp\lrp{\frac{3 LR^2}{2}} \cdot \frac{d}{\epsilon^2}\right),
        \end{align*}
        we have the following guarantee:
        \begin{align*}
        W_1\lrp{p_{n\delta}, p^*} \leq \epsilon,
        \end{align*}
        where $p_{n\delta}$ is the distribution of $x_{n\delta}$ in Algorithm~\ref{olmcmc} and the distribution $p^*(y) \propto e^{-U(y)}$.
    \end{theorem}    For potentials where $LR^2$ is a constant, the number of iterations taken by overdamped MCMC scales as $\widetilde{\Omega}(d/\epsilon^2)$. This matches the rate obtained in the strongly log-concave setting by \citet{durmus}.
        
    Intuitively, $LR^2$ measures the extent of nonconvexity. When this quantity is large, it is possible for $U$ to contain numerous local minima that are deep. It is therefore reasonable that the runtime of the algorithm should be exponential in this quantity.

    The assumption on the strong convexity parameter, $m$, is made to simplify the presentation of the theorem. Note that this assumption is without loss of generality, since we can always take the radius $R$ to be sufficiently large in Assumption \ref{ass:strongconvexity}. Similarly, our assumption on the target accuracy can also be easily removed, but we make this assumption in the interest of clarity. 

    The proof of Theorem \ref{t:overdamped} is relegated to Appendix~\ref{s:overdamped_proof}. The proof follows by carefully combining the continuous-time argument of \cite{reflectioneberle} together with the discretization bound of \cite{durmus}.

    \begin{section}{Underdamped Langevin diffusion}\label{sec:underdampedsection}
        In this section, we present our results for \emph{underdamped Langevin diffusion}. 
         The underdamped Langevin diffusion is a  second-order stochastic process described by the following SDE:
        \begin{align} \label{e:exactunderdampedlangevindiffusion}
        dy_t &= v_t dt,\\
        \nonumber
        dv_t &= -2 v_t -\frac{\c}{L}\nabla U(y_t) dt + \sqrt{\frac{4\c}{L}} dB_t,
        \end{align}
        where we define the constant:
        \begin{align*}
        \numberthis \label{d:c}
        \c :=& 1/\lrp{1000 \kappa},
        \end{align*}
        where $\kappa = L/m$ is the condition number.
        Similar to the case of overdamped Langevin diffusion, it can be verified that the
        invariant distribution of the  SDE is $p^*(y, v)\propto e^{-U(y) -  \frac{L}{2\c} \|v\|_2^2}$.
        This ensures that the marginal along $y$ is the distribution that we are interested in. Based on the SDE in \eqrefmike{e:exactunderdampedlangevindiffusion}, we define the discretized underdamped Langevin diffusion as:
        \begin{align} \label{e:discreteunderdampedlangevindiffusion}
        dx_t &= u_t dt,\\
        \nonumber
        du_t &= -2 u_t -\frac{\c}{L}\nabla U\lrp{x_{\step{t}{\delta}\delta}} dt + \sqrt{\frac{4\c}{L}} dB_t,
        \end{align}
        where $\delta$ is the step size of discretization. 
        The SDE in \eqrefmike{e:discreteunderdampedlangevindiffusion}  is implementable as the following algorithm:
        
        \begin{algorithm}[h] 
            \caption{Underdamped Langevin MCMC} \label{ulmcmc}
            \SetKwInOut{Input}{Input~}
            \SetKwInOut{Output}{Output}
            \Input{Step-size $\delta<1$, number of iterations $n$, initial point $(x^{(0)},0)$, smoothness parameter $L$, condition number $\kappa$ and gradient oracle $\nabla U(\cdot)$.}
            
            \For {$i=0,1,\ldots,n-1$} 
            {
                Sample $(x_{(i+1)\delta},u_{(i+1)\delta})\sim Z^{(i)}(x_{i\delta},u_{i\delta})$
            }
        \end{algorithm} 
        In this algorithm $Z^{(i)}(x_{i\delta}, u_{i\delta})\in \Re^{2d}$ is a Gaussian random vector with the following mean and covariance (which are functions of the previous iterates $(x_{i\delta},u_{i\delta})$):
        \begin{align*}
        &\E{u_{(i+1)\delta}} = u_{i\delta} e^{-2 \delta} - \frac{\c}{2 L}(1-e^{-2 \delta}) \nabla U(x_{i\delta})\,,\\
        &\E{x_{(i+1)\delta}}  = x_{i\delta} + \frac{1}{2}(1-e^{-2 \delta})u_{i\delta} - \frac{\c}{2 L} \left( \delta - \frac{1}{2}\left(1-e^{-2 \delta}\right) \right) \nabla U(x_{i\delta}) \,,\\
        &\E{\left(x_{(i+1)\delta} - \E{x_{(i+1)\delta}}\right) \left(x_{(i+1)\delta} - \E{x_{(i+1)\delta}}\right)^{\top}}= \frac{\c}{ L } \left[\delta-\frac{1}{4}e^{-4\delta}-\frac{3}{4}+e^{-2\delta}\right] \cdot I_{d\times d}\,,\\
        &\E{\left(u_{(i+1)\delta} - \E{u_{(i+1)\delta}}\right) \left(u_{(i+1)\delta} - \E{u_{(i+1)\delta}}\right)^{\top}} = \frac{\c}{ L}(1-e^{-4 \delta})\cdot I_{d\times d}\,,\\
        &\E{\left(x_{(i+1)\delta} - \E{x_{(i+1)\delta}}\right) \left(u_{(i+1)\delta} - \E{u_{(i+1)\delta}}\right)^{\top}}= \frac{\c}{2 L} \left[1+e^{-4\delta}-2e^{-2\delta}\right] \cdot I_{d \times d}\;.
        \end{align*}
        We show that the iterates at round $i$ of Algorithm \ref{ulmcmc} and the solution to the SDE in \eqrefmike{e:discreteunderdampedlangevindiffusion} at time $t=i\delta$ have the same distribution
        (see Lemma~\ref{l:gaussianexpressionforsamplingxtut} in Appendix \ref{ass:marginal_correctness}). 
        
        In Theorem \ref{t:maintheoremunderdamped0},
        we establish a bound on the rate at which the distribution of the iterates produced by this algorithm converge to the target distribution $p^*$.
        \begin{theorem}\label{t:maintheoremunderdamped0}
            Assume that $m\ge \frac{\exp\lrp{-6 LR^2}}{64 R^2}$ and let $0<\epsilon \leq \frac{d R^2}{\sqrt{d/m + R^2}}$ be the desired accuracy. Also let the initial point $x^{(0)}$ be such that $\lv x^{(0)}\rv_2 \le R$. Assume also that $e^{72LR^2} \geq 2$.
            
            Then if the step size scales as:
            \begin{align*}
            \delta = \frac{\epsilon}{R + \sqrt{d/m}} \cdot e^{-12LR^2} \cdot 2^{-35} \min \lrp{\frac{1}{LR^2}, \frac{1}{\kappa}},
            \end{align*}
            and the number of iterations as:
            \begin{align*}
            n = \widetilde{\Omega} \lrp{\frac{\sqrt{d}}{\epsilon} \exp\left(18LR^2\right)},
            \end{align*}
            we have the guarantee that
            \begin{align*}
            W_1\lrp{p_{n\delta}, p^*} \leq \epsilon,
            \end{align*}
            where $p_{n\delta}$ is the distribution of $x_{n\delta}$ and we have $p^*(y) \propto e^{-U(y)}$.
        \end{theorem}
        If we consider potentials for which $LR^2$ is a constant, the iteration complexity of underdamped Langevin MCMC grows as $\widetilde{\mathcal{O}}(\sqrt{d}/\epsilon)$, which is a quadratic improvement over the first-order overdamped Langevin MCMC algorithm. Again, the iteration complexity grows exponentially in $LR^2$ which is to be expected. As before, the condition on the strong convexity parameter and the target accuracy is made in the interest of clarity and can be removed. 
        
        The heart of the proof of this theorem is a somewhat intricate coupling argument. We begin by defining two processes, $(x_t,u_t)$ and $(y_t,v_t)$, and then couple them appropriately. The first set of variables, $(x_t,u_t)$, represent a solution to the discretized SDE in Eq.~\eqref{e:discreteunderdampedlangevindiffusion}. On the other hand, the variables $(y_t,v_t)$  represent a solution of the continuous-time SDE in Eq.~\eqref{e:exactunderdampedlangevindiffusion} with the initial conditions being $(y_0,v_0)\sim p^*(y,v)$. Thus the variables $(y_t,v_t)$ evolve according to the invariant distribution for all $t>0$. The noise that underlies both processes is \emph{coupled}, and with an appropriate choice of a Lyapunov function we are able to demonstrate that the distributions of these variables converge in $1$-Wasserstein distance.

        We present the coupling construction and a proof sketch in the subsequent sections. We relegate most of the technical details to the appendix.
        
        \begin{subsection}{A coupling construction}\label{ss:coupling_underdamped}
        
            Let $\beta=1/poly(L,1/m,d,R,1/\Cm)$ be a small constant (see proof of Theorem \ref{t:maintheoremunderdamped0} for the exact value), and let $\ell(x)=\q(\lrn{x}_2)$ be a smoothed approximation of $\lrn{x}_2$ at a scale of $\beta$, as defined in \eqref{d:ell}. 
            
            Additionally, let $\nu=1/poly(L,1/m,d,R,1/\Cm)$ be another small constant (see proof of Theorem \ref{t:maintheoremunderdamped0} for the exact value). In designing our coupling, we ensure that certain values are only updated at intervals of size $\nu$. These are needed to ensure that the stochastic process that we work with is sufficiently regular. 
            
            While reading the proofs it might be convenient for the reader to think of both $\beta$ and $\nu$ to be arbitrarily close to zero, and to think of $\ell(x)$ as equal to $\lrn{x}_2$; $\beta$ and $\nu$ do not impact the bound on the iteration complexity in Theorem \ref{t:maintheoremunderdamped0}. For a detailed discussion see Appendix~\ref{ss:small_constants}.
            
            We define a time $\Ts$ as
            \begin{align}
            \label{d:ts}
            \Ts :=& \frac{3 \log 100}{\c^2}.
            \end{align}
            We then choose $\nu$ to be such that $\frac{\Ts}{\nu}$ is a positive integer,
            and define the constant
            \begin{align*}
            \numberthis \label{d:cm}
            \Cm 
            :=& \min\lrbb{\frac{e^{-6LR^2}}{6000\kappa (1+L R^2)}, \frac{e^{-6LR^2}}{200 \Ts}, \frac{\c^2}{3}}\\
            =& \min\lrbb{\frac{e^{-6LR^2}}{2^{13}\kappa (1+L R^2)}, \frac{e^{-6LR^2}}{2^{29} \cdot  \log\lrp{100} \cdot \kappa^2}, \frac{1}{2^{22} \kappa^2}}.
            \end{align*}
            This constant $\Cm$ will be the rate at which our Lyapunov function contracts.  
            
            With these definitions in place we are ready to define a coupling between variables $(x_t, u_t)$ that evolve according to the discretized process described in Eq.~\eqref{d:xt}, and variables $(y_t, v_t)$ that evolve according to  the SDE in Eq.~\eqref{d:yt}.
            
            Let the initial conditions for these processes be given by,
            \begin{align*}
            (x_0,u_0) &= (x^{(0)},0),\\
            (y_0, v_0) & \sim p^*(y,v).
            \numberthis \label{d:x0}
            \end{align*}
            Define a variable $\tau_t$ that will be useful in determining how the noise underlying the processes is coupled.
            We initialize this variable as follows:  $\tau_0 = 0$, if $\sqrt{\lrn{x_0 - y_0}_2^2 + \lrn{x_0 - y_0 + u_0 - w_0}_2^2} \geq \sqrt{5} R$, and $\tau_0 = -\Ts$ otherwise. 
            
            Let $A_t$ and $B_t$ denote independent $d$-dimensional Brownian motions. We then let the complete set of variables $\lrp{x_t, u_t, y_t, v_t, \tau_{\step{t}{\nu}}}$ evolve according to the following stochastic dynamics:
            \begin{align*}
            \numberthis \label{d:xt}
            & d x_t = u_t dt\\
            \numberthis \label{d:ut}
            & d u_t = -2 u_t dt - \frac{\c}{L}\nabla U\lrp{x_{\step{t}{\delta} \delta}} dt + 2\sqrt{\frac{\c}{L}} dB_t\\
            \numberthis \label{d:yt}
            & d y_t = v_t dt\\
            \numberthis \label{d:vt}
            & d v_t = -2 v_t - \frac{\c}{L}\nabla U(y_t) dt + 2\sqrt{\frac{\c}{L}} dB_t \\
            &\qquad \qquad - \ind{k\nu \geq \tau_{\step{t}{\nu}} + \Ts} \cdot \lrp{4\sqrt{\frac{\c}{L}}\gamma_t \gamma_t^T dB_t + 2 \sqrt{\frac{\c}{L}} \bar{\gamma}_t \bar{\gamma}_t^T dA_t},
            \end{align*}
            where the functions $\mathcal{M}$, $\gamma_t$ and $\bar{\gamma}_t$ are defined as follows:
            \begin{align*}
            \mathcal{M}(r) := &
            \threecase{1}{r \in [\beta, \infty)}{\frac{1}{2} + \frac{1}{2} \cos \lrp{r \cdot \frac{2\pi}{\beta}}}{r \in[\beta/2, \beta]}{0}{r \in[0,\beta/2]}\\
            \gamma_t :=& \lrp{\mathcal{M}\lrp{\lrn{z_t + w_t}_2}}^{1/2} \frac{z_t + w_t}{\lrn{z_t + w_t}_2}\\
            \bar{\gamma}_t :=& \lrp{1 - \lrp{1-2\mathcal{M}\lrp{\lrn{z_t + w_t}_2}}^2}^{1/4}\frac{z_t + w_t}{\lrn{z_t + w_t}_2},
            \numberthis \label{d:gamma}
            \end{align*}
            and where for convenience we have defined
            \begin{align*}
            &z_t := x_t - y_t \\
            &w_t:= u_t-v_t. \numberthis \label{d:z}
            \end{align*}    
            Note that the function $\mathcal{M}$ essentially is a Lipschitz approximation to the indicator function $\ind{r > 0}$.
            
            Let us unpack the definition of the SDE. First, note that when the indicator $\ind{k\nu \geq \tau_{\step{t}{\nu}} + \Ts}$ is equal to zero,  then both $(x_t,u_t)$ and $(y_t,v_t)$ are evolved by the same Brownian motion $B_t$. This is called a \emph{synchronous coupling} between the processes. 
            
            Second, when this indicator is equal to one, the processes are evolved by the same Brownian motion in the directions perpendicular to $z_t + w_t$, and (roughly) by the reflected Brownian motion along the direction $z_t + w_t$. This is called a \emph{reflection coupling} between the two processes.

            In the following lemma, we show that the variables $(y_t, v_t)$ have the same marginal distributions as the solution to the SDE defined in  Eq.~\eqref{d:yt}. 
            \begin{lemma}\label{l:marginal_yt}
            The dynamics in defined by Eq.~\eqref{d:yt} and Eq.~\eqref{d:vt} is distributionally equivalent to the dynamics defined by Eq.~\eqref{e:exactunderdampedlangevindiffusion}.
            \end{lemma}
            We give the proof in Appendix \ref{ass:marginal_correctness}. It is easy to verify that $(x_t, u_t)$ have the same marginal distribution as the solution to the SDE defined in Eq.~\eqref{d:xt} so we omit the proof.
            
            Finally, we define an update rule for $\tau$ which dictates how the noise is coupled. For any $k\in \Z^+$, $\tau_k$ is defined as follows:
            \begin{align*}
            & \tau_k := \twocase{k\nu}{\text{if } \lrp{k\nu - \tau_{k-1} \geq \Ts \ \text{AND}\ \sqrt{\lrn{z_{k\nu}}_2^2 + \lrn{z_{k\nu} + w_{k\nu}}_2^2} \geq \sqrt{5} R}}{\tau_{k-1}}{otherwise.}
            \numberthis \label{d:taut}
            \end{align*}
            
            From the dynamics in Eq.~\eqref{d:vt}, we see that $\tau_k$ is used for determining whether $(x_t,y_t,u_t,v_t)$ evolves by synchronous or reflection coupling over the interval $t\in[k\nu, (k+1)\nu)$. From its definition in Eq.~\eqref{d:taut}, we see that, roughly speaking, $\tau_k$ is ``the last time (up to $k\nu$) that $(z_t, w_t)$ ends up outside the ball $\sqrt{\lrn{z_{t}}_2^2 + \lrn{z_{t} + w_t}_2^2} = \sqrt{5}R$,'' but with a caveat: we do not update the value of $\tau_k$ more than once in a $\Ts$ interval of time.

            Let $\lrp{\Omega, \F_t, P}$ be the probability space, where $\F_t$ is the $\sigma$-algebra generated by $(y_0, v_0)$, $B_s$ and $A_s$ for all $s\in[0,t)$. In the following Lemma, we prove that $\lrp{x_t, u_t, v_t, y_t, \tau_{\step{t}{\nu}}}$ has a unique strong solution $(x_t,u_t,y_t,v_t,\tau_{\step{t}{\nu}})(\omega)$ ($\omega \in \Omega$), which is adapted to the filtration $\F_t$. Furthermore, with probability one, $(x_t,u_t,y_t,v_t)(\omega)$ is $t$-continuous:
        \begin{lemma}\label{l:existence_of_process}
            Let $B_t$ and $A_t$ be two independent Brownian motions, and let $\F_t$ be the $\sigma$-algebra generated by $B_s$, $A_s$; $s\leq t$, and $(x_0,u_0,y_0,v_0)$.
            
            For all $t\ge 0$, the stochastic process $(x_t, u_t, y_t, v_t, \tau_{\step{t}{\nu}}) (\omega)$ defined in Eqs.~\eqref{d:xt}--\eqref{d:taut} has a unique solution such that $(x_s, u_s, y_s, v_s)$ is $t$-continuous with probability one, and satisfies the following, for all $s\ge 0$,
            \begin{enumerate}
                \item $(x_s, u_s, y_s, v_s, \tau_{\step{s}{\nu}})$ is adapted to the filtration $\F_s$.
                \item $\E{\lrn{x_s}_2^2 + \lrn{y_s}_2^2 + \lrn{u_s}_2^2 + \lrn{v_s}_2^2} \leq \infty$.
            \end{enumerate}
        \end{lemma}
        We defer the proof of this lemma to Appendix~\ref{ass:existence_coupling}.
            
            Finally, for notational convenience, we define the following quantities, for any $k \in \Z^+$:
            \begin{align*}
            \numberthis \label{d:mut}
            \mu_k :=& \ind{k\nu \geq \tau_{k} + \Ts}\\
            \numberthis \label{d:r}
            r_t :=& \lrp{1+2\c}\l(z_t) + \l(z_t + w_t)\\
            \nabla_t :=& \nabla U(x_{t}) - \nabla U(y_t)\\
            \nablat_t :=& \nabla U(x_{\step{t}{\delta} \delta}) - \nabla U(x_t).
            \numberthis \label{d:nablat}
            \end{align*}
            As described above, when $\mu_k=0$ the processes are synchronously coupled, and when $\mu_k=1$ they are coupled via reflection coupling. Roughly, $r_t$ corresponds to the sum of $\lv z_t \rv_2$ and $\lv z_t+w_t \rv_2$. $\nabla_t$ is the difference of the gradients of $U$ at $x_t$ and $y_t$, while $\nablat_t$ is the difference of the gradients at $x_{\step{t}{\delta} \delta}$ and $x_t$.
        \end{subsection}

        \begin{subsection}{Lyapunov Function}
            In this section, we define a Lyapunov function that will be useful in demonstrating that the distributions of $(x_t,u_t)$ and $(y_t,v_t)$ converge in 1-Wasserstein distance.
            
            We follow \citet{reflectioneberle} in our specification of the \emph{distance function} $f$ that 
            is used in the definition of our Lyapunov function. We define two constants,
            \begin{align*}
            \Cf := \frac{L}{4}, \qquad \text{and, }\qquad \Rf := 12 R,
            \numberthis \label{d:fparameters}
            \end{align*}
            and auxiliary
            functions $\psi(r)$, $\Psi(r)$ and $g(r)$, all from $ \Re^+$ to $\Re^+$:
            \begin{align}
            \nonumber
            & h(r) := \threecase{1}{ r \in[0, \Rf]}{1 - \frac{1}{\Rf}\lrp{r-\Rf}}{r\in[\Rf, 2\Rf]}{0}{r\in [2\Rf, \infty)} \\
            \label{d:psietal}& \psi(r) := e^{- 2\Cf \int_0^r h(s) ds}\,, \qquad \Psi(r) := \int_0^r \psi(s) ds\,, \qquad \\
            \nonumber
            & \qquad g(r) := 1- \frac{1}{2}
            \frac{\int_0^{r} h(s) \frac{\Psi(s)}{\psi(s)} ds}{\int_0^{\infty} h(s)\frac{\Psi(s)}{\psi(s)} ds}\,.
            \end{align}
            Let us summarize some important properties of the functions $\psi$ and $g$:
            \begin{itemize}
                \item $\psi$ is decreasing, $\psi(0) = 1$, and $\psi(r) = \psi(2\Rf)$ for any $r>2\Rf$.
                \item $g$ is decreasing, $g(0) = 1$, and $g(r) = \frac{1}{2}$ for any $r>2\Rf$.
            \end{itemize}
            Finally we define $f$ as
            \begin{equation}
            \label{d:f}
            f(r) := \int_0^r \psi(s) g(s) ds.
            \end{equation}
            In Lemma \ref{lem:fpropertiesall} in Appendix \ref{s:onf}, we state and prove various several useful properties of the distance function $f$.
        
            Additionally define the stochastic processes:
            \begin{align}
            \label{d:xi}
            \xi_t =& \int_0^t e^{-\Cm (t-s)} \c{\lrn{x_s-x_{\step{s}{\delta} \delta}}_2} ds,\\
            \label{d:sigma}
            \sigma_t =& \int_0^t \mu_{\step{s}{\nu}}\cdot e^{-\Cm (t-s)} \cdot \ind{r_s \geq \sqrt{12}R} 4 r_s ds,\\
            \phi_t =& \int_{0}^{t} \mu_{\step{s}{\nu}} \cdot e^{-\Cm (t-s)} \lin{\nabla_{w_s} \lrp{f(r_s)}, 4\sqrt{\frac{\c}{L}} \lrp{\gamma_s \gamma_s^T dB_s + \frac{1}{2} \bar{\gamma}_s \bar{\gamma}_s^T dA_s}}. \label{d:phi}
            \end{align}
            These processes essentially track the discretization error arising due to a finite step size $\delta$ and $\nu$. We refer to Lemma \ref{l:existence_of_phi} in Appendix \ref{ass:existence_coupling} for a proof of existence of $\phi_t$.

            Then following stochastic process $\L_t$  acts as our Lyapunov function:
            \begin{align*}
            \numberthis \label{d:L}
            \L_t
            &:= \mu_{k} \cdot \lrp{f\lrp{r_t} - \xi_{t}}
            + \lrp{1-\mu_{k}} \cdot \exp\lrp{-\Cm\lrp{t-\tau_{k}}} \cdot  \lrp{f\lrp{r_{\tau_{k}}}- \xi_{\tau_k} } 
            - \lrp{\sigma_t + \phi_t},
            \nonumber
            \end{align*}
            where $k:= \step{t}{\nu}$. Note that $\L_t$ (the Lyapunov function at time $t$) depends on $r_{\tau_{k}}$ (at time $\tau_k$). In Lemma \ref{l:combining_4_cases}, we demonstrate that this function contracts at a rate of $e^{-\Cm t}$. The convergence bound then follows by showing that the convergence of this Lyapunov function implies convergence of the distributions in $1$-Wasserstein distance.
        \end{subsection}
        \begin{subsection}{Proof Sketch}
            We present a full proof of Theorem~\ref{t:maintheoremunderdamped0} in Appendix~\ref{appendix:underdamped}. In this section we provide a high-level sketch of our proof. 
            
            The proof proceeds by a path-wise analysis of the evolution of the Lyapunov function. In Figure~\ref{fig:sketch}, we illustrate a sample path of the process.
            \begin{figure}[H]
                \begin{subfigure}{.5\textwidth}
                \centering
                \includegraphics[scale=0.48]{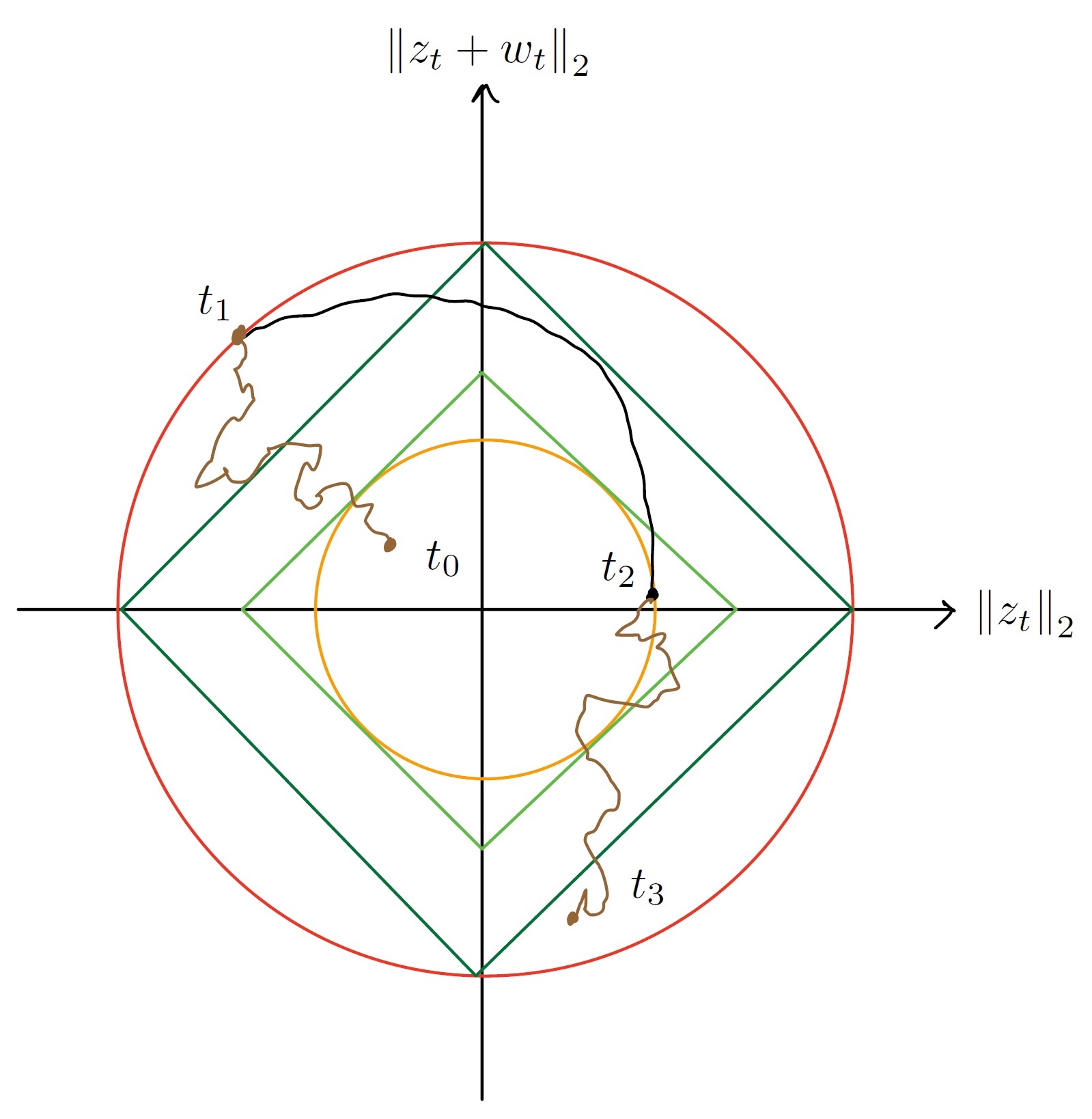}
                \caption{Illustration of Coupling} \label{fig:sketch}
                \label{fig:sub1}
                \end{subfigure}%
                \begin{subfigure}{.5\textwidth}
                \centering
                \includegraphics[scale=1.2]{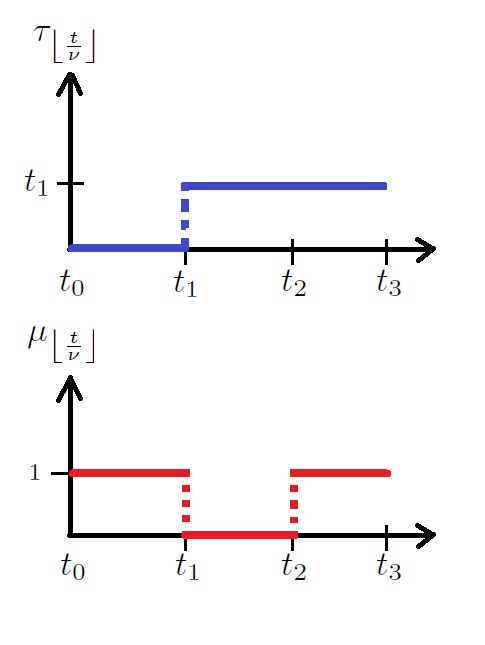}
                \caption{Update for $\tau$ and $\nu$} \label{fig:sketch}
                \end{subfigure}%
            \end{figure}
            First, let us highlight the features of the figure.
            
            \begin{enumerate}
                \item The \textcolor{red}{red circle} represents the set $\sqrt{\lrn{z_t}_2^2 + \lrn{z_t + w_t}_2^2} = \sqrt{5} R$. It affects the updates of $\tau_{\step{t}{\nu}}$, which, in turn, dictates how the processes are coupled.
                \item The \textcolor{orange}{orange circle} represents $\sqrt{\lrn{z_t}_2^2 + \lrn{z_t + w_t}_2^2} = \frac{23}{50} \cdot \sqrt{5} R$. In relation to the \textcolor{red}{red circle}, it represents the contraction of $\sqrt{\lrn{z_t}_2^2 + \lrn{z_t + w_t}_2^2}$ when evolved according to synchronous coupling.
                \item The \textcolor{OliveGreen}{dark green diamond} represents $(1+2\c) \lrn{z_t}_2 + \lrn{z_t + w_t}_2 = \sqrt{5} R$. It is a lower bound on $(1+2\c) \lrn{z_t}_2 + \lrn{z_t + w_t}_2 = \sqrt{5} R$ when $\sqrt{\lrn{z_t}_2^2 + \lrn{z_t + w_t}_2^2} = \sqrt{5} R$.
                \item The \textcolor{Green}{light green diamond} represents $2 \lrp{(1+2\c) \lrn{z_t}_2 + \lrn{z_t + w_t}_2} = 2 \cdot \frac{23}{50} \sqrt{5}R$. It represents an upper bound on $(1+2\c) \lrn{z_t}_2 + \lrn{z_t + w_t}_2 = \sqrt{5} R$ when $\sqrt{\lrn{z_t}_2^2 + \lrn{z_t + w_t}_2^2} = \frac{23}{50} \cdot \sqrt{5} R$.
                \item It is not drawn, but note that the red circle is contained in $(1+2\c) \lrn{z_t}_2 + \lrn{z_t + w_t}_2 \leq \sqrt{12}R$, which is the radius used for defining $f$ in Eq.~\eqref{d:fparameters}.
                \item The \textcolor{Brown}{brown squiggly lines} ($t_0\to t_1$) and ($t_2\to t_3$) represent the evolution of the process under \emph{reflection coupling}. 
                \item The black line $t_1\to t_2$ represents the evolution of the process under \emph{synchronous coupling}.
            \end{enumerate}
            Below, we describe how $(z_t, w_t)$ evolves over $t \in [t_0, t_3]$, and illustrate the main ideas behind the proof. To simplify matters, assume that 
            \begin{enumerate}
                \item $k_i := t_i/\nu$ are integers, for $i=0,1,2,3$.
                \item $t_3 - t_2 = \Ts$.
                \item $\xi_t = \sigma_t = 0$ as these terms correspond to discretization errors.
                \item $r_t \approx \lrn{z_t}_2 + (1+2\c) \lrn{z_t + w_t}_2$.
            \end{enumerate}
            Then 
            \begin{itemize}
                \item From $t_0 \to t_1$: \\
                 Suppose that the process starts somewhere inside the red circle and stays inside for until time $t_1$, then $\tau_{\step{t}{\nu}}=t_0$ and $\mu_{\step{t}{\nu}} = 1$ for $t\in[t_0,t_1)$, and the process $(z_t, w_t)$ undergoes reflection coupling. 
                 
                 In this case, we can show that when $r_t \le \sqrt{12}R$ then $f(r_t) - \phi_t$ contracts at a rate of $\exp(-\Cm t)$  with probability one (see Lemma~\ref{l:muk=1}). This in turn implies that our Lyapunov function $\L_t$ also contracts at the same rate with probability one (see Lemma~\ref{l:mu11} and Lemma~\ref{l:mu10}).

                \item From $t_1 \to t_2$: \\
                At $t = t_1$, we update $\tau_{k_1}$ so that $\tau_{k_1} = t_1$. Thus $\mu_s=0$ for all $s\in[t_1, t_2)$. During this period, $(z_t, w_t)$ evolves under synchronous coupling. In Lemma \ref{l:muk=0_norm_contraction_2}, we show that $\sqrt{\lrn{z_{t_2}}_2^2 + \lrn{z_{t_2} + w_{t_2}}_2^2} \leq \frac{23}{50} \sqrt{\lrn{z_{t_1}}_2^2 + \lrn{z_{t_1} + w_{t_1}}_2^2}$. This implies that $f(r_{t_2}) \leq e^{-\Cm (t_2 - t_1)} f(r_{t_1})$ (Lemma \ref{l:muk=0_norm_contraction_3}). Again, this contraction is with probability one. Intuitively, we use synchronous coupling because when the value of $\lrn{z_t}_2 + \lrn{z_t + w_t}_2$ is large, Assumption \ref{ass:strongconvexity} guarantees contraction even in the absence of noise.
                
                This contraction in $f$ consequently results in a contraction of the Lyapunov function (see Lemma~\ref{l:mu00}).
                
                \item After a duration $\Ts$ of synchronous coupling, we have $\mu_{k_2} = 1$ and we resume reflection coupling over $[t_2, t_3]$. Note that at $t= t_2$, the Lyapunov function $\L_t$, undergoes a jump in value, from $\exp\lrp{-\Cm (t_2 - t_1)} f(r_{t_1})$ to $f(r_{t_2})$ (see \eqref{d:L}). We show in Lemma \ref{l:mu01} that this jump is negative with probability one. 
            \end{itemize}
        \end{subsection}
    \end{section}
    
    \section{Discussion}
In this paper, we study algorithms for sampling from distributions which satisfy a more general
structural assumption than log-concavity, in time polynomial in dimension and accuracy. We also
demonstrate that when using underdamped dynamics the runtime can be improved, mirroring the strongly
convex case.

There are a few natural questions that we hope to answer in further investigation of
non-log-concave sampling problems.  First, it would be interesting to determine other structural assumptions that
may be imposed on the target distribution that are more general than log-concavity but still admit
tractable sampling guarantees; for example, we would like to uncover assumptions that may alleviate the exponential dependence on $LR^2$. Conversely, existing guarantees may be extended to weaker assumptions,
such as weak convexity outside a ball.
Secondly, one might also wish to consider algorithms which have access to more than a gradient oracle, such as the Metropolis Hastings filter, or discretizations which use higher-order information.

\section*{Acknowledgements}

This work was supported in part by the Mathematical Data Science program of the Office of Naval Research under grant number N00014-18-1-2764.
    
\nocite{*}
\bibliography{ref}
    
    \newpage
    \appendix
    \flushleft{\textbf{\LARGE{Appendix}}}

    We outline here the organization of the Appendix.
    
    In Appendix \ref{s:index}, we list the variables used in this paper, in alphabetical order, with references to their definitions. In Appendix \ref{ss:small_constants}, we give a description of two small constants, $\beta$ and $\nu$, which are used throughout our analysis to ensure regularity in time and space. 
    
    In Appendix \ref{s:overdamped_proof}, we give a proof of Theorem \ref{t:overdamped}. In Appendix \ref{appendix:underdamped}, we give a proof of Theorem \ref{t:maintheoremunderdamped0}.
    
    In Appendix \ref{s:fproperties}, we specify the construction of the distance function $f$, which is used to demonstrate contraction. In Appendix \ref{s:bounding_energy}, we bound the moments of some of the relevant quantities; these are used in discretization bounds of Appendix \ref{s:overdamped_proof} and \ref{appendix:underdamped}. In Appendix \ref{ass:existence_coupling}, we gives proofs of the existence of our coupling constructions. In Appendix \ref{ass:marginal_correctness}, we prove that our coupling constructions have the correct marginals. We also prove that Algorithm \ref{ulmcmc} exactly implements \eqref{e:discreteunderdampedlangevindiffusion}.

    \newpage

    \newcolumntype{L}{>{\flushleft\arraybackslash}m{12cm}}
    \renewcommand{\arraystretch}{1.6}
    \begin{section}{Index of notation}\label{s:index}
        \begin{tabular}{l L}
            $\Cf$ & Parameter of $f$ \eqref{d:f}. See \eqref{d:o:fparameters}(overdamped) and \eqref{d:fparameters} (underdamped).\\
            $\beta$ & Constant in defining $\l$. See also Section \ref{ss:small_constants}.\\
            $\c$ & See \eqref{d:c}\\
            $\Cm$ & Underdamped contraction rate, see \eqref{d:cm}.\\
            $\Co$ & Overdamped contraction rate, see \eqref{d:co}.\\
            $d$ & Dimension of $x$\\
            $f$ & See \eqref{d:f}\\
            $\kappa$ & Condition number, defined after Assumption \ref{ass:strongconvexity}\\
            $\l$ & Twice continuously differentiable approximation to $\lrn{\cdot}_2$ with $\beta$ error. See Lemma \ref{l:def_l}.\\
            $L$ & Lipschitz gradient parameter, see Assumption \ref{ass:smoothness}.\\
            $\L$ & Lyapunov function. See \eqref{d:o:L} (overdamped) and \eqref{d:L} (underdamped)\\
            $m$ & Contraction parameter outside the $R$ ball. See Assumption \ref{ass:strongconvexity}\\
            $\mathcal{M}$ & See \eqref{d:o:gamma} (overdamped) \eqref{d:gamma} (underdamped)\\
            $q$ & See Lemma \ref{l:def_mr}\\
            $r$ & See \eqref{d:mut}.\\
            $R$ & See Assumption \ref{ass:strongconvexity}\\
            $\Rf$ & Parameter of $f$ \eqref{d:f}. See \eqref{d:o:fparameters}(overdamped) and \eqref{d:fparameters} (underdamped).\\
            $\Ts$ & See \eqref{d:ts}.\\
            $\tau_k$ & See \eqref{d:taut}\\
            $w_t$ & Short for $u_t - v_t$, defined in \eqref{d:z}\\
            $z_t$ & Short for $x_t - y_t$, defined in \eqref{d:z}\\
            $\gamma$ & See \eqref{d:gamma}\\
            $\mu$ & See \eqref{d:mut}\\
            $\nu$ & Coupling stepsize in underdamped Coupling, used in \eqref{d:taut}). See also Section \ref{ss:small_constants}.\\
            $\xi$ & See \eqref{d:xi}.\\
            $\sigma$ & See \eqref{d:sigma}.\\
            $\phi$ & See \eqref{d:phi}.\\
            $\nabla_t$ and $\nablat_t$ & See \eqref{d:o:nablat} and \eqref{d:nablat}
        \end{tabular}

    \end{section}

    \begin{section}{Two Small Constants}\label{ss:small_constants}
            \textbf{On $\l$ and $\beta$:}\\
            In this paper, we will take $\beta = 1/poly(L, 1/m, d, R)$ to be a small constant. See the proofs of Theorem~\ref{t:overdamped} and Theorem~\ref{t:maintheoremunderdamped0} for the exact values of $\beta$. Intuitively, $\beta$ is a radius inside of which we perform the following smoothing:
            
            We define a function $q(r)$ in \eqref{d:q}, which is a smoothed approximation of $|r|$, such that it has continuous second derivatives everywhere. Specifically, for $r\leq \beta/2$, $q(r)$ is a cubic spline.
            \begin{align*}
            \q(r) = \threecase{\frac{\beta}{3} + \frac{8}{3\beta^2} \cdot r^3}{r \in [0, \beta/4]}{\frac{5\beta}{12} -r + \frac{4}{\beta} \cdot r^2 - \frac{8}{3\beta^2} \cdot r^3}{ r \in [\beta/4, \beta/2]}{r}{r \in [\beta/2, \infty].}
            \numberthis \label{d:q}
            \end{align*}
            This allows us to define a smoothed version of $\lrn{x}_2$, which has continuous second derivatives everywhere:
            \begin{align*}
            \l(x)  = \q(\lrn{x}_2).
            \numberthis \label{d:ell}
            \end{align*}
            In various parts of our proof, we replace $\lrn{\cdot}_2$ by its smooth approximation $\l(\cdot)$, defined in Lemma~\ref{l:def_l}, parametrized by $\beta$; a small $\beta$ means that $\l(\cdot)$ and $\lrn{\cdot}_2$ are close. We need to be careful as $\l(\cdot)$ is strongly convex, with parameter $1/\beta^2$, in a $\beta/2$ radius around zero. We thus need to design our dynamics to ensure that the coupling has no noise in this region (see Eq.~\eqref{d:gamma}).
            
            When reading the proofs, it helps to think of $\l(\cdot) = \lrn{\cdot}_2$ and $\beta=0$, as we can take $\beta$ to be arbitrarily small without additional computation costs. In our proof, it suffices to let $\beta = 1/poly(L, 1/m, d, R)$.
            
            \textbf{On $\nu$:}\\
            In order to demonstrate the existence of a strong solution to the coupling presented in Section \ref{ss:coupling_underdamped} (Lemma \ref{l:existence_of_process}), we switch between synchronous and reflection coupling at deterministic, finite intervals of width $\nu$. 
            
            This is not necessary strictly speaking, as there are results that ensure the existence of solutions of an SDE when the diffusion and drift coefficients are discontinuous but have finite variation. However, we choose to use a discretized coupling as the existence of its solution can be verified by using standard results.
            
            This discretized coupling scheme adds an error term $\sigma_t$ (see Eq.~\eqref{d:sigma}). We show in Lemma~\ref{l:sigma_bound} that this is $o(\nu^2)$. 
            
            When reading the proofs, it helps to think of $\nu=0$ and $\sigma_t=0$, as we can take $\nu$ to be arbitrarily small without additional computation costs. In the proof, it suffices to let $\nu = 1/poly(L, 1/m, d, R)$. See the proof and Theorem~\ref{t:maintheoremunderdamped0} for the exact value of $\nu$.
            
            Note that $\nu$ is distinct from (and unrelated to) $\delta$, which is the step-size of the underdamped Langevin MCMC algorithm (Algorithm \ref{ulmcmc}). $\delta$, and the corresponding discretization error $\xi_t$, cannot be made arbitrarily small without additional computation costs.

        \begin{lemma}\label{l:def_l}
            For a given $\beta>0$, let $m(r)$ be as defined in Lemma~\ref{l:def_mr}. Let $\l(x) : \Re^n \to \Re^+$ be defined as in \eqref{d:ell}, reproduced below for ease of reference:
            \begin{align*}
            \l(x)  = \q(\lrn{x}_2).
            \end{align*}
            Then,
            \begin{enumerate}
                \item For all $x$, $\ell(x)$ satisfies $\beta/3 \leq \ell(x)$ and $\lrabs{\ell(x) - \lrn{x}_2} \leq \beta/3$. In addition, for $\lrn{x}_2\geq \beta/2$, $\ell(x) = \lrn{x}_2$.
                \item $\nabla \ell(x) = \q'\lrp{\lrn{x}_2} \frac{x}{\lrn{x}_2}$, for all $x$, $\lrn{\nabla \ell(x)}_2 \leq 1$, for $\lrn{x}_2 \geq \beta/2$, $\nabla \ell(x) = \frac{x}{\lrn{x}_2}$.
                \item for $\lrn{x}_2 \geq \beta/2$, $\nabla^2 \ell(x) = \q''\lrp{\lrn{x}_2} \frac{xx^T}{\lrn{x}_2^2} + \q'\lrp{\lrn{x}_2}\frac{1}{\lrn{x}_2} \lrp{I - \frac{x x^T}{\lrn{x}_2^2}} $.\\
                \item $\nabla \l(x)$ and $\nabla^2 \l(x)$ are defined everywhere and continuous. In particular, for $\lrn{x}_2 \leq \beta/2$, 
                \begin{align*}
                \lrn{\nabla \l(x)}_2 \leq 4, \text{ and } \lrn{\nabla^2 \l(x)}_2 \leq \frac{8}{\beta}.
                \end{align*}
            \end{enumerate}
        \end{lemma}
        \begin{proof}
            \begin{enumerate}
                \item Immediate from  Lemma~\ref{l:def_mr}.2.
                \item By Chain rule, $\nabla \ell(x) = q'(\lrn{x}_2)\frac{x}{\lrn{x}_2}$. Furthermore, From the Lemma~\ref{l:def_mr}.1, we verify that $\nabla \ell(x)$ is defined everywhere, including at $0$. The remaining claims follow from Lemma~\ref{l:def_mr}.3
                \item This is just chain rule, together with Lemma~\ref{l:def_mr}.1, which guarantees the existence of $q''(\lrn{x}_2)/\lrn{x}_2^2$ for all $x$.
                \item Existence and continuity follow from Lemma~\ref{l:def_mr}.
            \end{enumerate}
        \end{proof}
        
        \begin{lemma}\label{l:def_mr}
            Let $\beta$ be any positive real. Let $\q(r)$ be defined as in \eqref{d:q}, reproduced below for ease of reference:
            \begin{align*}
            \q(r) = \threecase{\frac{\beta}{3} + \frac{8}{3\beta^2} \cdot r^3}{r \in [0, \beta/4]}{\frac{5\beta}{12} -r + \frac{4}{\beta} \cdot r^2 - \frac{8}{3\beta^2} \cdot r^3}{ r \in [\beta/4, \beta/2]}{r}{r \in [\beta/2, \infty].}
            \end{align*}
            Then, 
            \begin{enumerate}
                \item $\q(r)$, $\q'(r)/r$ and $\q''(r)/r^2$ exist for all $r$, and are continuous.
                \item For all $r$, $q(r)$ satisfies $\beta/3 \leq \q(r)$ and $\lrabs{r - \q(r) } \leq \beta/3$. In addition, $\q(r) = r$ for $r\geq \beta/2$.
                \item $\q'(r)$ is monotonically nondecreasing, $\q'(r)=1$ for $r\geq \beta/2$, and $\q'(r) = 0$ for $r=0$.
                \item $\q''(r) = 0$ for all $r\geq \beta/2$.
            \end{enumerate}
        \end{lemma}
        \begin{proof}Taking derivatives, we verify that
            \begin{align*}
            \q'(r) =& \threecase{\frac{8}{\beta^2} \cdot r^2}{r \in [0, \beta/4]}{-1 + \frac{8}{\beta} \cdot r - \frac{8}{\beta^2} \cdot r^2}{ r \in [\beta/4, \beta/2]}{1}{r \in [\beta/2, \infty];}\\
            \q''(r) =& \threecase{\frac{16}{\beta^2} \cdot r}{r \in [0, \beta/4]}{\frac{8}{\beta} - \frac{16}{\beta^2} \cdot r}{ r \in [\beta/4, \beta/2]}{0}{r \in [\beta/2, \infty].}
            \end{align*}
            All the claims can then be verified algebraically.
        \end{proof}
    \end{section}
    
    \begin{section}{Proofs for overdamped Langevin Monte Carlo}\label{s:overdamped_proof}
        \begin{subsection}{Coupling construction for overdamped Langevin MCMC}
            Let $\beta$ be a small constant (see proof of Theorem \ref{t:overdamped} for the exact value), and let $\ell(x)=\q(\lrn{x}_2)$ be a smoothed approximation of $\lrn{x}_2$ as defined in \eqref{d:ell}. See Appendix~\ref{ss:small_constants} for a detailed discussion.
            
            We begin by establishing the convergence of the continuous-time process in Eq.~\eqref{e:exactlangevindiffusion}
            to the invariant distribution. Similar to \citet{reflectioneberle}, we construct a coupling 
            between the SDEs described  by Eq.~\eqref{e:exactoverdampedlangevindiffusion} and Eq.~\eqref{e:discreteoverdampedlangevindiffusion}.
            We initialize the coupling at
            \begin{align*}
            & x_0 = 0\\
            & y_0 \sim p^*(y),
            \end{align*}
            and evolve the pair $(x_t, y_t)$ according to the dynamics
            \begin{align*}
            \numberthis \label{d:o:xt}
            & d x_t = - \nabla U\lrp{x_{\step{t}{\delta}\delta}} dt + \sqrt{2} dB_t\\
            \numberthis \label{d:o:yt}
            & d y_t = \nabla U\lrp{y_{t}} dt + \sqrt{2} dB_t - 2\sqrt{2} \gamma_t \gamma_t^T dB_t + \sqrt{2} \bar{\gamma}_t \bar{\gamma}_t^T dA_t,
            \end{align*}
            where the terms $\gamma_t$ and $\bar{\gamma}_t$ are defined as:
            \begin{align*}
            \gamma_t :=& \lrp{\mathcal{M}\lrp{\lrn{z_t}_2}}^{1/2} \frac{z_t}{\lrn{z_t}_2}\\
            \bar{\gamma}_t :=& \lrp{1 - \lrp{1-2\mathcal{M}\lrp{\lrn{z_t}_2}}^2}^{1/4}\frac{z_t}{\lrn{z_t}_2},\\
            \text{with}\qquad\qquad
            z_t := & x_t - y_t,\\
            \mathcal{M}(r) := & 
            \threecase{1}{r \in [\beta, \infty)}{\frac{1}{2} + \frac{1}{2} \cos \lrp{r \cdot \frac{2\pi}{\beta}}}{r \in[\beta/2, \beta]}{0}{r \in[0,\beta/2].}
            \numberthis \label{d:o:gamma}
            \end{align*}
            We use the convention that $0/0=0$ when $\lrn{z_t}_2 = 0$. It can be verified that $\gamma_t$ and $\bar{\gamma}_t$ are Lipschitz and gradient-Lipschitz for all $z_t \in \Re^d$.
            
            In the following Lemma, we show that $y_t$ evolved according to Eq.~\eqref{d:o:yt} has the same marginal distributions as $y_t$ evolved according to the SDE in Eq.~\eqref{e:exactoverdampedlangevindiffusion}.  
            \begin{lemma}\label{l:o:marginal_yt}
            The dynamics in Eq.~\eqref{d:o:yt} is distributionally equivalent to the dynamics defined in  Eq.~\eqref{e:exactoverdampedlangevindiffusion}.
            \end{lemma}
            We defer the proof to Appendix~\ref{ass:marginal_correctness}.
            
            For notational convenience, we define
            \begin{align*}
            \nabla_t :=& \nabla U(x_{t}) - \nabla U(y_t)\\
            \nablat_t :=& \nabla U(x_{\step{t}{\delta} \delta}) - \nabla U(x_t).
            \numberthis \label{d:o:nablat}
            \end{align*}

            Finally, we construct the Lyapunov function that we will use to show convergence. Let $f(r_t)$ be as defined in Eq.~\eqref{d:f}, with
            \begin{align*}
            \Cf := \frac{L}{4}, \quad \text{and, }\quad 
            \Rf := R.
            \numberthis \label{d:o:fparameters}
            \end{align*}
            
            Define a constant, 
            \begin{align*}
            \numberthis \label{d:co}
            \Co := \min\lrbb{\frac{1}{8R^2} e^{-LR^2/2}, m},
            \end{align*}
            
            and finally, define two stochastic processes 
            \begin{align}
            \label{d:o:xi}
            \xi_t :=& L \int_0^t e^{-\Co (t-s)} {\lrn{x_s-x_{\step{s}{\delta} \delta}}_2} ds\\
            \label{d:o:phi}
            \phi_t :=& \int_{0}^{t} e^{-\Co (t-s)} f'\lrp{\lrn{z_s}_2} \lin{\frac{z_s}{\lrn{z_s}_2},  \lrp{2\sqrt{2} \gamma_s \gamma_s^T dB_s + \sqrt{2} \bar{\gamma}_s \bar{\gamma}_s^T dA_s}}.
            \end{align}
            
            With these definitions, the following stochastic process $\L_t$ acts as our Lyapunov function:
            \begin{align*}
            \numberthis \label{d:o:L}
            \L_t
            :=& f(\l(z_t)) - \xi_t - \phi_t.
            \end{align*}
        \end{subsection}
        
        \begin{subsection}{Proof of Theorem~\ref{t:overdamped}}
            The proof follows in three steps. In Step 1 we analyze the evolution of $f(\l(z_t))$ using $\Ito$'s Lemma. In Step 2 we use this to show that the Lyapunov function $\L_t$ which is defined in Eq.~\eqref{d:o:L} contracts at a sufficiently fast rate. Finally in Step 3 we relate this contraction in the Lyapunov function to a bound on the iteration complexity of Algorithm~\ref{olmcmc}.
                
                We note that the technique in establishing Step 1 is essentially taken from \cite{reflectioneberle}.
                
                \textbf{Step 1:} By $\Ito$'s Lemma applied to $f(\l\lrp{z_t})$,
                
                \begin{align*}
                d f(\l(z_t))
                &= \underbrace{\lin{\nabla_z f(\l(z_t)), -\nabla_t - \nablat_t}}_{=:\spadesuit} dt
                + \underbrace{\frac{1}{2} \tr\lrp{\nabla^2_z f(\l(z_t)) \lrp{8 \gamma_t \gamma_t^T + 2 \bar{\gamma}_t \bar{\gamma}_t^T }}}_{=:\heartsuit} dt \\
                &\qquad\qquad\qquad + \lin{\nabla_z f(\l(z_t)), 2\sqrt{2} \gamma_t \gamma_t^T dB_t + \sqrt{2} \bar{\gamma}_t \bar{\gamma}_t^T dA_t}.
                \end{align*}
                
                We first bound the term $\spadesuit$. We can verify using Lemma~\ref{l:def_l} that $\nabla_z f(\l(z_t)) = 0$ when $z_t = 0$. For the case when $\lrn{z_t}_2 \neq 0$ we have,
                \begin{align*}
                \nabla f(\l(z_t)) = f'(\l(z_t)) q'(\lrn{z_t}_2) \frac{z_t}{\lrn{z_t}_2},
                \end{align*}
                where $q(\cdot)$ is the function used to define $\l$ (see Lemma~\ref{l:def_mr}).
                Thus
                \begin{align*}
                \spadesuit
                &= \lin{\nabla f(\l(z_t)), - \nabla_t - \nablat_t}\\
                &= f'(\l(z_t)) \cdot q'(z_t) \cdot \lin{\frac{z_t}{\lrn{z_t}_2}, - \nabla_t - \nablat_t}\\
                &\overset{(i)}{\leq} f'(\l(z_t))\cdot q'(z_t) \lin{\frac{z_t}{\lrn{z_t}_2}, - \nabla_t} + \lrn{\nablat_t}_2\\
                &\overset{(ii)}{\leq} \ind{\lrn{z_t}_2 \in [0,\beta]} \cdot L \beta
                + \ind{\lrn{z_t}_2 \in [R, \infty]} \cdot \lrp{- m \lrn{z_t}_2}\\
                &\qquad + \ind{\lrn{z_t}_2 \in [\beta, R)} \cdot f'(\l(z_t)) \cdot \lrp{L \lrn{z_t}_2} + \lrn{\nablat_t}_2,
                \end{align*}
                where $(i)$ is by the Cauchy-Schwarz inequality, along with the fact that $\lrabs{f'(r)} \leq 1$ (see (F2) of Lemma~\ref{lem:fpropertiesall}), and Lemma \ref{l:def_mr}.3. The inequality in $(ii)$ can be verified by considering three disjoint events. When $\lrn{z_t}_2   \in [0,\beta]$, the bound follows by Cauchy-Schwarz, (F2) of Lemma \ref{lem:fpropertiesall}, combined with Lemma~\ref{l:def_mr}.3. While when $\lrn{z_t}_2   \in [R,\infty]$ the bound follows from Assumption~\ref{ass:strongconvexity}. When $\lrn{z_t}_2 \in [\beta, R]$, we bound the term using Cauchy-Schwarz, Assumption~\ref{ass:smoothness}, and Lemma~\ref{l:def_mr}.3.
                
                Next, we consider the other term $\heartsuit = \frac{1}{2} \tr\lrp{\nabla^2_z f(\l(z_t)) \lrp{8 \gamma_t \gamma_t^T + 2 \bar{\gamma}_t \bar{\gamma}_t^T }}$. We can verify using Lemma~\ref{l:def_l}.4 that $\nabla^2 f(\l(z_t)) = 0$ when $\lrn{z_t}_2 = 0$. Thus for $\lrn{z_t}_2=0$, the following holds:
                \begin{align*}
                \heartsuit = 0 = \ind{\lrn{z_t}_2 \in [\beta, R]} 4 f''(\l(z_t)).
                \end{align*}
                Alternatively, when $\lrn{z_t}_2 \neq 0$,
                \begin{align*}
                \nabla^2 f(\l(z_t)) 
                =& f''(\l(z_t)) q'(\lrn{z_t}_2)^2 \frac{z_t z_t^T}{\lrn{z_t}_2^2}
                 + f'(\l(z_t)) q'(\lrn{z_t}_2) \frac{1}{\lrn{z_t}_2} \lrp{I - \frac{z_t z_t^T}{\lrn{z_t}_2^2}}\\
                &\qquad\qquad\qquad + f'(\l(z_t)) q''(\lrn{z_t}_2) \frac{z_t z_t^T}{\lrn{z_t}_2^2}.
                \end{align*}
                
                Exapanding using the definition of $\heartsuit$,
                \begin{align*}
                \heartsuit
                =& \frac{1}{2} \tr\lrp{\nabla^2_z f(\l(z_t)) 8 \gamma_t \gamma_t^T + 2 \bar{\gamma}_t \bar{\gamma}_t^T }\\
                \overset{(i)}{=}& \underbrace{\frac{1}{2} \tr\lrp{f''(\l(z_t)) \cdot q'(\lrn{z_t}_2)^2 \cdot \frac{z_t z_t^T}{\lrn{z_t}_2^2} \lrp{8\gamma_t \gamma_t^T + 2\bar{\gamma}_t \bar{\gamma}_t^T}}}_{=:\heartsuit_1}\\
                &\quad + \underbrace{\frac{1}{2} \tr\lrp{f'(\l(z_t))\cdot q'(\lrn{z_t}_2) \cdot  \frac{1}{\lrn{z_t}_2} \lrp{I - \frac{z_t z_t^T}{\lrn{z_t}_2^2}} \lrp{8\gamma_t \gamma_t^T + 2\bar{\gamma}_t \bar{\gamma}_t^T}}}_{=:\heartsuit_2}\\
                &\quad + \underbrace{\frac{1}{2} \tr\lrp{f'(\l(z_t)) q''(z_t) \frac{z_t z_t^T}{\lrn{z_t}_2^2} \lrp{8\gamma_t \gamma_t^T + 2\bar{\gamma}_t \bar{\gamma}_t^T}}}_{=:\heartsuit_3},
                \end{align*}
                where $(i)$ is by the expression for $\nabla^2 f(\l(z_t))$ above. 
                
                Before proceeding, we verify by definition of $\gamma_t$ and $\bar{\gamma}_t$ in Eq.~\eqref{d:gamma} that
                \begin{align*}
                \tr\lrp{\frac{z_t z_t^T}{\lrn{z_t}_2^2} \lrp{8\gamma_t \gamma_t^T + 2\bar{\gamma}_t \bar{\gamma}_t^T}} = 8\lrn{\gamma_t}_2^2 + 2\lrn{\bar{\gamma}_t}_2^2.
                \numberthis \label{e:t:qoinfdm}
                \end{align*}

                First we simplify $\heartsuit_1$:
                \begin{align*}
                \heartsuit_1
                =& \frac{1}{2} f''(\l(z_t)) \cdot q'(\lrn{z_t}_2)^2 \cdot \lrp{8\lrn{\gamma_t}_2^2 + 2\lrn{\bar{\gamma}_t}_2^2}\\
                \overset{(i)}{\leq}& \ind{\lrn{z_t}_2 \in [\beta, R]} \lrp{f''(\l(z_t)) \cdot \lrp{4\lrn{\gamma_t}_2^2 + \lrn{\bar{\gamma}_t}_2^2}}\\
                \overset{(ii)}{=}& \ind{\lrn{z_t}_2 \in [\beta, R]} 4 f''(\l(z_t)),
                \end{align*}
                where the inequality $(i)$ is because $f''(r) \geq 0$ for all $r > 0$ (by Lemma \ref{lem:fpropertiesall}.(F5)), $q'(r) \geq 0$ for all $r$ (by Lemma \ref{l:def_mr}.3) and $q'(r) = 1$ for all $r\geq \beta/2$ (Lemma \ref{l:def_mr}.3). The equality in $(ii)$ is because $\gamma_t = 1$ for $\lrn{z_t}_2 \geq \beta$ (by its definition in Eq.~\eqref{d:o:gamma}).
                
                Next, using Eq.~\eqref{e:t:qoinfdm}, we can immediately verify that $\heartsuit_2 = 0$. 
                
                Finally, we focus on $\heartsuit_3$,
                \begin{align*}
                \heartsuit_3
                = \frac{1}{2} f'(\l(z_t)) q''(z_t) \lrp{8\lrn{\gamma_t}_2^2 + 2\lrn{\bar{\gamma}_t}_2^2} = 0,
                \end{align*}
                where we use the fact that $q''(\lrn{z_t}_2) = 0$ if $\lrn{z_t}_2 \geq \beta/2$ (by Lemma \ref{l:def_mr}.4) and $\gamma_t = \bar{\gamma}_t = 0$ if $\lrn{z_t}_2 \leq \beta/2$ (by its definition in Eq.~\eqref{d:o:gamma}).
                
                Putting together the bounds on $\heartsuit_1$, $\heartsuit_2$ and $\heartsuit_3$, we can upper bound $\heartsuit$ as
                \begin{align*}
                \heartsuit \leq \ind{\lrn{z_t}_2 \in [\beta, R]} 4 f''(\l(z_t)).
                \end{align*}
                
                Combining the upper bounds on $\spadesuit$ and $\heartsuit$,
                \begin{align*}
                \spadesuit + \heartsuit
                \leq& \underbrace{\ind{\lrn{z_t}_2 \in [\beta, R]} \lrp{L \lrn{z_t}_2 f'(\l(z_t)) + 4 f''(\l(z_t))}}_{=:\clubsuit}\\
                &\quad + \ind{\lrn{z_t}_2 \in [R, \infty]} \cdot \lrp{- m \lrn{z_t}_2} + \lrn{\nablat_t}_2 + L \beta.
                \end{align*}
                
                Let us now focus on $\clubsuit$.  By Lemma \ref{lem:fpropertiesall}, 
                \begin{align*}
                \clubsuit =& \ind{\lrn{z_t}_2 \in [\beta, R]} \cdot \lrp{L \lrn{z_t}_2 f'(\l(z_t)) + 4 f''(\l(z_t))}\\
                \overset{(i)}{\leq}& \ind{\lrn{z_t}_2 \in [\beta, R]} \cdot  \lrp{L \cdot \l(z_t) f'(\l(z_t)) + 4 f''(\l(z_t)) + L \beta/3}\\
                \overset{(ii)}{\leq}& \ind{\lrn{z_t}_2 \in [\beta, R]} \cdot  \lrp{- \Co f(\l(z_t))}  + L \beta/3\\
                \overset{(iii)}{\leq}& \ind{\lrn{z_t}_2 \in [0, R]} \lrp{- \Co f(\l(z_t))}  + \lrp{L + 8\Co} \beta\\
                \overset{(iv)}{\leq}& \ind{\lrn{z_t}_2 \in [0, R]} \lrp{- \Co f(\l(z_t))}  + 10 L \beta,
                \end{align*}
                where $(i)$ is because $\lrabs{\lrn{z_t}_2 - \l(z_t)} \leq \beta/3$ (by Lemma \ref{l:def_l}.1) and because $\lrabs{f'(r)} \leq 1$ for all $r> 0$ (by Lemma \ref{lem:fpropertiesall}.(F2)). The inequality in $(ii)$ is by Lemma \ref{lem:fpropertiesall} (F4), our definition of $\Co$ in \eqref{d:co}, and the fact that $\lrn{z_t}_2 \in [\beta, R]$ implies $\l(z_t) \leq R$ (Lemma \ref{l:def_l}.1). Inequality $(iii)$ is again by Lemma \ref{l:def_l}.1 and Lemma \ref{lem:fpropertiesall} (F3). Finally, $(iv)$ is by \eqref{d:co} and $m \leq L$, a consequence of Assumptions \ref{ass:smoothness} and \ref{ass:strongconvexity}. 
                
                Thus,
                \begin{align*}
                \spadesuit+ \heartsuit
                \leq& \ind{\lrn{z_t}_2 \in [0, R]} \lrp{- \Co f(\l(z_t))} + \ind{\lrn{z_t}_2 \in [R, \infty]} \cdot \lrp{- m \lrn{z_t}_2} + 11 L\beta + \lrn{\nablat_t}_2\\
                \leq& - \Co f(\l(z_t)) + 12 L\beta + \lrn{\nablat_t}_2,
                \end{align*}
                where the second line is by Lemma \ref{l:def_l}.1 and \ref{lem:fpropertiesall}.(F3), and by $m \leq L$.
                
                Putting this together with the expression for $df(\l(z_t))$,
                \begin{align*}
                d f(\l(z_t)) 
                \leq& \lrp{- \Co f(\l(z_t)) + 12 L\beta + \lrn{\nablat_t}_2} dt
                 + \lin{\nabla_z f(\l(z_t)), 2\sqrt{2} \gamma_t \gamma_t^T dB_t + \sqrt{2} \bar{\gamma}_t \bar{\gamma}_t^T dA_t}\\
                \leq& \lrp{- \Co f(\l(z_t)) + 12 L\beta + L\lrn{x_t - x_{\step{t}{\delta}\delta}}_2} dt\\ &\qquad \qquad + \lin{\nabla_z f(\l(z_t)), 2\sqrt{2} \gamma_t \gamma_t^T dB_t + \sqrt{2} \bar{\gamma}_t \bar{\gamma}_t^T dA_t}.
                \end{align*}
                The second inequality uses the definition of $\nablat_t$ in Eq.~\eqref{d:o:nablat} and Assumption \ref{ass:smoothness}.
                
                \textbf{Step 2:} If we consider the evolution of the Lyapunov function $\L_t$ (defined in Eq.~\eqref{d:o:L}), we can verify that
                \begin{align*}
                d \L_t 
                =& d \lrp{f(\l(z_t)) - \phi_t - \xi_t}\\
                \overset{(i)}{\leq}& -\Co \lrp{f(\l(z_t)) - \phi_t - \xi_t} dt + 12 L\beta dt\\
                =& -\Co \L_t dt + 12 L\beta dt,
                \end{align*}
                where the simplification in inequality $(i)$ can be verified by taking time derivatives of stochastic processes $\phi_t$ and $\xi_t$ defined in Eq.~\eqref{d:o:phi} and Eq.~\eqref{d:o:xi}.
                
                Applying Gr\"onwall's inequality,
                \begin{align*}
                \L_t 
                \leq& e^{-\Co t} \L_0 + \int_0^t e^{-\Co (t-s)} 12 L \beta ds \leq  e^{-\Co t} \L_0 + \frac{12 L \beta}{\Co}.
                \end{align*}
                Using the definition of $\L_t$ in Eq.~\eqref{d:o:L} we get,
                \begin{align*}
                f(\l(z_t)) \leq e^{-\Co t} f(\l(z_0)) + \xi_t + \phi_t.
                \end{align*}
                Taking expectations with respect to the Brownian motion yields:
                \begin{align*}
                \E{f(\l(z_t))} \leq e^{-\Co t}\E{f(\l(z_0))} + \E{\xi_t} + \E{\phi_t}.
                \numberthis \label{e:t:rlksd}
                \end{align*}
                
                By the definition of $\phi_t$ in Eq.~\eqref{d:o:phi}, we verify that $\E{\phi_t}=0$, and by definition of $\xi_t$ in Eq.~\eqref{d:o:xi},
                \begin{align*}
                 \E{\xi_t} 
                =& \int_0^t e^{-\Co (t - s)} \E{\lrn{x_t - x_{\step{t}{\delta}\delta}}_2}ds\\
                \leq& \int_0^t e^{-\Co (t - s)} \E{\lrn{\lrp{s-\step{s}{\delta} \delta} \nabla U(x_{\step{s}{\delta}\delta}) + \int_{\step{s}{\delta}\delta}^s dB_r}_2}ds\\
                \leq& \int_0^t e^{-\Co (t - s)} \lrp{\E{\delta L \lrn{x_{\step{s}{\delta}\delta}}_2} + \sqrt{\delta d}} ds\\
                \leq& \int_0^t e^{-\Co (t - s)} \lrp{2\delta L \sqrt{R^2 + d/m} + \sqrt{\delta d}} ds\\
                \leq& \frac{2\delta L \sqrt{R^2 + d/m} + \sqrt{\delta d}}{\Co}
                \end{align*}
                
                We can also bound the initial value of $\E{f(\l(z_0))}$ as follows:
                \begin{align*}
                \E{f(\l(z_0))} \overset{(i)}{=} \E{f(\l(y_0))} \overset{(ii)}{\leq}\E{\l(y_0)}
                \overset{(iii)}{\leq} \E{\lrn{y_0}_2} + \beta/3 \overset{(iv)}{\leq} \sqrt{R^2 + \frac{d}{m}} + \beta/3,
                \end{align*}
                where $(i)$ is because $x^(0) = 0$ in Eq.~\eqref{d:o:xt}, $(ii)$ is by Lemma \ref{lem:fpropertiesall}.(F3), $(iii)$ is by Lemma \ref{l:def_l}.1, and finally $(iv)$ is by Lemma \ref{l:energy_bound_pstar}.
                
                Let $n$ be the number of time steps, so that $t=n\delta$. Substituting into the inequality in Eq.~\eqref{e:t:rlksd}, we get
                \begin{align*}
                \E{f(\l(z_{n\delta}))} \leq e^{-\Co (n\delta)} \lrp{32 \sqrt{R^2 + \frac{d}{m}} + \beta/3} + \frac{2\delta L \sqrt{R^2 + d/m} + \delta d}{\Co} + \frac{12L\beta}{\Co}.
                \end{align*}
                
                \textbf{Step 3:}
                We translate our bound on $\E{f(\l(z_{n\delta}))}$ to a bound on $\E{\lrn{z_{n\delta}}_2}$, which implies a bound in 1-Wasserstein distance. By Lemma~\ref{lem:fpropertiesall}(F3),
                \begin{align*}
                &\E{\lrn{z_{n\delta}}_2} \\
                & \qquad \leq 2 e^{L (R + \beta)^2/2}\lrp{e^{-\Co (n\delta)} \lrp{32 \sqrt{R^2 + \frac{d}{m}} + \beta/3} + \frac{2\delta L \sqrt{R^2 + d/m} + \delta d}{\Co} + \frac{12L\beta}{\Co}}\\
                &\qquad \leq 4 e^{L R^2/2}\lrp{e^{-\Co (n\delta)} \lrp{32 \sqrt{R^2 + \frac{d}{m}}} + \frac{2\delta L \sqrt{R^2 + d/m} + \delta d}{\Co}},
                \end{align*}
                where for the second inequality, it suffices to let $\beta = \delta d/6$
                
                For a given $\epsilon$, the first term is less than $\epsilon/2$ if 
                \begin{align*}
                n\delta 
                \geq& 10 \lrp{\log \lrp{\frac{R^2 + d/m}{\epsilon}} + LR^2} \cdot \frac{1}{\Co}.
                \end{align*}
                The second term is less than $\epsilon/2$ if 
                \begin{align*}
                \delta 
                \leq& \frac{1}{10} e^{-LR^2/2}\min \lrbb{\frac{\epsilon}{\sqrt{R^2 + d/m}}, \frac{\epsilon^2 \Co}{d}}.
                \end{align*}
                By the definition of $\Co$ in Eq.~\eqref{d:co},
                \begin{align*}
                \Co 
                \leq& \frac{1}{8} \min\lrbb{\frac{\exp\lrp{-LR^2/2}}{R^2}, m}= \frac{\exp \lrp{LR^2/2}}{8 R^2},
                \end{align*}
                where the equality is by our assumption on the strong convexity parameter $m$ in the theorem statement. Recall that we also assume that $\epsilon \leq \frac{d R^2}{\sqrt{d/m + R^2}}$. Thus we can verify that $$\min \lrbb{\frac{\epsilon}{\sqrt{R^2 + d/m}}, \frac{\epsilon^2 \Co}{d}} = \frac{\epsilon^2 \Co}{d}.$$
                
                Putting everything together, we obtain a guarantee that $\E{\lrn{z_t}_2} \leq \epsilon$ if 
                \begin{align*}
                \delta = \frac{\epsilon^2 \exp\lrp{-LR^2}}{2^{10}R^2 d} ,
                \end{align*}
                and
                \begin{align*}
                n \geq 2^{18} \log \lrp{\frac{R^2 + d/m}{\epsilon}} \cdot R^4 \cdot \exp\lrp{\frac{3 LR^2}{2}} \cdot \frac{d}{\epsilon^2},
                \end{align*}
                as prescribed by the theorem statement.

        \end{subsection}

    \end{section}

    \newpage
    
    \begin{section}{Proofs for Underadmped Langevin Monte Carlo}\label{appendix:underdamped}
        \begin{subsection}{Overview}
            The main idea behind the proof is to show that $\L_t$ contracts with probability one by a factor of $e^{-\Cm \nu}$, going from $t=(k-1)\nu$ to $t=k\nu$. The result can be found in Lemma \ref{l:combining_4_cases} in Section \ref{s:putting_together}. The proof considers four cases:
            
            \begin{enumerate}
                \item $\mu_{k-1}=1,\mu_k=1$. In Lemma \ref{l:mu11} in Section \ref{s:putting_together}, we show that $\L_{k\nu} \leq e^{-\Cm\nu}\L_{(k-1)\nu}$. The proof of this result in turn uses Lemma \ref{l:muk=1} in Section \ref{s:reflection}, which shows that $\L_t$ contracts at a rate of $-\Cm$ over the interval $t\in[(k-1)\nu,k\nu]$.
                
                \item $\mu_{k-1}=1,\mu_k=0$. In Lemma \ref{l:mu10} in Section \ref{s:putting_together}, we show that $\L_{k\nu} \leq e^{-\Cm\nu}\L_{(k-1)\nu}$. The proof of this result is almost identical to the preceding case $\mu_{k-1}=1,\mu_k=1$. (In particular, $\L_t$ undergoes no jump in value at $t=k\nu$, in spite in the change in value from $\mu_{k-1}=1$ to $\mu_k=0$. See proof for details.)
                
                \item $\mu_{k-1}=0,\mu_k=0$. In Lemma \ref{l:mu00} in Section \ref{s:putting_together}, we show that $\L_{k\nu} \leq e^{-\Cm\nu}\L_{(k-1)\nu}$. The proof of this result is mainly based on the definition of $\L_t$.
                
                \item $\mu_{k-1}=0,\mu_k=1$. In Lemma \ref{l:mu01} in Section \ref{s:putting_together}, we show that $\L_{k\nu} \leq e^{-\Cm\nu}\L_{(k-1)\nu}$. This case is somewhat tricky, as $\L_t$ undergoes a jump in value at $t=k\nu$. Specifically, $\L_t$ jumps from $ e^{-\Cm \Ts} \lrp{f(r_{\tau_{k-1}}) - \xi_{\tau_{k-1}}} - \lrp{\sigma_{k\nu} + \phi_{k\nu}}$ to $f(r_{k\nu}) - \xi_{k\nu} -  \lrp{\sigma_{k\nu} + \phi_{k\nu}}$. We prove that this jump is always negative (Lemma \ref{l:muk=0_norm_contraction_3}, Section \ref{s:synchronous}). The proof of Lemma \ref{l:muk=0_norm_contraction} in turn relies on a contraction result in Lemma \ref{l:muk=0_norm_contraction_2}.
            \end{enumerate}

            Having proven Lemma \ref{l:combining_4_cases}, we prove Theorem \ref{t:maintheoremunderdamped0} by applying Lemma \ref{l:combining_4_cases} recursively, and showing that $\E{\L_t}$ sandwiches the Wasserstein distance $W_1(p_t,p^*)$.
            
        \end{subsection}

        \begin{subsection}{Contraction under Reflection Coupling}\label{s:reflection}
            Our main result is stated as Lemma \ref{l:muk=1}. It shows that $\mu_k f(r_t)$ contracts at a rate of $\exp(-\Cm t)$, plus some discretization error terms.
            \begin{lemma}\label{l:muk=1}
                For any positive integer $k$, with probability one we have,
                \begin{align*}
                & \mu_k \cdot \lrp{f(r_{(k+1)\nu}) - \xi_{(k+1)\nu}}  - \lrp{\sigma_{(k+1)\nu} + \phi_{(k+1)\nu}}\\
                &\qquad \qquad \qquad \leq e^{-\Cm \nu} \lrp{ \mu_k \cdot \lrp{f(r_{k\nu}) - \xi_{k\nu}}  - \lrp{\sigma_{k\nu} + \phi_{k\nu}}} + 5\beta \nu.
                \end{align*}
            \end{lemma}
            
            \begin{proof}
                If $\mu_k=0$, both sides of the inequality are identically zero. To simplify notation, we leave out the factor of $\mu_k$ in subsequent expressions and assume that $\mu_k = 1$ unless otherwise stated. 
                
                For the rest of this proof, we will consider time $s\in[k\nu,(k+1)\nu)$ for some $k$.
                
                Let us first establish some useful derivatives of the function $f$:
                \begin{align*}
                \nabla_z f(r_s) 
                &= f'(r_s) \cdot (1+2\c) q'(\lrn{z_s}_2) \cdot \frac{z_s}{\lrn{z_s}_2}  + f'(r_s) \cdot \q'(z_s+w_s) \cdot \frac{z_s+w_s}{\lrn{z_s+w_s}_2},\\
                \nabla_w f(r_s) 
                &= f'(r_s) \cdot q'(\lrn{z_s + w_s}_2) \cdot \frac{z_s + w_s}{\lrn{z_s+w_s}_2},\\
                \nabla^2_w f(r_s) &= f'(r_s) \cdot q'(\lrn{z_s + w_s}_2) \cdot \frac{1}{\lrn{z_s + w_s}_2} \lrp{I - \frac{(z_s + w_s) (z_s + w_s)^T}{\lrn{z_s + w_s}_2^2}}\\
                &\qquad + f'(r_s) q''(\lrn{z_s + w_s}_2) \frac{(z_s + w_s) (z_s + w_s)^T}{\lrn{z_s + w_s}_2^2}\\
                &\qquad \qquad + f''(r_s) q'(\lrn{z_s + w_s}_2)^2 \frac{(z_s + w_s) (z_s + w_s)^T}{\lrn{z_s + w_s}_2^2}.
                \numberthis \label{e:t:nablaf_reflection}
                \end{align*}
                The derivatives follow from Lemma~\ref{l:def_l} and by the definition of $r_t$ in Eq.~\eqref{d:r}. From Lemma~\ref{l:def_l}.3, $\nabla^2_w f(\lrp{1+2\c}\l(z) + \l(z + w))$ exists everywhere and is continuous, \\with $\at{\nabla^2_w f(\lrp{1+2\c}\l(z) + \l(z + w))}{z+w=0} = 0$. Note that, we use the convention $0/0=0$.
                
                For any $s\in [k\nu, (k+1)\nu)$, we have:
                \begin{align*}
                 d \mu_k \cdot f(r_s)
                &\overset{(i)}{=} \mu_k \cdot \lin{\nabla_z f(r_s), d z_s} + \lin{\nabla_w f(r_s), d w_s} \\
                &\quad + \mu_k \cdot \frac{8\c}{L} \gamma_s^T \nabla^2_w f(r_s) \gamma_s ds + \frac{2\c}{L} \bar{\gamma}_s^T \nabla^2_w f(r_s) \bar{\gamma_s} ds\\
                &\overset{(ii)}{=}  \mu_k \cdot \underbrace{\lrp{\lin{\nabla_z f(r_s), w_s} + \lin{\nabla_w f(r_s), -2w_s - \frac{\c}{L}\nabla_s - \frac{\c}{L}\nablat_s}}}_{=:\spadesuit} ds\\
                &\quad + \mu_k \cdot \underbrace{\lrp{\lrp{\frac{8\c}{L} \gamma_s^T \nabla^2_w f(r_s) \gamma_s + \frac{2\c}{L} \bar{\gamma}_s^T \nabla^2_w f(r_s) \bar{\gamma_s}}}}_{=:\heartsuit} ds\\
                &\quad +  \mu_k \cdot \lin{\nabla_w f(r_s), \lrp{4\sqrt{\frac{\c}{L}}\gamma_t \gamma_t^T dB_t + 2 \sqrt{\frac{\c}{L}} \bar{\gamma}_t \bar{\gamma}_t^T dA_t}} ds,
                \numberthis \label{e:t:pdfvm}
                \end{align*}
                where $(i)$ follows from $\Ito$'s Lemma, and $(ii)$ follows from Eqs.~\eqref{d:xt} - \eqref{d:vt}, and the definition of $\nabla_t$ and $\nablat_t$ in Eq.~\eqref{d:nablat}.
                
                In the sequel, we upper bound the terms $\spadesuit, \heartsuit, \clubsuit$ separately. Before we proceed, we verify the following inequalities:
                \begin{align*}
                &q'(\lrn{z_s}_2) \lin{\frac{z_s}{\|z_s\|_2}, w_s} 
                 = q'(\lrn{z_s}_2)\lin{\frac{z_s}{\|z_s\|_2}, z_s + w_s -z_s}
                 \overset{(i)}{\leq} q'(\lrn{z_s}_2) \lrp{\|z_s+w_s\|_2 - \|z_s\|_2},
                 \end{align*}
                 where $(i)$ is by Cauchy-Schwarz, and:
                 \begin{align*}
                 &q'(\lrn{z_s + w_s}_2) \lin{\frac{z_s+w_s}{\|z_s+w_s\|_2}, - w_s - \frac{\c}{L}\nabla_s}\\ &\qquad \qquad = q'(\lrn{z_s + w_s}_2) \lin{\frac{z_s+w_s}{\|z_s+w_s\|_2}, - z_s - w_s + z_s - \frac{\c}{L}\nabla_s}\\
                & \qquad \qquad \overset{(i)}{\leq} q'(\lrn{z_s + w_s}_2)\lrp{-\|z_s+w_s\|_2 + \|z_s\|_2 + \lin{\frac{z_s+w_s}{\|z_s+w_s\|_2}, -\frac{\c}{L}\nabla_s}}\\
                &\qquad \qquad  \overset{(ii)}{\leq} q'(\lrn{z_s + w_s}_2)\lrp{-\|z_s+w_s\|_2 + (1+\c) \|z_s\|_2},
                \end{align*}
                where $(i)$ is again by Cauchy-Schwarz and $(ii)$ is by Cauchy-Schwarz combined with Assumption~\ref{ass:smoothness}. Finally:
                \begin{align*}
                & q'(z_s+w_s) \lin{\frac{z_s + w_s}{\lrn{z_s + w_s}_2}, - \frac{\c}{L} \nablat_s}   \leq q'(\lrn{z_s+w_s}_2) \frac{\c}{L} \lrn{\nablat_s}_2,
                \numberthis \label{e:t:reflection_derivative_intermediate}
                \end{align*}
                where the inequality above is by Cauchy-Schwarz along with the fact that $q'(r) \geq 0 $ for all $r$ from Lemma \ref{l:def_mr}.
                
                \textbf{Bounding $\spadesuit$:} From Eqs.~\eqref{e:t:pdfvm} and~\eqref{e:t:nablaf_reflection}:
                \begin{align*}
                \spadesuit
                &= (1+2\c) f'(r_s) q'(\lrn{z_s}_2) \lin{\frac{z_s}{\lrn{z_s}_2}, w_s} \\
                &\quad + f'(r_s) q'(\lrn{z_s+w_s}_2) \lin{\frac{z_s+w_s}{\lrn{z_s+w_s}_2}, w_s }\\
                &\quad + f'(r_s) q'(\lrn{z_s + w_s}_2) \lin{\frac{z_s + w_s}{\lrn{z_s+w_s}_2}, -2w_s - \frac{\c}{L}\nabla_s - \frac{\c}{L}\nablat_s} \\
                &= (1+2\c) f'(r_s) q'(\lrn{z_s}_2) \lin{\frac{z_s}{\lrn{z_s}_2}, w_s} \\
                &\quad + f'(r_s) q'(\lrn{z_s+w_s}_2) \lin{\frac{z_s+w_s}{\lrn{z_s+w_s}_2}, -w_s - \frac{\c}{L}\nabla_s - \frac{\c}{L}\nablat_s } =: \spadesuit_1.
                \numberthis \label{e:t:reflection_first_order_1}
                \end{align*}
        
                We again highlight the fact that $q'(\lrn{z}_2) \frac{z}{\lrn{z}_2}$ is defined for all $z$, particularly at $\lrn{z}_2=0$, as $q(r) = o(r^2)$ near zero (see Lemma \ref{l:def_mr}).
                
                Substituting the inequality in Eq.~\eqref{e:t:reflection_derivative_intermediate} into $\spadesuit_1$:
                \begin{align*}
    \spadesuit_1
                &= (1+2\c) f'(r_s) q'(\lrn{z_s}_2) \lin{\frac{z_s}{\lrn{z_s}_2}, w_s} \\
                &\quad + f'(r_s) q'(\lrn{z_s+w_s}_2) \lin{\frac{z_s+w_s}{\lrn{z_s+w_s}_2}, -w_s - \frac{\c}{L}\nabla_s - \frac{\c}{L}\nablat_s }\\
                & \leq (1+2\c) f'(r_s) q'(\lrn{z_s}_2)\lrp{\|z_s+w_s\|_2 - \|z_s\|_2}\\
                &\quad + f'(r_s) q'(\lrn{z_s + w_s}_2)\lrp{-\|z_s+w_s\|_2 + (1+\c)\|z_s\|_2} + \frac{\c}{L} \lrn{\nablat_s}_2,
                \end{align*}
                where the inequality uses Cauchy-Schwarz and (F2) of Lemma~\ref{lem:fpropertiesall}.
                
                Now consider a few cases. We will use the expression for $q'(r)$ from Eq.~\eqref{l:def_mr} a number of times:
                \begin{enumerate}
                    \item If $\lrn{z_s}_2 \in [\beta, \infty), \lrn{z_s + w_s}_2 \in [\beta,\infty)$, then $q'(\lrn{z_s}_2) = q'(\lrn{z_s + w_s}_2) = 1$, so that 
                    \begin{align*}
                    \spadesuit_1
                    &\leq f'(r_s) \lrp{\lrn{z_s + w_s}_2 - \lrn{z_s}_2 - \lrn{z_s + w_s}_2 + (1+\c) \lrn{z_s}_2}  + \frac{\c}{L} \lrn{\nablat_s}_2\\
                    &= f'(r_s) \lrp{\c \lrn{z_s}_2} + \frac{\c}{L} \lrn{\nablat_s}_2\\
                    &\leq 2\c f'(r_s) r_s + \beta + \frac{\c}{L} \lrn{\nablat_s}_2,
                    \end{align*}
                    where we use the definition of $r_t$ defined in Eq.~\eqref{d:r} and Lemma~\ref{l:def_l}.1.
                    \item If $\lrn{z_s}_2 \in [0, \beta), \lrn{z_s + w_s}_2 \in [\beta,\infty)$, then $q'(\lrn{z_s}_2)\in [0,1]$ and $q'(\lrn{z_s + w_s}_2) = 1$, so that
                    \begin{align*}
                    \spadesuit_1
                    &\overset{(i)}{\le} f'(r_s) \lrp{ (1+2\c) q'(\lrn{z_s}_2)\lrn{w_s}_2 - \lrn{z_s + w_s}_2 + (1+\c) \lrn{z_s}_2 }  + \frac{\c}{L} \lrn{\nablat_s}_2\\
                    &\overset{(ii)}{\le} f'(r_s) \lrp{2\c \lrn{w_s}_2 + 3 \lrn{z_s}_2}  + \frac{\c}{L} \lrn{\nablat_s}_2\\
                    &\overset{(iii)}{\le} f'(r_s) \lrp{2\c \lrn{w_s}_2 + 3\beta} + \frac{\c}{L} \lrn{\nablat_s}_2\\
                    &\overset{(iv)}{\le} 2\c f'(r_s) r_s + 5\beta + \frac{\c}{L} \lrn{\nablat_s}_2,
                    \end{align*}
                    where (i) uses $\lrn{z_s+w_s}_2 - \lrn{z_s}_2\leq \lrn{w_s}_2$, $(ii)$ uses $\lrn{w_s}_2 - \lrn{z_s + w_s}_2 \leq \lrn{z)s}_2$, $(iii)$ uses our upper bound in $\lrn{z_s}_2$ and $(iv)$ uses the definition of $r_t$ in Eq.~\eqref{d:r} and Lemma~\ref{l:def_l}.1.
                    \item If $\lrn{z_s}_2 \in [\beta, \infty), \lrn{z_s + w_s}_2 \in [0,\beta)$, then $q'(\lrn{z_s}_2) = 1$ and $q'(\lrn{z_s + w_s}_2) \in [0,1]$, so that
                    \begin{align*}
                \spadesuit_1
                &\overset{(i)}{\le} f'(r_s)\lrp{(1+2\c)  \lrp{\|z_s+w_s\|_2 - \|z_s\|_2} -\|z_s+w_s\|_2 + (1+\c)\|z_s\|_2} + \frac{\c}{L} \lrn{\nablat_s}_2\\
                    &= f'(r_s) \lrp{2\c \lrn{z_s + w_s}_2 - \c \lrn{z_s}_2} + \frac{\c}{L} \lrn{\nablat_s}_2\\
         &\overset{(ii)}{\le} f'(r_s) \lrp{3 \c \lrn{z_s + w_s}_2 - \frac{\c}{2} r_s + 2\beta} + \frac{\c}{L} \lrn{\nablat_s}_2\\
         &\le f'(r_s) \lrp{- \frac{\c}{2}r_s} + 5 \beta + \frac{\c}{L} \lrn{\nablat_s}_2,
                    \end{align*}
                    where $(i)$ uses our expression for $q'(\cdot)$, and $(ii)$ uses the expression for $r_t$ in Eq.~\eqref{d:r}, the fact that $\c\leq 1/1000$ and Lemma~\ref{l:def_l}.1.
                    
                    \item Finally, if $\lrn{z_s}_2 \in [0, \beta), \lrn{z_s + w_s}_2 \in [0,\beta)$, then $q'(\lrn{z_s}_2) \in [0,1]$ and $q'(\lrn{z_s + w_s}_2) \in [0,1]$, so that
                    \begin{align*}
                    \spadesuit_1
                    \leq& f'(r_s) \lrp{3\beta}  + \frac{\c}{L} \lrn{\nablat_s}_2\le f'(r_s) \lrp{-\frac{\c}{2}r_s} + 5\beta  + \frac{\c}{L} \lrn{\nablat_s}_2,
                    \end{align*}
                    where we again use the expression for $r_s$ in Eq.~\eqref{d:r} and Lemma \ref{l:def_l}.1.
                \end{enumerate}
                Combining the four cases above we find that, 
                \begin{align*}
                \spadesuit
                \leq\spadesuit_1
                \leq& \ind{\lrn{z_s+w_s}_2 \in [0,\beta)} \cdot \lrp{f'(r_s) \lrp{-\frac{\c}{2}r_s} + 4\beta  + \frac{\c}{L} \lrn{\nablat_s}_2}\\
                &\quad + \ind{\lrn{z_s+w_s}_2 \in [\beta, \infty)}\cdot \lrp{2\c f'(r_s) r_s + 5\beta + \frac{\c}{L}\lrn{\nablat_s}_2},
                \numberthis \label{e:t:wsfdfje:1}
                \end{align*}
                where we use Lemma~\ref{lem:fpropertiesall}.(F2), Lemma~\ref{l:def_l}.1 and Eq.~\eqref{d:mut}.
                
                \textbf{Bounding $\heartsuit$:}
                \begin{align*}
                \heartsuit &\overset{(i)}{=} \lrp{\frac{8\c}{L} \gamma_s^T \nabla^2_w f(r_s) \gamma_s + \frac{2\c}{L} \bar{\gamma}_s^T \nabla^2_w f(r_s) \bar{\gamma_s}}\\
                &\overset{(ii)}{=} \frac{8\c}{L} \cdot  \gamma_s^T \lrp{f'(r_s) \cdot q'(\lrn{z_s + w_s}_2) \cdot \frac{1}{\lrn{z_s + w_s}_2} \lrp{I - \frac{(z_s + w_s) (z_s + w_s)^T}{\lrn{z_s + w_s}_2^2}}} \gamma_s\\
                &\quad + \frac{8\c}{L} \cdot   {\gamma}_s^T \lrp{f'(r_s) q''(\lrn{z_s + w_s}_2) \frac{(z_s + w_s) (z_s + w_s)^T}{\lrn{z_s + w_s}_2^2}} {\gamma}_s\\
                &\quad + \frac{8\c}{L} \cdot   {\gamma}_s^T \lrp{f''(r_s) q'(\lrn{z_s + w_s}_2)^2 \frac{(z_s + w_s) (z_s + w_s)^T}{\lrn{z_s + w_s}_2^2}} \gamma_s\\
                &\quad + \frac{2\c}{L} \cdot  \bar{\gamma}_s^T \lrp{f'(r_s) \cdot q'(\lrn{z_s + w_s}_2) \cdot \frac{1}{\lrn{z_s + w_s}_2} \lrp{I - \frac{(z_s + w_s) (z_s + w_s)^T}{\lrn{z_s + w_s}_2^2}}} \bar{\gamma}_s\\
                &\quad + \frac{2\c}{L} \cdot   \bar{\gamma}_s^T \lrp{f'(r_s) q''(\lrn{z_s + w_s}_2) \frac{(z_s + w_s) (z_s + w_s)^T}{\lrn{z_s + w_s}_2^2}} \bar{\gamma}_s\\
                &\quad + \frac{2\c}{L} \cdot   \bar{\gamma}_s^T \lrp{f''(r_s) q'(\lrn{z_s + w_s}_2)^2 \frac{(z_s + w_s) (z_s + w_s)^T}{\lrn{z_s + w_s}_2^2}} \bar{\gamma}_s\\
                &\overset{(iii)}{=} \frac{8\c}{L} \cdot   \lrp{ \lrp{f''(r_s) q'(\lrn{z_s + w_s}_2)^2 + f'(r_s) q''(\lrn{z_s + w_s}_2)} }\cdot \lrn{\gamma_s}_2^2\\
                &\quad + \frac{2\c}{L} \cdot   \lrp{f''(r_s) q'(\lrn{z_s + w_s}_2)^2 + f'(r_s) q''(\lrn{z_s + w_s}_2)}\cdot \lrn{\bar{\gamma}_s}_2^2,
                \end{align*}
                where $(i)$ is by Eq.~\eqref{e:t:nablaf_reflection}, $(ii)$ is by Lemma~\ref{e:t:nablaf_reflection} and $(iii)$ is because $\lin{\gamma_s, \frac{z_s + w_s}{\lrn{z_s + w_s}_2}} = \lrn{\gamma_s}_2$ and $\lin{\bar{\gamma}_s, \frac{z_s + w_s}{\lrn{z_s + w_s}_2}} = \lrn{\bar{\gamma}_s}_2$ (see Eq.~\eqref{d:gamma}).
                
                From Lemma \ref{l:def_mr}.4, $q''(\lrn{z_s + w_s}_2) = 0$ for $\lrn{z_s + w_s}_2 \geq \beta/2$ and from Eq.~\eqref{d:gamma}, $\gamma_s = \bar{\gamma}_s = 0$ for $\lrn{z_s + w_s}_2 \leq \beta/2$. Thus the above simplifies to 
                \begin{align*}
                \heartsuit
                & \overset{(i)}{\leq} \frac{8\c}{L} \cdot   \lrp{f''(r_s) q'(\lrn{z_s + w_s}_2)^2 }\cdot \lrn{\gamma_s}_2^2+ \frac{2\c}{L} \cdot   \lrp{f''(r_s) q'(\lrn{z_s + w_s}_2)^2} \cdot \lrn{\bar{\gamma}_s}_2^2\\
                & \overset{(ii)}{\leq} \frac{8\c}{L} \cdot   \lrp{f''(r_s) q'(\lrn{z_s + w_s}_2)^2 }\cdot \lrn{\gamma_s}_2^2\\
                &\le \ind{\lrn{z_s + w_s}_2 \geq \beta} \cdot \frac{8\c}{L} \cdot   f''(r_s),
                \numberthis \label{e:t:wsfdfje:2}
                \end{align*}
                where $(i)$ is by Lemma~\ref{lem:fpropertiesall} (F5), which implies that $\frac{2\c}{L} \cdot   \lrp{f''(r_s) q'(\lrn{z_s + w_s}_2)^2} \cdot \lrn{\bar{\gamma}_s}_2^2 \leq 0$. The inequality in $(ii)$ is because $f''(r)\leq 0$ for all $r$ (Lemma~\ref{lem:fpropertiesall}.(F5)), along with the facts that $\ind{\lrn{z_s + w_s}_2 \geq \beta} \cdot q'(\lrn{z_s + w_s}_2) = \ind{\lrn{z_s + w_s}_2 \geq \beta}$ (by Lemma~\ref{l:def_mr}.3), and $\ind{r\geq \beta} q'(r)^2 = \ind{r\geq \beta}$  (by Eq.~\eqref{d:gamma}).
                
                Combining our upper bounds on $\spadesuit$ and $\heartsuit$ from Eq.~\eqref{e:t:wsfdfje:1} and Eq.~\eqref{e:t:wsfdfje:2},
                \begin{align*}
                \spadesuit+\heartsuit
                &\le \ind{\lrn{z_s+w_s}_2 < \beta} \cdot \lrp{f'(r_s) \lrp{-\frac{\c}{2}r_s} + 5\beta  + \frac{\c}{L} \lrn{\nablat_s}_2}\\
                &\quad + \ind{\lrn{z_s+w_s}_2 \geq \beta}\cdot \lrp{2\c f'(r_s) r_s + 5\beta + \frac{\c}{L}\lrn{\nablat_s}_2}\\
                &\quad + \ind{\lrn{z_s + w_s}_2 \geq \beta} \cdot \frac{8\c}{L} \cdot   f''(r_s)\\
                &\overset{(i)}{=} \ind{\lrn{z_s + w_s}_2 \geq \beta, r_s \leq \sqrt{12}R} \cdot \lrp{\frac{8\c}{L} f''(r_s) + 2\c f'(r_s)\cdot r_s}\\
                &\quad + \ind{\lrn{z_s + w_s}_2 \geq \beta, r_s > \sqrt{12}R} \cdot \lrp{2\c f'(r_s) r_s}\\
                &\quad + \ind{\lrn{z_s+w_s}_2 < \beta} \cdot \lrp{f'(r_s) \lrp{-\frac{\c}{2}r_s}}\\
                &\quad + 5\beta + \frac{\c}{L}\lrn{\nablat_s}_2\\
                &\overset{(ii)}{=} \ind{\lrn{z_s + w_s}_2 \geq \beta, r_s \leq \sqrt{12}R} \cdot \lrp{\frac{8\c}{L}\lrp{f''(r_s) + \frac{L}{4} f'(r_s)\cdot r_s}}\\
                &\quad + \ind{\lrn{z_s + w_s}_2 \geq \beta, r_s > \sqrt{12}R} \cdot \lrp{2\c f'(r_s) r_s}\\
                &\quad + \ind{\lrn{z_s+w_s}_2 < \beta} \cdot \lrp{f'(r_s) \lrp{-\frac{\c}{2}r_s}}\\
                &\quad + 5\beta + \frac{\c}{L}\lrn{\nablat_s}_2,
                \end{align*}
                where $(i)$ and $(ii)$ follow from algebraic manipulations. Continuing forward we find that,
                \begin{align*}
                \spadesuit+\heartsuit&\overset{(i)}{\leq} \ind{\lrn{z_s + w_s}_2 \geq \beta, r_s \leq \sqrt{12}R} \cdot \lrp{-\frac{8\c}{L}  \cdot \frac{e^{-6LR^2}}{48R^2} f(r_s)}\\
                &\quad + \ind{\lrn{z_s + w_s}_2 \geq \beta, r_s > \sqrt{12}R} \cdot \lrp{2\c f'(r_s) r_s}\\
                &\quad + \ind{\lrn{z_s+w_s}_2 < \beta} \cdot \lrp{-\frac{\c e^{-6LR^2}}{4} f(r_s)}\\
                &\quad + 5\beta + \frac{\c}{L}\lrn{\nablat_s}_2\\
                &\overset{(ii)}{\leq} \ind{\lrn{z_s + w_s}_2 \geq \beta, r_s \leq \sqrt{12}R} \cdot \lrp{- \Cm f(r_s)}\\
                &\quad + \ind{\lrn{z_s + w_s}_2 < \beta} \cdot \lrp{- \Cm f(r_s)}\\
                &\quad + \ind{r_s > \sqrt{12}R} \cdot 2r_s + 5\beta + \frac{\c}{L}\lrn{\nablat_s}_2\\
              &\overset{(iii)}{\leq} - \Cm f(r_s) + \ind{r_s > \sqrt{12}R} \cdot \lrp{\Cm f(r_s) + 2r_s} + 5\beta + \frac{\c}{L}\lrn{\nablat_s}_2\\
               &\overset{(iv)}{\leq} - \Cm f(r_s) + \ind{r_s > \sqrt{12}R} \cdot \lrp{4 r_s} + 5\beta + \frac{\c}{L}\lrn{\nablat_s}_2,
                \numberthis \label{e:t:wsfdfje:3}
                \end{align*}
                 where $(i)$ is by Lemma~\ref{lem:fpropertiesall} (F4) combined with the choice of $\Cf$ and $\Rf$, third line is by Lemma~\ref{lem:fpropertiesall} (F2) and Lemma~\ref{lem:fpropertiesall} (F3). $(ii)$ follows immediately from the definition of $\Cm$ in \eqref{d:cm}. $(iii)$ can be verified from algebra, and finally $(iv)$ is from the fact that $\Cm \leq 1$ and $f(r) \leq r$ for all $r$ (Lemma~\ref{lem:fpropertiesall} (F3)).
                
                
                Thus, by combining the bounds on $\spadesuit$ and $\heartsuit$ in Eqs.~\eqref{e:t:wsfdfje:3} back into Eq.~\eqref{e:t:pdfvm},
                \begin{align*}
                d \mu_k f(r_s)
                \leq& - \mu_k \Cm f(r_s) ds \\
                &\quad + \mu_k\lrp{\ind{r_s > \sqrt{12}R} \cdot 4 r_s  + 5\beta + \frac{\c}{L} \lrn{\nablat_s}_2} ds\\
                &\quad + \mu_k \lin{\nabla_w f(r_s), 4\sqrt{\frac{\c}{L}} \lrp{\gamma_s \gamma_s^T dB_s + \frac{1}{2} \bar{\gamma}_s \bar{\gamma}_s^T dA_s}}\\
                \leq& - \mu_k \Cm f(r_s) ds\\
                &\quad + \mu_k\lrp{\ind{r_s > \sqrt{12}R} \cdot 4 r_s ds + 5\beta + \c \lrn{x_s - x_{\step{s}{\delta} \delta}}_2} ds\\
                &\quad + \mu_k \lin{\nabla_w f(r_s), 4\sqrt{\frac{\c}{L}} \lrp{\gamma_s \gamma_s^T dB_s + \frac{1}{2} \bar{\gamma}_s \bar{\gamma}_s^T dA_s}}.
                \numberthis \label{e:t:gfkd}
                \end{align*}
                
                By taking the time derivative of Eq.~\eqref{d:xi}-\eqref{d:phi}, we can verify that for $s\in[k\nu, (k+1)\nu)$,
                \begin{align*}
                d \mu_k \xi_s =& - \mu_k \cdot \Cm \xi_s ds + \mu_k \cdot \c \lrn{x_s - x_{\step{t}{\delta} \delta}}_2 ds,\\
                d \sigma_s =& - \mu_k \Cm \sigma_s ds + \mu_k \cdot \ind{r_s \geq \sqrt{12}R} \cdot 4 r_s ds,\\
                d \phi_s =& -  \mu_k \Cm \phi_s ds + \mu_k \cdot \lin{\nabla_w f(r_s), 4\sqrt{\frac{\c}{L}}\lrp{\gamma_s \gamma_s^T dB_s + \frac{1}{2} \bar{\gamma}_s \bar{\gamma}_s^T dA_s}}.
                \end{align*}
                
                By combining with Eq.~\eqref{e:t:gfkd} we get 
                
                \begin{align*}
                d \lrp{\mu_k \cdot \lrp{f(r_s) - \xi_s}- \sigma_s - \phi_s }
                \leq& - \Cm \lrp{\mu_k \cdot \lrp{f(r_s) - \xi_s}  - \lrp{\sigma_s + \phi_s}} + 5\beta ds.
                \end{align*}
                
    An application of Gr\"onwall's Lemma over the interval $s\in[k\nu, (k+1)\nu)$ gives us the claimed result:
                \begin{align*}
                & \mu_k \cdot \lrp{f(r_{(k+1)\nu}) - \xi_{(k+1)\nu}}  - \lrp{\sigma_{(k+1)\nu} + \phi_{(k+1)\nu}}\\
                &\qquad \qquad \qquad \leq e^{-\Cm \nu} \lrp{ \mu_k \cdot \lrp{f(r_{k\nu}) - \xi_{k\nu}}  - \lrp{\sigma_{k\nu} + \phi_{k\nu}}} + 5\beta \nu.
                \end{align*}
            \end{proof}
        \end{subsection}
        \begin{subsection}{Main results for synchronous coupling}\label{s:synchronous}
            Our main result in this section is Lemma~\ref{l:muk=0_norm_contraction_3}, which shows that over a period of $\Ts$, $f(r_s)$ contracts by an amount $\exp\lrp{-\Cm \Ts}$ with probability one. Note that this is weaker than showing a contraction rate of $\exp(-\Cm t)$ for all $t$, but is sufficient for our purposes.
            \begin{lemma}\label{l:muk=0_norm_contraction_3}
                Assume that $e^{72LR^2} \geq 2$. With probability one, for all $k$, 
                \begin{align*}
                 &\ind{k\nu = \tau_{k-1}+ \Ts} \cdot \lrp{f\lrp{r_{k\nu}} -\xi_{k\nu}}\\
                &\qquad \qquad \qquad \leq \ind{k\nu = \tau_{k-1}+ \Ts} \cdot \exp\lrp{-\Cm\Ts} \cdot \lrp{f\lrp{r_{\tau_{k-1}}} - \xi_{\tau_{k-1}}} + 5\beta.
                \end{align*}
                
            \end{lemma}
            \begin{proof}
                From our definition of $\c$ in Eq.~\eqref{d:c}, $r_t$ in Eq.~\eqref{d:r}, and from Lemma~\ref{l:def_l}.1, it can be verified that
                \begin{align*}
                r_{k\nu}
                \leq& 1.002 \lrp{\lrn{z_{k\nu}}_2 + \lrn{z_{k\nu} + w_{k\nu}}_2} + 2\beta \\
                \leq& \sqrt{2.002} \sqrt{\lrn{z_{k\nu}}_2^2 + \lrn{z_{k\nu} + w_{k\nu}}_2^2} + 2\beta.
                \end{align*}
                On the other hand, by $\lrn{\cdot}_1 \geq \lrn{\cdot}_2$ and by Lemma~\ref{l:def_l}, 
                \begin{align*}
                r_{\tau_{k-1}} \geq \sqrt{\lrn{z_{\tau_{k-1}}}_2^2 + \lrn{z_{\tau_{k-1}} + w_{\tau_{k-1}}}_2^2} - 2\beta.
                \end{align*}
                Combining the inequality in the display above with the statement of Lemma~\ref{l:muk=0_norm_contraction_2} gives:
                \begin{align*}
                \ind{k\nu = \tau_{k-1}+ \Ts} \cdot {r_{k\nu}} 
                & \leq \sqrt{\frac{47}{50}} \cdot \ind{k\nu = \tau_{k-1}+ \Ts}\cdot {r_{\tau_{k-1}}}\\
                &\quad + \ind{k\nu = \tau_{k-1}+ \Ts} \cdot \c\int_{\tau_{k-1}}^{k\nu} e^{- \frac{\c^2}{3}({k\nu}-t)} \lrn{x_t - x_{\step{t}{\delta} \delta}}_2 dt + 5\beta.
                \end{align*}
                Combining the above with (F2), (F3) and (F6) of Lemma~\ref{lem:fpropertiesall}, and by using the definition of $f$ in Eq.~\eqref{d:fparameters},
                \begin{align*}
                & \ind{k\nu = \tau_{k-1}+ \Ts} \cdot f(r_{k\nu}) \\
                &\qquad \leq \ind{k\nu = \tau_{k-1}+ \Ts} \cdot \exp\lrp{-\frac{1-\sqrt{47/50}}{4}e^{-6 LR^2}}f(r_{\tau_{k-1}})\\
                &\qquad \qquad + \ind{k\nu = \tau_{k-1}+ \Ts} \cdot \c\int_{\tau_{k-1}}^{k\nu} e^{- \frac{\c^2}{3}({k\nu}-t)} \lrn{x_t - x_{\step{t}{\delta} \delta}}_2 dt + 5\beta\\
                &\qquad \leq \ind{k\nu = \tau_{k-1}+ \Ts} \cdot \exp\lrp{-\Cm\Ts}f\lrp{r_{\tau_{k-1}}}\\
                &\qquad\qquad + \ind{k\nu = \tau_{k-1}+ \Ts} \cdot \c\int_{\tau_{k-1}}^{k\nu} e^{- \frac{\c^2}{3}({k\nu}-t)} \lrn{x_t - x_{\step{t}{\delta} \delta}}_2 dt + 5\beta\\
                & \qquad \overset{(i)}{\leq} \ind{k\nu = \tau_{k-1}+ \Ts} \cdot \exp\lrp{-\Cm\Ts}f\lrp{r_{\tau_{k-1}}}\\
                &\qquad\qquad + \ind{k\nu = \tau_{k-1}+ \Ts} \cdot \c\int_{\tau_{k-1}}^{k\nu} e^{- \Cm({k\nu}-t)} \lrn{x_t - x_{\step{t}{\delta} \delta}}_2 dt + 5\beta,
                \numberthis \label{e:t:tinv:9}
                \end{align*}
                where the first line in $(i)$ follows from the definition of $\Ts$ and $\Cm$ in Eq.~\eqref{d:ts} and Eq.~\eqref{d:cm} along with the fact that $(1-\sqrt{47/50})/4 \geq 1/200$. The second line in $(i)$ is because $\Cm \leq \frac{\c^2}{3}$ from Eq.~\eqref{d:cm}.
                
                By definition of $\xi_t$ in Eq.~\eqref{d:xi}, 
                \begin{align*}
                & \ind{k\nu = \tau_{k-1}+ \Ts} \xi_{k\nu} \\
                &\qquad =  \ind{k\nu = \tau_{k-1}+ \Ts} \cdot \int_{0}^{k\nu} e^{- \Cm (k\nu - t)} \c \lrn{x_t - x_{\step{t}{\delta} \delta}}_2 dt\\
                &\qquad = \ind{k\nu = \tau_{k-1}+ \Ts} \cdot e^{- \Cm (k\nu - \tau_{k-1})} \int_{0}^{\tau_{k-1}} e^{- \Cm (\tau_{k-1} - t)} \c\lrn{x_t - x_{\step{t}{\delta} \delta}}_2 dt\\
                &\qquad \qquad  +  \ind{k\nu = \tau_{k-1}+ \Ts} \cdot \int_{\tau_{k-1}}^{k\nu} e^{-\Cm(k\nu - t)} \c \lrn{x_t - x_{\step{t}{\delta} \delta}}_2 dt\\
                &\qquad = \ind{k\nu = \tau_{k-1}+ \Ts} \cdot \exp\lrp{- \Cm (k\nu - \tau_{k-1})} \xi_{\tau_{k-1}} \\
                &\qquad \qquad + \c \int_{\tau_{k-1}}^{k\nu} \exp\lrp{-\Cm(k\nu - t)}\lrn{x_t - x_{\step{t}{\delta} \delta}}_2 dt\\
                &\qquad =  \ind{k\nu = \tau_{k-1}+ \Ts} \cdot \exp\lrp{- \Cm \Ts} \xi_{\tau_{k-1}} \\
                &\qquad \qquad + \c \int_{\tau_{k-1}}^{k\nu} \exp\lrp{-\Cm(k\nu - t)}\lrn{x_t - x_{\step{t}{\delta} \delta}}_2 dt.
                \numberthis \label{e:t:tinv:2}
                \end{align*}
                By subtracting the left and the right hand sides of Eq.~\eqref{e:t:tinv:2} and Eq.~\eqref{e:t:tinv:9} thus gives us that,
                \begin{align*}
                & \ind{k\nu = \tau_{k-1}+ \Ts} \cdot \lrp{f\lrp{r_{k\nu}} - \xi_{k\nu}} \\
                &\qquad \qquad \leq \ind{k\nu = \tau_{k-1}+ \Ts} \cdot \exp\lrp{-\Cm\Ts} \cdot \lrp{f\lrp{r_{\tau_{k-1}}} - \xi_{\tau_{k-1}}} + 5\beta.
                \end{align*}
            \end{proof}
            
            We now state and prove several auxillary lemmas which are required for the proof of Lemma \ref{l:muk=0_norm_contraction_3}.

            \begin{lemma}
                \label{l:synchronous_algebra}
                If $\lrn{z_s}_2^2 + \lrn{z_s + w_s}_2^2 \geq 2.2 R^2$, then
                \begin{align*}
                \lin{z_s, w_s} + \lin{z_s + w_s, - w_s - \frac{\c}{L} \nabla_s} \leq - \frac{\c^2}{3} \lrp{\lrn{z_s}_2^2 + \lrn{z_s + w_s}_2^2}.
                \end{align*}
            \end{lemma}
            
            \begin{proof} We begin by expanding the differentials $d\lrn{z_s}_2^2 + d\lrn{z_s+w_s}_2^2$:
                \begin{align*}
                 d \lrn{z_s}_2^2 + d\lrn{z_s + w_s}_2^2
                &= 2\lin{z_s, w_s} + 2\lin{z_s + w_s, - w_s - \frac{\c}{L} \nabla_s}\\
                &= {-2 \|w_s\|_2^2 - 2\lin{z_s,  \frac{\c}{L}\nabla_s} - 2\lin{w_s,  \frac{\c}{L}\nabla_s}} \\
                &= {-2\|w_s\|_2^2 - 2\lin{z_s,  \frac{\c}{L}\nabla_s} + \|w_s\|_2^2 + \frac{\c^2}{L^2}\|\nabla_t\|_2^2 - \|w_t + \frac{\c}{L}\nabla_t\|_2^2} \\
                &\leq {-\|w_s\|_2^2 - 2\lin{z_s,  \frac{\c}{L}\nabla_s} + \frac{\c^2}{L^2}\|\nabla_s\|_2^2}\\
                &\leq {-\|w_s\|_2^2 - 2\lin{z_s,  \frac{\c}{L}\nabla_s} + \c^2 \|z_s\|_2^2} =:\spadesuit.
                \numberthis \label{e:uq}
                \end{align*}
                
                Now consider two cases.
                
                \textbf{Case 1: ($\lrn{z_s}_2 \leq R$})
                By Young's inequality,
                \begin{align*}
                \lrn{z_s + w_s}_2^2 \leq 11 \lrn{w_s}_2^2 + 1.1 \lrn{z_s}_2^2.
                \end{align*}
                Furthermore, by our assumption that $\lrn{z_s}_2^2 + \lrn{z_s + w_s}_2^2 \geq 2.2 R^2$, 
                \begin{align*}
                11\lrn{w_s}_2^2
                & \geq \lrn{z_s + w_s}_2^2 - 1.1 \lrn{z_s}_2^2\\
                &= \lrn{z_s}_2^2 + \lrn{z_s + w_s}_2^2 - 1.1 \lrn{z_s}_2^2 - \lrn{z_s}_2^2\\
                &\geq  2.2R^2 - 2.1 R^2\\
                &\geq 0.1 R^2\\
                &\geq 0.1 \lrn{z_s}_2^2,\\
                \implies \lrn{z_s}_2^2 \leq & \frac{1000}{9} \lrn{w_s}_2^2.
                \numberthis \label{e:t:nsqus}
                \end{align*}
                 With this implication $\spadesuit$ can now be upper bounded by
                \begin{align*}
                \spadesuit
                &=  - \lrn{w_s}_2^2 - 2 \lin{z_s, \frac{\c}{L}\nabla_s} + \c^2 \lrn{z_s}_2^2\\
                &\overset{(i)}{\leq} -\lrn{w_s}_2^2 + 2\c \lrn{z_s}_2^2 + \c^2 \lrn{z_s}_2^2\\
                &\overset{(ii)}{\leq} -\lrn{w_s}_2^2 + 3\c \lrn{z_s}_2^2\\
                &\overset{(iii)}{\leq} - \frac{2}{3} \lrn{w_s}_2^2\\
                &\overset{(iv)}{\leq} - \frac{\c^2}{3} \lrp{\lrn{z_s}_2^2 + \lrn{z_s + w_s}_2^2},
                \end{align*}
                where $(i)$ is by Assumption~\ref{ass:smoothness} and Cauchy-Schwarz, and $(ii)$ is because $\c := \frac{1}{1000 \kappa}  \leq \frac{1}{1000}$. The inequality $(iii)$ is by the implication in Eq.~\eqref{e:t:nsqus}, which gives $3\c \lrn{z_s}_2^2 \leq \frac{1000 \c}{3} \lrn{w_s}_2^2 \leq \frac{1}{3} \lrn{w_s}_2^2$. Finally, $(iv)$ can be verified as follows:
                \begin{align*}
                \|z_s\|_2^2 + \|z_s + w_s\|_2^2
                &\overset{(i)}\leq  3\|z_s\|_2^2 + 2\|w_s\|_2^2 \\
                &\overset{(ii)}\leq  \frac{1000}{3} \|w_t\|_2^2 + 2\|w_t\|_2^2 \\
                &\leq  \frac{1006}{3} \|w_t\|_2^2.\\
                \implies \qquad \qquad \lrp{\lrn{z_s}_2^2 + \lrn{z_s + w_s}_2^2} &\leq \frac{1006}{3} \lrn{w_s}_2^2\\
                &\leq \frac{1}{2\c} \lrn{w_s}_2^2.\\
                \implies \qquad \qquad \qquad \qquad \qquad
                \frac{2}{3} \lrn{w_s}_2^2 
                &\geq \frac{4\c}{3} \lrp{\lrn{z_s}_2^2 + \lrn{z_s + w_s}_2^2}\\
               &\overset{(iii)}{\geq} \frac{\c^2}{3} \lrp{\lrn{z_s}_2^2 + \lrn{z_s + w_s}_2^2},
                \end{align*}
                where $(i)$ is by Young's inequality, $(ii)$ is by Eq.~\eqref{e:t:nsqus}, and $(iii)$ is by $\c \leq \frac{1}{1000}$.

                \textbf{Case 2: ($\lrn{z_s}_2 \geq R$}) We have,
                \begin{align*}
                \spadesuit = & - \lrn{w_s}_2^2 - 2 \lin{z_s, \frac{\c}{L}\nabla_s} + \c^2 \lrn{z_s}_2^2\\
                \overset{(i)}{\leq} & - \lrn{w_s}_2^2 - 2\c^2 \lrn{z_s}_2^2 + \c^2 \lrn{z_s}_2^2\\
                \leq & - \lrn{w_s}_2^2 - \c^2 \lrn{z_s}_2^2\\
                \leq & - \c^2 \lrp{\lrn{w_s}_2^2 + \lrn{z_s}_2^2}\\
                \overset{(ii)}{\leq} & - \frac{\c^2}{3} \lrp{\lrn{z_s}_2^2 + \lrn{z_s + w_s}_2^2},
                \end{align*}
                where $(i)$ is by Assumption~\ref{ass:strongconvexity} and $(ii)$ is because
                \begin{align*}
                 \lrn{z_s}_2^2 + \lrn{z_s + w_s}_2^2
                \leq& 3\lrn{z_s}_2^2 + 2\lrn{w_s}_2^2\\
                \leq& 3 \lrp{\lrn{z_s}_2^2 + \lrn{w_s}_2^2}.
                \end{align*}
Hence, we have proved the result under both cases.
            \end{proof}
            \begin{lemma}\label{l:muk=0_norm_contraction}
                With probability one, 
                \begin{align*}
                & \lrp{1-\mu_k} \cdot \lrp{\sqrt{\lrn{z_{(k+1)\nu}}_2^2 + \lrn{z_{(k+1)\nu} + w_{(k+1)\nu}}_2^2} - \sqrt{2.2} R}_+\\
                & \qquad \qquad \qquad \leq \lrp{1-\mu_k} \cdot e^{-\frac{\c^2 \nu}{3}} \lrp{\sqrt{\lrn{z_{k\nu}}_2^2 + \lrn{z_{k\nu} + w_{k\nu}}_2^2} - \sqrt{2.2} R}_+\\
                & \qquad \qquad \qquad \qquad + \lrp{1-\mu_k} \cdot \c\int_{k\nu}^{(k+1)\nu} e^{- \frac{\c^2}{3}({(k+1)\nu}-t)} \lrn{x_t - x_{\step{t}{\delta} \delta}}_2 dt.
                \end{align*}
            \end{lemma}
            \begin{proof} When $\mu_k=1$, the inequality holds trivially ($0=0$), so for the rest of this proof, we consider the case $\mu_k=0$. To simplify notation, we leave out the multiplier $(1-\mu_k)$ in all subsequent expressions.
                
                We can verify from Eqs. \eqref{d:xt}-\eqref{d:vt} and Eq.~\eqref{d:mut} that when $\mu_k = 0$, for any $s\in[k\nu, (k+1)\nu)$,
                \begin{align*}
                d z_s =& w_s ds\\
                d \lrp{z_s + w_s} =& \lrp{-w_s +  \frac{\c}{L} \lrp{\nabla_s + \nablat_s}}ds.
                \end{align*}
                Thus, for any $s\in[k\nu, (k+1)\nu)$,
                \begin{align*}
                & d \lrp{\lrp{\sqrt{\|z_s\|_2^2 + \|z_s + w_s\|_2^2} - \sqrt{2.2} R}_+^2 }\\
                & \qquad \overset{(i)}{=}  \frac{\lrp{\sqrt{\|z_s\|_2^2 + \|z_s + w_s\|_2^2} - \sqrt{2.2} R}_+}{\sqrt{\|z_s\|_2^2 + \|z_s + w_s\|_2^2}}\lin{\cvec{z_s}{z_s + w_s}, \cvec{w_s}{-w_s - \frac{\c}{L} \lrp{\nabla_s + \nablat_s}}} ds\\
                & \qquad \overset{(ii)}{\le} - \frac{\c^2}{3} \frac{\lrp{\sqrt{\|z_s\|_2^2 + \|z_s + w_s\|_2^2} - \sqrt{2.2} R}_+}{\sqrt{\|z_s\|_2^2 + \|z_s + w_s\|_2^2}} \cdot { \lrp{\lrn{z_s}_2^2 + \lrn{z_s + w_s}_2^2}}ds \\
                &\qquad\qquad + \frac{\lrp{\sqrt{\|z_s\|_2^2 + \|z_s + w_s\|_2^2} - \sqrt{2.2} R}_+}{\sqrt{\|z_s\|_2^2 + \|z_s + w_s\|_2^2}} \cdot  \lrp{\lrn{z_s + w_s}_2 \lrn{\nablat_s}_2} ds\\
                & \qquad \leq  - \frac{\c^2}{3} {\lrp{\sqrt{\|z_s\|_2^2 + \|z_s + w_s\|_2^2} - \sqrt{2.2} R}_+} \cdot {\sqrt{\|z_s\|_2^2 + \|z_s + w_s\|_2^2}}ds \\
                & \qquad\qquad + \frac{\c}{L} {\lrp{\sqrt{\|z_s\|_2^2 + \|z_s + w_s\|_2^2} - \sqrt{2.2} R}_+} \cdot \lrn{\nablat_s}_2 ds\\
                & \qquad \le - \frac{\c^2}{3} \cdot \lrp{\sqrt{\|z_s\|_2^2 + \|z_s + w_s\|_2^2} - \sqrt{2.2} R}_+^2 ds\\
                & \qquad \qquad + \c {\lrp{\sqrt{\|z_s\|_2^2 + \|z_s + w_s\|_2^2} - \sqrt{2.2} R}_+} \cdot \lrn{x_s - x_{\step{s}{\delta} \delta}}_2 ds,
                \end{align*}
                where $(i)$ is by the expression for $d z_s$ and $d w_s$ established above, and $(ii)$ is by Lemma~\ref{l:synchronous_algebra} and Cauchy-Schwarz, the last two inequalities follow by algebraic manipulations.
                
                Dividing throughout by $\lrp{\sqrt{\|z_s\|_2^2 + \|z_s + w_s\|_2^2} - \sqrt{2.2} R}_+$ gives us that
                \begin{align*}
                & d \lrp{\sqrt{\|z_s\|_2^2 + \|z_s + w_s\|_2^2} - \sqrt{2.2} R}_+\\ 
                &\qquad \qquad \leq \lrp{- \frac{\c^2}{3}\lrp{\sqrt{\|z_s\|_2^2 + \|z_s + w_s\|_2^2} - \sqrt{2.2} R}_+ + \c\lrn{x_s - x_{\step{s}{\delta} \delta}}_2} dt.
                \end{align*}        
                We can verify that the inequality implies that
                \begin{align*}
                & d \lrp{\lrp{\sqrt{\|z_s\|_2^2 + \|z_s + w_s\|_2^2} - \sqrt{2.2} R}_+ - \c\int_{k\nu}^s e^{- \frac{\c^2}{3}(s-t)} \lrn{x_t - x_{\step{t}{\delta} \delta}}_2 dt}\\
                & \qquad \qquad \leq - \frac{\c^2}{3} \lrp{\lrp{\sqrt{\|z_s\|_2^2 + \|z_s + w_s\|_2^2} - \sqrt{2.2} R}_+ - \c\int_{k\nu}^s e^{- \frac{\c^2}{3}(s-t)} \lrn{x_t - x_{\step{t}{\delta} \delta}}_2 dt} dt.
                \end{align*}
                Thus by Gr\"onwall's Lemma,
                \begin{align*}
 &\lrp{\sqrt{\|z_{(k+1)\nu}\|_2^2 + \|z_{(k+1)\nu} + w_{(k+1)\nu}\|_2^2} - \sqrt{2.2} R}_+ - \c\int_{k\nu}^{(k+1)\nu} e^{- \frac{\c^2}{3}({(k+1)\nu}-t)} \lrn{x_t - x_{\step{t}{\delta} \delta}}_2 dt\\
                &\leq e^{- \frac{\c^2}{3} ((k+1)\nu-k\nu)} \lrp{\lrp{\sqrt{\|z_{k\nu}\|_2^2 + \|z_{k\nu} + w_{k\nu}\|_2^2} - \sqrt{2.2} R}_+ - \c\int_{k\nu}^{k\nu} e^{- \frac{\c^2}{3}({k\nu}-t)} \lrn{x_t - x_{\step{t}{\delta} \delta}}_2 dt }\\
                &= e^{- \frac{\c^2 \nu}{3}} \lrp{\sqrt{\|z_{k\nu}\|_2^2 + \|z_{k\nu} + w_{k\nu}\|_2^2} - \sqrt{2.2} R}_+.
                \end{align*}
                This proves the statement of the Lemma.
            \end{proof}

            \begin{lemma}\label{l:muk=0_norm_contraction_2}
                Assume that $e^{72LR^2} \geq 2$. With probability one, for all positive integers $k$,
                \begin{align*}
                &\ind{k\nu = \tau_{k-1}+ \Ts} \cdot \sqrt{\lrn{z_{k\nu}}^2 + \lrn{z_{k\nu} + w_{k\nu}}^2} \\
                &\qquad \qquad\qquad \qquad\leq \sqrt{\frac{23}{50}} \cdot \ind{k\nu = \tau_{k-1}+ \Ts}\cdot \sqrt{\lrn{z_{\tau_{k-1}}}^2 + \lrn{z_{\tau_{k-1}} + w_{\tau_{k-1}}}^2}\\
                &\qquad \qquad\qquad \qquad\qquad  + \ind{k\nu = \tau_{k-1}+ \Ts} \cdot \c\int_{\tau_{k-1}}^{k\nu} e^{- \frac{\c^2}{3}({k\nu}-t)} \lrn{x_t - x_{\step{t}{\delta}\delta}}_2 dt + 3\beta.
                \end{align*}
            \end{lemma}
            \begin{proof}
                By our choice $\nu$ we know that $\Ts/\nu$ is an integer, thus we have,
                \begin{align*}
                k\nu = \tau_{k-1} + \Ts 
                \Rightarrow 
                (k-1)\nu < \tau_{k-1} + \Ts.
                \end{align*}
                Thus,
                \begin{align*}
                 \ind{k\nu = \tau_{k-1} + \Ts}
                &\overset{(i)}{=} \ind{k\nu = \tau_{k-1} + \Ts} \cdot \ind{(k-1)\nu < \tau_{k-1} + \Ts}\\
                &\overset{(ii)}{=} \ind{k\nu = \tau_{k-1} + \Ts} \cdot (1-\mu_{k-1}) \\
                &\overset{(iii)}{=} \ind{k\nu = \tau_{k-1} + \Ts} \cdot \prod_{i\in S_{k-1}} (1-\mu_i),
                \numberthis \label{e:t:qos}
                \end{align*}
                where $S_{k-1} := \lrbb{\frac{\tau_{k-1}}{\nu}, \frac{\tau_{k-1}}{\nu} + 1, ..., k-1}$ (as defined in Lemma \ref{l:over_time_synchronous}).
                Above, $(i)$ is because $k\nu = \tau_{k-1} + \Ts \Rightarrow (k-1)\nu < \tau_{k-1} + \Ts$, $(ii)$ is because $(k-1)\nu < \tau_{k-1} + \Ts \Rightarrow \mu_{k-1}= 0$ (see Eq.~\eqref{d:mut}) and $(iii)$ is by Part 2 of Lemma~\ref{l:over_time_synchronous}.

                We can now recursively apply Lemma~\ref{l:muk=0_norm_contraction} as follows: (to simplify notation, let $\alpha:= \ind{k\nu = \tau_{k-1} + \Ts}$):
                \begin{align*}
                & \alpha \cdot \prod_{i\in S_{k-1}} (1-\mu_i) \cdot \lrp{\sqrt{\lrn{z_{k\nu}}_2^2 + \lrn{z_{k\nu} + w_{k\nu}}_2^2} - \sqrt{2.2} R}_+\\
                &\qquad \leq \alpha \cdot \prod_{i\in S_{k-1}} (1-\mu_i) \cdot e^{-\frac{\c^2}{3} \nu} \lrp{\sqrt{\lrn{z_{(k-1)\nu}}_2^2 + \lrn{z_{(k-1)\nu} + w_{(k-1)\nu}}_2^2} - \sqrt{2.2} R}_+\\
                &\qquad \qquad  + \alpha \cdot \prod_{i\in S_{k-1}} (1-\mu_i) \cdot \c\int_{(k-1)\nu}^{k\nu} e^{- \frac{\c^2}{3}({k\nu}-t)} \lrn{x_t - x_{\step{t}{\delta} \delta}}_2 dt\\
                & \qquad \leq \alpha \cdot \prod_{i\in S_{k-1}} (1-\mu_i) \cdot e^{-\frac{\c^2}{3} \Ts}\lrp{\sqrt{\lrn{z_{\tau_{k-1}}}_2^2 + \lrn{z_{\tau_{k-1}} + w_{\tau_{k-1}}}_2^2} - \sqrt{2.2} R}_+\\
                &\qquad \qquad + \alpha \cdot \prod_{i\in S_{k-1}} (1-\mu_i) \cdot \c\int_{\tau_{k-1}}^{k\nu} e^{- \frac{\c^2}{3}({k\nu}-t)} \lrn{x_t - x_{\step{t}{\delta} \delta}}_2 dt,
                \numberthis \label{e:t:qos2}
                \end{align*}
                where the last inequality uses the fact that $\nu \cdot (k- \tau_{k-1}) = \Ts$ in the definition of $\alpha$.
                
                Thus, we have,
                \begin{align*}
                & \ind{k\nu = \tau_{k-1}+ \Ts} \cdot \lrp{\sqrt{\lrn{z_{k\nu}}_2^2 + \lrn{z_{k\nu} + w_{k\nu}}_2^2} - \sqrt{2.2} R}_+\\
                & \qquad \overset{(i)}{=} \ind{k\nu = \tau_{k-1}+ \Ts} \cdot \prod_{i\in S_{k-1}} (1-\mu_i) \cdot \lrp{\sqrt{\lrn{z_{k\nu}}_2^2 + \lrn{z_{k\nu} + w_{k\nu}}_2^2} - \sqrt{2.2} R}_+\\
                & \qquad \overset{(ii)}{\le} \ind{k\nu = \tau_{k-1}+ \Ts} \cdot \prod_{i\in S_{k-1}} (1-\mu_i) \cdot \lrp{\sqrt{\lrn{z_{k\nu}}_2^2 + \lrn{z_{k\nu} + w_{k\nu}}_2^2} - \sqrt{2.2} R}_+\\
                &\qquad \qquad + \ind{k\nu = \tau_{k-1}+ \Ts} \cdot \prod_{i\in S_{k-1}} (1-\mu_i) \cdot \c\int_{\tau_{k-1}}^{k\nu} e^{- \frac{\c^2}{3}({k\nu}-t)} \lrn{x_t - x_{\step{t}{\delta} \delta}}_2 dt\\
                & \qquad \overset{(iii)}{=} \ind{k\nu = \tau_{k-1}+ \Ts} \cdot e^{-\frac{\c^2}{3} \Ts}\lrp{\sqrt{\lrn{z_{\tau_{k-1}}}_2^2 + \lrn{z_{\tau_{k-1}} + w_{\tau_{k-1}}}_2^2} - \sqrt{2.2} R}_+\\
                &\qquad \qquad + \ind{k\nu = \tau_{k-1}+ \Ts} \cdot \c\int_{\tau_{k-1}}^{k\nu} e^{- \frac{\c^2}{3}({k\nu}-t)} \lrn{x_t - x_{\step{t}{\delta} \delta}}_2 dt\\
                & \qquad \overset{(iv)}{\le} \ind{k\nu = \tau_{k-1}+ \Ts} \cdot \frac{1}{100} \lrp{\sqrt{\lrn{z_{\tau_{k-1}}}_2^2 + \lrn{z_{\tau_{k-1}} + w_{\tau_{k-1}}}_2^2} - \sqrt{2.2} R}_+\\
                &\qquad \qquad + \ind{k\nu = \tau_{k-1}+ \Ts} \cdot \c\int_{\tau_{k-1}}^{k\nu} e^{- \frac{\c^2}{3}({k\nu}-t)} \lrn{x_t - x_{\step{t}{\delta} \delta}}_2 dt,
                \numberthis \label{e:t:qos3}
                \end{align*}
                where $(i)$ is by Eq.~\eqref{e:t:qos}, $(ii)$ is by Eq.~\eqref{e:t:qos2}, $(iii)$ is by Eq.~\eqref{e:t:qos} again, and $(iv)$ is by the definition $\Ts = \frac{3}{\c^2} \log(100)$.
                
                Let $j := \tau_{k-1}/\nu$. Then by the first part of Lemma~\ref{l:over_time_synchronous}, we know that $\tau_j = \tau_{k-1}= j\nu$. From the update rule for $\tau_k$, Eq.~\eqref{d:taut}, this must imply that
                \begin{align*}
    \sqrt{\lrn{z_{\tau_{k-1}}}_2^2 + \lrn{z_{\tau_{k-1}} + w_{\tau_{k-1}}}_2^2} 
                =& \sqrt{\lrn{z_{j\nu}}_2^2 + \lrn{z_{j\nu} + w_{j\nu}}_2^2}
                \geq \sqrt{5} R.
                \numberthis \label{e:t:qos4}
                \end{align*}
                
                Thus finally,
                \begin{align*}
                & \ind{k\nu = \tau_{k-1}+ \Ts} \cdot \lrp{\sqrt{\lrn{z_{k\nu}}_2^2 + \lrn{z_{k\nu} + w_{k\nu}}_2^2}}\\
                &\qquad \overset{(i)}{\leq} \ind{k\nu = \tau_{k-1}+ \Ts} \cdot \lrp{\sqrt{\lrn{z_{k\nu}}_2^2 + \lrn{z_{k\nu} + w_{k\nu}}_2^2} - \sqrt{2.2} R}_+\\
                &\qquad \qquad + \ind{k\nu = \tau_{k-1}+ \Ts} \cdot \sqrt{2.2} R\\
                &\qquad \overset{(ii)}{\leq} \ind{k\nu = \tau_{k-1}+ \Ts} \frac{1}{100} \lrp{\sqrt{\lrn{z_{\tau_{k-1}}}_2^2 + \lrn{z_{\tau_{k-1}} + w_{\tau_{k-1}}}_2^2} - \sqrt{2.2} R}_+\\
                &\qquad \qquad + \ind{k\nu = \tau_{k-1}+ \Ts} \cdot \sqrt{2.2} R\\
                &\qquad \qquad + \ind{k\nu = \tau_{k-1}+ \Ts} \cdot \c\int_{\tau_{k-1}}^{k\nu} e^{- \frac{\c^2}{3}({k\nu}-t)} \lrn{x_t - x_{\step{t}{\delta} \delta}}_2 dt\\
                &\qquad \overset{(iii)}{\leq} \ind{k\nu = \tau_{k-1}+ \Ts} \frac{1}{100} \lrp{\sqrt{\lrn{z_{\tau_{k-1}}}_2^2 + \lrn{z_{\tau_{k-1}} + w_{\tau_{k-1}}}_2^2} - \sqrt{2.2} R}\\
                &\qquad \qquad + \ind{k\nu = \tau_{k-1}+ \Ts} \cdot \sqrt{\frac{22}{50}} \sqrt{\lrn{z_{\tau_{k-1}}}_2^2 + \lrn{z_{\tau_{k-1}} + w_{\tau_{k-1}}}_2^2}\\
                &\qquad \qquad + \ind{k\nu = \tau_{k-1}+ \Ts} \cdot \c\int_{\tau_{k-1}}^{k\nu} e^{- \frac{\c^2}{3}({k\nu}-t)} \lrn{x_t - x_{\step{t}{\delta} \delta}}_2 dt\\
                &\qquad \overset{(iv)}{\leq} \ind{k\nu = \tau_{k-1}+ \Ts} \sqrt{\frac{23}{50}} \sqrt{\lrn{z_{\tau_{k-1}}}_2^2 + \lrn{z_{\tau_{k-1}} + w_{\tau_{k-1}}}_2^2}\\
                &\qquad \qquad + \ind{k\nu = \tau_{k-1}+ \Ts} \cdot \c\int_{\tau_{k-1}}^{k\nu} e^{- \frac{\c^2}{3}({k\nu}-t)} \lrn{x_t - x_{\step{t}{\delta} \delta}}_2 dt,
                \end{align*}
                where $(i)$ is by an algebraic manipulation, $(ii)$ is by Eq.~\eqref{e:t:qos3}, $(iii)$ is by Eq.~\eqref{e:t:qos4} and $(iv)$ is because $1/100 + \sqrt{22/50} \leq \sqrt{23/50}$.
            \end{proof}
            
            \begin{lemma}\label{l:over_time_synchronous} Let $k$ be a positive integer, then:
            \begin{enumerate}
                \item Let $j = \tau_k/\nu$. Then for all $i \in \lrbb{j,j+1, ..., k}$, $\tau_i = \tau_k = j\nu$.
                \item If $\mu_k = 0$, then $\mu_i = 0$ for all $i\in\lrbb{\tau_k/\nu...k}$, $\mu_i = 0$. Equivalently,
                \begin{align*}
                \ind{\mu_k = 0} = \prod_{i\in S_k} \ind{\mu_i = 0},
                \end{align*}
                where $S_k := \lrbb{\frac{\tau_k}{\nu},...,k}$.
            \end{enumerate}
        \end{lemma}
        \begin{proof}
            For the first claim: By definition of the update for $\tau_k$, if $j = \tau_k/\nu$ for any $k$, then $j\nu = \tau_{j} = \tau_k$. Note that $\tau_i$ is nondecreasing with $i$, so that $j = \tau_k \leq k$, which implies that $\tau_j \leq \tau_{j+1} \leq ... \leq \tau_j$. Since $\tau_j = \tau_k$, the inequalities must hold with equality.
            
            For the second claim: By the definition of $\mu_k$; $\mu_k = 0$ implies that $k\nu < \tau_k + \Ts$. From the first claim, we know that for all $i \in \lrbb{\tau_{k}/\nu ...k}$, $\tau_i = \tau_k$. Thus $i \nu \leq k\nu < \tau_k + \Ts = \tau_i + \Ts$.
        \end{proof}
        \end{subsection}
    
        \begin{subsection}{Discretization Error Bound}\label{s:discretization}
            In this section, we bound the various \emph{discretization errors}. First, in Section \ref{ss:e_xit}, we establish a bound on $\E{\xi_t}$. Then in Lemma \ref{l:sigma_bound}, we bound $\E{\sigma_t}$. Finally, in Lemma \ref{l:phi_bound}, we show that $\E{\phi_t}=0$ as it is a martingale. 
            
            \begin{subsubsection}{Bound on $\E{\xi_t}$}\label{ss:e_xit}
                In this subsection, we establish a bound on $\E{\xi_t}$. This term represents the discretization error that arises because in the SDE in Eq.~\eqref{d:ut}, the update to $u_t$ uses the gradient $\nabla U\lrp{x_{\step{t}{\delta}\delta}}$ instead of $\nabla U\lrp{x_t}$. Our main result is Lemma \ref{l:xi_bound}, which in turn relies on the uniform bound for all $t \ge 0$ on $\E{\lrn{x_{t} - x_{\step{t}{\delta} \delta}}_2^8}$ established in Corollary~\ref{l:xt-xinu} (based on the moment bounds established in Appendix~\ref{s:bounding_energy}).
                \begin{lemma}\label{l:xi_bound}
                    For all $t\ge 0$,
                    \begin{align*}
                    \E{\xi_t}
                    \leq \delta \cdot \frac{2^9 \c \lrp{R + \sqrt{d/m}}}{\Cm}.
                    \end{align*}
                \end{lemma}
                \begin{proof}
                    By the bound in Corollary~\ref{l:xt-xinu},
                    \begin{align*}
                    \E{\lrn{x_{t} - x_{\step{t}{\delta} \delta}}_2^8} \leq \delta^8 2^{72} \lrp{R^2 + \frac{d}{m}}^4.
                    \end{align*}
                    Further, by Jensen's inequality,
                    \begin{align*}
                    \E{\lrn{x_t - x_{\step{t}{\delta} \delta}}_2} \leq \delta \cdot 2^9 \lrp{R + \sqrt{\frac{d}{m}}}.
                    \end{align*}
                    By integrating from up to time $t$,
                    \begin{align*}
                    \E{\xi_t}
                    =& \int_0^t e^{-\Cm (t-s)} \E{\c{\lrn{x_t-x_{\step{s}{\delta} \delta}}_2}} ds\\
                    \leq& \int_0^t e^{-\Cm (t-s)} \c \delta 2^9 \lrp{R + \sqrt{\frac{d}{m}}}ds\\
                    \leq& \delta \cdot \frac{2^9 \c \lrp{R + \sqrt{d/m}}}{\Cm}.
                    \end{align*}
                \end{proof}
                \begin{corollary}\label{l:xt-xinu}
                    For all $t\ge 0$,
                    \begin{align*}
                    \E{\lrn{x_{t} - x_{\step{t}{\delta} \delta}}_2^8} \leq \delta^8 2^{72} \lrp{R^2 + \frac{d}{m}}^4.
                    \end{align*}
                \end{corollary}
                \begin{proof}
                    This follows directly by combining the results of Lemma~\ref{l:energy_bound_all_time} and Lemma~\ref{l:vndj}.
                \end{proof}
                
                \begin{lemma}\label{l:vndj}
                    Suppose that the step size $\delta \leq \frac{1}{1000}$. Then for all $t\in[\step{t}{\delta} \delta, (\step{t}{\delta}+1)\delta]$,
                    \begin{align*}
                    \E{\lrn{x_{t} - x_{\step{t}{\delta} \delta}}_2^8} \leq \delta^8 \lrp{1.1 \E{\lrp{\lrn{x_{\step{t}{\delta} \delta}}_2^8 + \lrn{u_{\step{t}{\delta} \delta}}_2^8}} + 2^{12} \lrp{R^2 + \frac{d}{m}}^4}.
                    \end{align*}
                \end{lemma}
                \begin{proof} 
                    \begin{align*}
                    \E{\lrn{x_t - x_{\step{t}{\delta} \delta}}_2^8}
                    &= \E{\lrn{\int_{\step{t}{\delta} \delta}^t w_s ds }_2^8}\\
                    &\leq \delta^7 \int_{\step{t}{\delta} \delta}^t \E{\lrn{w_s}_2^8} ds\\
                    &= \delta^8 \lrp{1.1 \E{\lrp{\lrn{x_{\step{t}{\delta} \delta}}_2^8 + \lrn{u_{\step{t}{\delta} \delta}}_2^8}} + 2^{12} \lrp{R^2 + \frac{d}{m}}^4},
                    \end{align*}
                    where for the last inequality, we use Lemma~\ref{l:energy_bound_uniform}.
                    
                \end{proof}
                
            \end{subsubsection}
            \begin{subsubsection}{Bounds on $\E{\sigma_t}$ and $\E{\phi_t}$}\label{ss:e_sigmat}
                In this subsection, we bound $\E{\sigma_t}$ (Lemma \ref{l:sigma_bound}). This term represents the discretization error that arises because $\tau_k$ (and hence $\mu_k$) is updated at discrete time intervals of $\nu$. We highlight the fact that $\E{\sigma_t}$ is bounded by a term that depends on $\nu$, which can be made arbitrarily small. The main ingredient of this proof is a bound on $\E{\mu_k \cdot \ind{r_s\geq \sqrt{12}R}}$ in Lemma~\ref{l:exit_probability}.
                \begin{lemma}\label{l:sigma_bound}
                    For $\beta \leq 0.0001R$. There exists a $\Cd_5=poly(L, 1/m, d, R, \frac{1}{\Cm})$ and $\Cd_3=1/poly(L, 1/m, d, R)$, such that for all $\nu \leq \Cd_3$, for all positive integers $k$, and for all $t\ge 0$,
                    \begin{align*}
                    \E{\sigma_t}
                    \leq C_5 \nu^2 .
                    \end{align*}
                \end{lemma}
                \begin{proof} By the definition of $\sigma_t$ in Eq.~\eqref{d:sigma},
                    \begin{align*}
                    \E{\sigma_t}
                    &= \E{\int_0^t \mu_{\step{s}{\nu}}\cdot e^{-\Cm (t-s)} \ind{r_s \geq \sqrt{12}R} \cdot 4 r_s ds}\\
                    &= 4 \int_0^t e^{-\Cm (t-s)} \E{\mu_{\step{s}{\nu}}\ind{r_s \geq \sqrt{12}R} r_s }ds\\
                    &\overset{(i)}{\leq} 4 \int_0^t e^{-\Cm (t-s)} \nu^2 \cdot C_4\\
                    &\le \frac{4\nu^2 C_4}{\Cm} \\
                    &= \nu^2 \cdot C_5,
                    \end{align*}
                    where $(i)$ is by Corollary~\ref{l:error_muk}.
                \end{proof}

                \begin{lemma}\label{l:energy_bound_rs2}
                    For all $s\ge 0$, 
                    \begin{align*}
                    \E{r_s^2} \leq 2^{32} \lrp{R^2 + \frac{d}{m}}.
                    \end{align*}
                \end{lemma}
                
                \begin{proof}
                    Recall that,
                    \begin{align*}
                    r_s^2 
                    &= \lrp{(1+ 2\c) \lrn{z_s}_2 + \lrn{z_s + w_s}_2}^2\\
                    &\leq \lrp{(2+2\c) \lrn{z_s}_2 + \lrn{w_s}_2}^2\\
                    &\leq \lrp{2.1 \lrn{x_s}_2+ 2.1 \lrn{y_s}_2 + \lrn{u_s}_2 + \lrn{v_s}_2}^2\\
                    &\overset{(i)}{\leq} 16 \lrp{\lrn{x_s}_2^2 + \lrn{u_s}_2^2 + \lrn{y_s}_2^2 + \lrn{v_s}_2^2}\\
                    &\leq 2^{16} \lrp{2^{72} \lrp{R^2 + \frac{d}{m}}^4}^{1/4}\\
                    &= 2^{32} \lrp{R^2 + \frac{d}{m}},
                    \end{align*}
                    where $(i)$ is by Lemma~\ref{l:energy_bound_all_time} and Lemma~\ref{l:energy_bound_all_time_y}.
                \end{proof}
                
                \begin{lemma}\label{l:exit_probability}
                    For every $\beta \leq 0.0001R$, there exists a $\Cd_2=poly(L, 1/m, d, R)$, $\Cd_3=1/poly(L, 1/m, d, R)$, such that for all $\nu \leq \Cd_3$, for all positive integers $k$, and for all $s\in [k\nu, (k+1)\nu]$,
                    \begin{align*}
                    \E{\mu_{k} \cdot \ind{r_s \geq \sqrt{12}R}} \leq \Cd_2 \nu^4 .
                    \end{align*}
                \end{lemma}
                
                \begin{proof}
                    By definition of $\mu_k$ in Eq.~\eqref{d:mut}, we know that $\mu_k=1$ implies that $k\nu - \tau_k \geq \Ts$ which further implies that $\tau_k = \tau_{k-1}$ (otherwise $\tau_k$ must equal $k\nu$ by the definition of $\tau_t$, in which case $k\nu -\tau_k = 0 < \Ts$). This then implies that $k\nu - \tau_{k-1} \geq \Ts$. It must thus be the case that $\sqrt{\lrn{z_{k\nu}}_2^2 + \lrn{z_{k\nu} + w_{k\nu}}_2^2} < \sqrt{5} R$, because otherwise $\tau_k = k\nu$, which contradicts $\mu_k =1$. Thus,
                    \begin{align*}
                        \mu_k \leq \ind{\sqrt{\lrn{z_{k\nu}}_2^2 + \lrn{z_{k\nu} + w_{k\nu}}_2^2} < \sqrt{5}R}.
                        \numberthis \label{e:t:tnmqnw:1}
                    \end{align*}
                    By a standard inequality between $\lrn{\cdot}_1$ and $\lrn{\cdot}_2$,
                    \begin{align*}
                     \sqrt{\lrn{z_{k\nu}}_2^2 + \lrn{z_{k\nu} + w_{k\nu}}_2^2}
                    &\geq \frac{1}{\sqrt{2}}\lrp{\lrn{z_{k\nu}}_2 + \lrn{z_{k\nu} + w_{k\nu}}_2}\\
                    &\overset{(i)}{\geq} \frac{1}{\sqrt{2}}\lrp{\l\lrp{z_{k\nu}} + \l\lrp{z_{k\nu} + w_{k\nu}}} - \beta\\
                    &\overset{(ii)}{\geq} \frac{1}{1.002 \sqrt{2}} r_{k\nu} -\beta,
                    \end{align*}
                    where $(i)$ is by Lemma~\ref{l:def_l}.1, and $(ii)$ is by definition of $r_t$ in Eq.~\eqref{d:mut} and by definition of $\c$.
                    
                    Combining  with the inequality~\eqref{e:t:tnmqnw:1},
                    \begin{align*}
                    \mu_k 
                    &\leq \ind{\frac{1}{1.002 \sqrt{2}} r_{k\nu} -\beta < \sqrt{5}R}\\
                    &= \ind{r_{k\nu} < 1.002 \sqrt{10}R + \beta}\\
                    &\leq \ind{r_{k\nu} < \sqrt{11}R},
                    \numberthis \label{e:tnmqnw:1}
                    \end{align*}
                    where the final inequality uses our
                    assumption that $\beta \leq 0.0001 R$.
                    Thus,
                    \begin{align*}
                    \mu_k \cdot \ind{r_s \geq \sqrt{12} R} 
                    \leq& \ind{r_{k\nu} < \sqrt{11} R} \cdot \ind{r_s \geq \sqrt{12} R}\\
                    \leq& \ind{\lrabs{r_s - r_{\kd} }\geq 0.14 R}.
                    \end{align*}
                    Taking expectations,
                    \begin{align*}
                    \E{\mu_k \cdot \ind{r_s \geq \sqrt{12}R}}
                    &\leq \E{\ind{\lrabs{r_s - r_{\kd} }\geq 0.14 R}}\\
                    &\overset{(i)}{\leq} \frac{\E{(r_s - r_{\kd})^8}}{(0.14 R)^8}\\
                    &\overset{(ii)}{\leq} \frac{2^{10}\E{\lrn{z_s - z_{k\nu}}_2^8 + \lrn{w_s - w_{k\nu}}_2^8} + 2^{10}\beta^4}{(0.14 R)^8},
                 \numberthis\label{e:t:lkdfiow:1}
                    \end{align*}
                    where $(i)$ by Markov's inequality, $(ii)$ can be verified by using Lemma~\ref{l:def_l}.1 and some algebra.
                    
                    Next, by the dynamics of $z_t$ we have that 
               \begin{align*}
                    \lrn{z_s - z_{\kd}}_2^8
                    &= \lrn{\int_{\kd}^s w_s dt}_2^8\\
                    &\leq \lrp{s-k\nu}^7\int_{\kd}^s \lrn{w_s}_2^8 dt\\
                    &\leq 2^3 \lrp{s-k\nu}^7\int_{\kd}^s \lrn{u_s}_2^8 + \lrn{v_s}_2^8 dt.
                    \numberthis \label{e:t:wrkgnqw:1}
                    \end{align*}
                    Further by the definition of the dynamics of $w_t$ we get,
                 \begin{align*}
                    &\lrn{w_s - w_{k\nu}}_2^8\\
                    &= \lrn{\int_{k\nu}^s -2 w_t - \frac{\c}{L}\nabla U(x_{\step{t}{\delta}}) + \frac{\c}{L}\nabla U(y_t) dt + 4\sqrt{\frac{\c}{L}} \int_{\kd}^s \gamma_t \gamma_t^T dB_t + 2 \sqrt{\frac{\c}{L}} \int_{\kd}^s \bar{\gamma}_t \bar{\gamma}_t^T dA_t }_2^8\\
                    &\overset{(i)}{\leq} 2^{20} \lrp{s-k\nu}^7 \lrp{\int_{k\nu}^s \lrn{w_t}_2^8 + \frac{\c^8}{L^8} \lrn{\nabla U(y_t)}_2^8 + \frac{\c^8}{L^8} \lrn{\nabla U(x_{\step{t}{\nu}})}_2^8  dt}\\
                    &\qquad + 2^{12} \frac{\c^4}{L^4} \lrn{\int_{k\nu}^s \gamma_t \gamma_t^T dB_t}_2^8 + 2^{12} \frac{\c^4}{L^4} \lrn{\int_{k\nu}^s \bar{\gamma}_t \bar{\gamma}_t^T dA_t}_2^8\\
     &\overset{(ii)}{\leq} 2^{30} \lrp{s-k\nu}^7 \lrp{\int_{k\nu}^s \lrn{u_t}_2^8 + \lrn{v_t}_2^8 + {\c^8} \lrn{y_t}_2^8 + {\c^8} \lrn{x_{\step{t}{\nu}}}_2^8  dt}\\
                    &\qquad + 2^{12} \frac{\c^4}{L^4} \lrn{\int_{k\nu}^s \gamma_t \gamma_t^T dB_t}_2^8 + 2^{12} \frac{\c^4}{L^4} \lrn{\int_{k\nu}^s \bar{\gamma}_t \bar{\gamma}_t^T dA_t}_2^8\\
                    &\overset{(iii)}{\leq} 2^{30} \lrp{s-k\nu}^7 \lrp{\int_{k\nu}^s \lrn{u_t}_2^8 + \lrn{v_t}_2^8 + \lrn{y_t}_2^8 +  \lrn{x_{\step{t}{\nu}}}_2^8  dt}\\
                    &\qquad + 2^{12} \frac{1}{L^4} \lrn{\int_{k\nu}^s \gamma_t \gamma_t^T dB_t}_2^8 + 2^{12} \frac{1}{L^4} \lrn{\int_{k\nu}^s \bar{\gamma}_t \bar{\gamma}_t^T dA_t}_2^8,
                    \numberthis \label{e:t:wrkgnqw:2}
                    \end{align*}
                    where $(i)$ is by the triangle inequality and Young's inequality, $(ii)$ uses Assumption~\ref{ass:smoothness}, and $(iii)$ uses the fact that $\c \leq 1$.
                    
                    Therefore, summing the two inequalities above and taking expectations,
                 \begin{align*}
                 &\E{\lrn{z_s - z_{\kd}}_2^8 + \lrn{w_s - w_{k\nu}}_2^8}\\
                    &\qquad \leq \E{2^{30} \lrp{s-k\nu}^7 \lrp{\int_{\kd}^s \lrn{x_{\step{t}{\delta} \delta}}_2^8 + \lrn{u_t}_2^8 + \lrn{y_t}_2^8 + \lrn{v_t}_2^8 dt}}\\
                    &\qquad \qquad + \E{2^{12} \frac{1}{L^4} \lrn{\int_{k\nu}^s \gamma_t \gamma_t^T dB_t}_2^8 + 2^{12} \frac{1}{L^4} \lrn{\int_{k\nu}^s \bar{\gamma}_t \bar{\gamma}_t^T dA_t}_2^8}\\
                    &\qquad  \leq 2^{32} (s-k\nu)^{8}\lrp{R^2 + \frac{d}{m}}^4 + 2^{52} \cdot (s-k\nu)^4 \cdot \frac{1}{L^4},
                    \end{align*}
                    where the last inequlaity is by combining  Lemma~\ref{l:energy_bound_all_time}, Lemma \ref{l:energy_bound_all_time_y} and Lemma \ref{l:gamma_t_moment} and by noting that by their definition in Eq.~\eqref{d:gamma}, $\lrn{\gamma_t}_2\leq 1$ and $\lrn{\bar{\gamma}_t}_2\leq 1$ for all $t$, with probability one. 
                    
                    There exists $C_1 = poly(R, d, \frac{1}{m})$ and $C_3 = 1/poly(R, d, \frac{1}{m})$, such that for all $\nu<C_3$ and for all $s\in[k\nu, (k+1)\nu]$, the right-hand side of the inequality above is upper bounded by
                    \begin{align*}
                    \E{\lrn{z_s - z_{\kd}}_2^8 + \lrn{w_s - w_{k\nu}}_2^8} \leq& \nu^4 C_1.
                    \end{align*}
                    
                    Combining the above with inequality \eqref{e:t:lkdfiow:1}, we find that there exists $C_2 = poly(R, d, \frac{1}{m})$ and $C_3 = 1/poly(R, d, \frac{1}{m})$, such that for all $\nu<C_3$ and for all $s\in[k\nu, (k+1)\nu]$
                    \begin{align*}
                    \E{\mu_k \cdot \ind{r_s \geq \sqrt{12}R}}
                    \leq& \frac{\E{(r_s - r_{\kd})^8}}{(0.14 R)^8}\\
                    \leq& \frac{2^{10}\E{\lrn{z_s - z_{k\nu}}_2^8 + \lrn{w_s - w_{k\nu}}_2^8}  + 2^{10} \beta^4}{(0.14 R)^8}\\
                    \leq& \nu^4 \Cd_2,
                    \end{align*}
                    where $\beta$ is absorbed into $\Cd_2$ due to our assumption that $\beta\leq 0.0001 R$.

                \end{proof}
            
                \begin{corollary}\label{l:error_muk}
                    For $\beta \leq 0.0001R$. There exists constants, $\Cd_3=1/poly(L, 1/m, d, R)$ and $\Cd_4=poly(L, 1/m, d, R)$, such that for all $\nu \leq \Cd_3$, for all positive integers $k$, and for all $s\in [k\nu, (k+1)\nu]$,
                    \begin{align*}
                    \E{\mu_k \ind{r_s \geq \sqrt{12}R} r_s}
                    \leq \sqrt{\E{\mu_k \ind{r_s \geq \sqrt{12}R}}}\sqrt{\E{r_s^2}}
                    \leq C_4 \nu^2.
                    \end{align*}
                \end{corollary}
                \begin{proof}
                    Proof follows by combining the results of Lemma~\ref{l:energy_bound_rs2} and Lemma~\ref{l:exit_probability}.
                \end{proof}

                \begin{lemma}\label{l:gamma_t_moment}
                    Let $\gamma_t$ be a $d$-dimensional adapted process satisfying $\lrn{\gamma_t}_2 \leq 1$ for all $t>0$ with probability one. Then
                    \begin{align*}
                    \E{\lrn{\int_0^t \gamma_s \gamma_s^T dB_s}_2^8} \leq 2^{20}t^4.
                    \end{align*}
                \end{lemma}
                \begin{proof}
                    Let us define $\beta_t := \int_0^t \gamma_s \gamma_s^T dB_s$. Define the function $l(\beta):= \lrn{\beta}_2^8$ for this proof. The derivates of this function are,
                    \begin{align*}
                    \nabla l(\beta) &= 8 l(\beta)^{3/4} \beta\\
                    \nabla^2 l(\beta) &= 8 l(\beta)^{3/4} I + 48 l(\beta)^{2/4} \beta\beta^T.
                    \end{align*}
                    By $\Ito$'s Lemma,
                    \begin{align*}
                    d l(\beta_t) 
                    =& \lin{8 l(\beta_t)^{3/4} \beta_t , \beta_t \beta_t^T dB_t} + 4 l(\beta_t)^{3/4} \lrn{\gamma_t}_2^2 dt + 24 l(\beta_t)^{2/4} \lrp{\lin{\beta_t, \gamma_t}}^2 \lrn{\gamma_t}_2^2 dt\\
                    \leq& \lin{8 l(\beta_t)^{3/4} \beta_t , \beta_t \beta_t^T dB_t} + 4 l(\beta_t)^{3/4}dt + 24 l(\beta_t)^{2/4} \lrn{\beta_t}_2^2 dt\\
                    =& \lin{4 l(\beta_t)^{3/4} \beta_t , \beta_t \beta_t^T dB_t} +  28 l(\beta_t)^{3/4}dt.
                    \end{align*}
                    Taking expectations, 
                    \begin{align*}
                    \ddt \E{l(\beta_t)} 
                    \leq& 28 \E{l(\beta_t)^{3/4}}
                    \leq 28 \E{l(\beta_t)}^{3/4}.
                    \end{align*}
                    Thus,
                    \begin{align*}
                    & \ddt \E{l(\beta_t)}^{1/4} \leq 28\\
                    \Rightarrow \qquad & \E{l(\beta_t)}^{1/4} \leq 28t\\
                    \Rightarrow \qquad & \E{l(\beta_t)} \leq 2^{20} t^4,
                    \end{align*}
                    as claimed.
                \end{proof}
            \end{subsubsection}

            \begin{lemma}\label{l:phi_bound}
                For all $t \ge0$, $\E{\phi_t}=0$.
            \end{lemma}
            \begin{proof}
                By the definition of $\phi_t$ it is a martingale. Hence, $\E{\phi_t}=0$.
            \end{proof}
        \end{subsection}
    
        \begin{subsection}{Putting it all together}\label{s:putting_together}
            In this section, we combine the results from Appendices~\ref{s:reflection}, \ref{s:synchronous} and \ref{s:discretization} to prove Theorem \ref{t:maintheoremunderdamped0}. The heart of the proof is Lemma \ref{l:combining_4_cases}, which shows that $\L_t$ contracts with probability one at a rate of $-\Cm$. This lemma essentially combines the results of Lemmas~\ref{l:mu01}, \ref{l:mu00} (proved in Appendix~\ref{s:reflection}) and Lemmas \ref{l:mu11}, \ref{l:mu10} (proved in Appendix~\ref{s:synchronous}).

            \begin{proof}[Proof of Theorem \ref{t:maintheoremunderdamped0}]
                From Lemma~\ref{l:combining_4_cases} we have,
                \begin{align*}
                \L_{k\nu} \leq e^{-\Cm k\nu} \L_0 + \frac{(3\nu + 5) \beta }{\Cm \nu}.
                \numberthis \label{e:t:vfdnkjs:1}
                \end{align*}
                while from Lemma~\ref{l:convenient_lower_bound},
                \begin{align*}
                f(r_{k\nu}) \leq 200 \L_{k\nu} + 400\xi_{k\nu} + \sigma_{k\nu} + \phi_{k\nu} + 400\beta.
                \end{align*}
                        Taking expectations, 
                \begin{align*}
                &\E{f(r_{k\nu})} \\
                & \qquad \overset{(i)}{\leq} 200 \E{\L_{k\nu}} + 400\E{\xi_{k\nu}} + \E{\sigma_{k\nu}} + \E{\phi_{k\nu}} + 400\beta\\
                & \qquad \overset{(ii)}{\leq} 200 e^{-\Cm k\nu} \E{\L_0} + 400\E{\xi_{k\nu}} + \E{\sigma_{k\nu}} + \E{\phi_{k\nu}} + \frac{2000 (\nu+1)}{\Cm \nu} \beta\\
                & \qquad= 200 e^{-\Cm k\nu} \E{f(r_0)} + 400\E{\xi_{k\nu}} + \E{\sigma_{k\nu}} + \E{\phi_{k\nu}} + \frac{2000 (\nu+1)}{\Cm \nu} \beta\\
                & \qquad\leq 200 e^{-\Cm k\nu} \E{r_0} + 400\E{\xi_{k\nu}} + \E{\sigma_{k\nu}} + \E{\phi_{k\nu}} + \frac{3000 (\nu+1)}{\Cm \nu} \beta,
                \end{align*}
                where $(i)$ is by Eq.~\eqref{e:t:vfdnkjs:1} and $(ii)$ can be verified from the initialization in Eq.~\eqref{d:x0} and the definition of the Lyapunov function $\L_t$ in Eq.~\eqref{d:L}.
                
                From Lemmas \ref{l:xi_bound}, \ref{l:sigma_bound} and \ref{l:phi_bound},
                \begin{align*}
                400\E{\xi_{k\nu}} + \E{\sigma_{k\nu}} + \E{\phi_{k\nu}} \leq \delta \cdot \frac{2^{18} \c \lrp{R + \sqrt{d/m}}}{\Cm}  + C_5 \nu^2,
                \end{align*}
                where $C_5 = poly(L,1/m,d,R,1/\Cm)$ as defined in Lemma~\ref{l:sigma_bound}.
                
                From Lemma~\ref{l:energy_bound_all_time_y}, our choice of $x_0 = u_0 = 0$ in Eq.~\eqref{d:x0} and our definition of $r_t$ in Eq.~\eqref{d:mut},
                \begin{align*}
                \E{r_0} 
                &\leq 3 \E{\lrn{y_0}_2 + \lrn{v_0}_2} 
                \leq 2^{10} \lrp{R + \sqrt{\frac{d}{m}}} + 3\beta.
                \end{align*}
                By plugging the bound on $\E{r_0}$ and $\E{\xi_{k\nu}}$ into the bound on $\E{f(r_{k\nu})}$ above gives us that
                \begin{align*}
                \E{f(r_{k\nu})} \leq e^{-\Cm k\nu} 2^{18} \lrp{R + \sqrt{\frac{d}{m}}} + \delta \cdot \frac{2^{18} \c \lrp{R + \sqrt{d/m}}}{\Cm}  + C_5 \nu^2 +  \frac{3000 (\nu+1)}{\Cm \nu} \beta.
                \end{align*}
                This inequality along with (F3) of Lemma~\ref{lem:fpropertiesall}, and Lemma~\ref{l:def_l}.1 also implies that,
                \begin{align*}
                 \E{\lrn{z_{k\nu}}_2}
                &\leq\E{r_{k\nu}} + \beta\\
                &\leq 2 e^{6 LR^2} \cdot \E{f(r_{ku})} + \beta\\
                &\leq e^{6 LR^2} \cdot e^{-\Cm k\nu} 2^{19} \lrp{R + \sqrt{\frac{d}{m}}} + e^{6 LR^2} \cdot \delta \cdot \frac{2^{19} \c \lrp{R + \sqrt{d/m}}}{\Cm}  \\
                &\qquad + 2 e^{6 LR^2} \cdot \lrp{ C_5 \nu^2 + \frac{3000 (\nu+1)}{\Cm \nu} \beta + \beta}.
                \end{align*}
                
                We can take $\nu$ and $\beta$ to be arbitrarily small without any additional computation cost, so let $\nu = \lrp{{2^{20}\delta \lrp{R + \sqrt{d/m}
                }/(\Cm C_5) }}^{-1/2}$ and $\beta = \min\lrbb{{2^{20}\delta \lrp{R + \sqrt{d/m}
                }/(\Cm) }, {2^{9}\delta \lrp{R + \sqrt{d/m}
                }},  {2^{9}\delta \nu \lrp{R + \sqrt{d/m}
                }}}$, so that the terms containing $\beta$ and $\nu$ are less than the other terms.
                
                We can ensure that the second term $\lrp{e^{6 LR^2} \cdot \delta \cdot \frac{2^{19} \c \lrp{R + \sqrt{d/m}}}{\Cm}}$ is less than $\epsilon/2$ by setting
                \begin{align*}
                \delta = \epsilon 2^{-20}e^{-6 LR^2} \frac{\Cm}{R + \sqrt{d/m}} \frac{1}{\c}.
                \end{align*}
                We can ensure that the first term $\lrp{e^{6 LR^2} \cdot e^{-\Cm k\nu} 2^{19} \lrp{R + \sqrt{\frac{d}{m}}}}$ is less than $\epsilon/2$ by setting 
                \begin{align*}
                k\nu \geq \frac{\log \frac{1}{\epsilon} + 6 LR^2 + \log\lrp{2^{20} \lrp{R^2 + \frac{d}{m}}}}{\Cm}.
                \end{align*}
                Recalling the definition of $\Cm := \min\lrbb{\frac{e^{-6 LR^2}}{6000\kappa L R^2}, \frac{e^{-6 LR^2}}{21  \cdot 10^7 \cdot  \log\lrp{100} \cdot \kappa^2}, \frac{1}{3\cdot 10^6 \kappa^2}}$ in Eq.~\eqref{d:cm}, and $\c := 1/(1000 \kappa)$, some algebra shows that it suffices to let
                \begin{align*}
                \delta = \frac{\epsilon}{R + \sqrt{d/m}} \cdot e^{-12 LR^2} \cdot 2^{-35} \min \lrp{\frac{1}{LR^2}, \frac{1}{\kappa}}.
                \end{align*}
                The number of steps of the algorithm is thus
                \begin{align*}
                n = \frac{k\nu}{\delta} 
                & \geq 2^{60} \cdot \frac{R + \sqrt{d/m}}{\epsilon} \cdot e^{18LR^2} \cdot \kappa \cdot \max\lrbb{LR^2, \kappa}^2 \cdot \lrp{\log \frac{1}{\epsilon} + LR^2 + \log \lrp{R^2 + \frac{d}{m}}}\\
                &= \widetilde{\mathcal{O}}\left(\frac{\sqrt{d}}{\epsilon}e^{18LR^2}\right).
                \end{align*}
                This completes the proof.
            \end{proof}
            \begin{lemma}\label{l:l_lowerbound_frk}
            With probability one, for all positive integers $k$, 
            \begin{align*}
            (1-\mu_k) \cdot f(r_{k\nu}) \leq (1-\mu_k) \cdot 2\lrp{ f(r_{\tau_k}) + \c\int_{\tau_k}^{k\nu} e^{- \frac{\c^2}{3}({k\nu}-t)} \lrn{x_t - x_{\step{t}{\delta} \delta}}_2 dt} + 6\beta.
            \end{align*}
        \end{lemma}
        \begin{proof}
            First, by Eq.~\eqref{d:r} and Lemma~\ref{l:def_l}.1, 
            \begin{align*}
            (1-\mu_k) \cdot r_{k\nu} 
            =& (1-\mu_k) \cdot \lrp{(1+2\c) \l\lrp{z_{k\nu}}_2 + \l\lrp{z_{k\nu} + w_{k\nu}}}\\
            \leq& (1-\mu_k) \cdot \lrp{(1+2\c) \lrn{z_{k\nu}}_2 + \lrn{z_{k\nu} + w_{k\nu}}} + 3\beta.
            \end{align*}
            Note that by Lemma~\ref{l:over_time_synchronous} we have,
            \begin{align*}
            1-\mu_k = \ind{\mu_k = 0} = \prod_{i\in S_k} \ind{\mu_i = 0} = \prod_{i\in S_k} (1-\mu_i)
            \numberthis \label{e:t:fgwkemf:1},
            \end{align*}
            where $S_k := \lrbb{\frac{\tau_k}{\nu},...,k}$.
            Thus using this characterization of $1-\mu_k$ we get,
            \begin{align*}
            & (1-\mu_k) \cdot \lrp{(1+2\c) \lrn{z_{k\nu}}_2 + \lrn{z_{k\nu} + w_{k\nu}}_2}\\
            & \qquad \qquad \overset{(i)}{\leq} (1-\mu_k) \cdot 2\sqrt{\lrn{z_{k\nu}}_2^2 + \lrn{z_{k\nu} + w_{k\nu}}_2^2}\\
            & \qquad \qquad\overset{(ii)}{\leq} (1-\mu_k) \cdot 2 \lrp{\lrp{\sqrt{\lrn{z_{k\nu}}_2^2 + \lrn{z_{k\nu} + w_{k\nu}}_2^2} - \sqrt{2.2} R}_+ + \sqrt{2.2}R},
            \end{align*}
            where $(i)$ is by defintion of $\c$ in Eq.~\eqref{d:c} and $(ii)$ inequality is by algebra. Unpacking this further we get that:
            \begin{align*}
            & (1-\mu_k) \cdot \lrp{(1+2\c) \lrn{z_{k\nu}}_2 + \lrn{z_{k\nu} + w_{k\nu}}_2}\\
            &  \quad \overset{(i)}{\le} (1-\mu_k) \cdot 2 \lrp{ \lrp{\prod_{i\in S_k} (1-\mu_i)}\cdot \lrp{\sqrt{\lrn{z_{k\nu}}_2^2 + \lrn{z_{k\nu} + w_{k\nu}}_2^2} - \sqrt{2.2} R}_+ + \sqrt{2.2}R}\\
            & \quad \overset{(ii)}{\le} (1-\mu_k) \cdot 2 \lrp{\lrp{\prod_{i\in S_k} (1-\mu_i)}\cdot e^{-\frac{\c^2}{3} (k\nu - \tau_k)}\lrp{\sqrt{\lrn{z_{\tau_k}}_2^2 + \lrn{z_{\tau_k} + w_{\tau_k}}_2^2} - \sqrt{2.2} R}_+}\\
            &\qquad + (1-\mu_k) \cdot 2 \lrp{ \lrp{\prod_{i\in S_k} (1-\mu_i)}\cdot \c\int_{\tau_k}^{k\nu} e^{- \frac{\c^2}{3}({k\nu}-t)} \lrn{x_t - x_{\step{t}{\delta} \delta}}_2 dt}\\ & \qquad + (1-\mu_k) \cdot 2 \lrp{\sqrt{2.2}R}\\
            & \quad \overset{(iii)}{\leq} (1-\mu_k) \cdot 2 \lrp{\lrp{\sqrt{\lrn{z_{\tau_k}}_2^2 + \lrn{z_{\tau_k} + w_{\tau_k}}_2^2} - \sqrt{2.2} R}_+ + \sqrt{2.2}R}\\
            &\qquad + (1-\mu_k) \cdot 2 \lrp{\c\int_{\tau_k}^{k\nu} e^{- \frac{\c^2}{3}({k\nu}-t)} \lrn{x_t - x_{\step{t}{\delta} \delta}}_2 dt}\\
            & \quad \overset{(iv)}{=} (1-\mu_k) \cdot 2 \lrp{\sqrt{\lrn{z_{\tau_k}}_2^2 + \lrn{z_{\tau_k} + w_{\tau_k}}_2^2}}\\
            &\qquad + (1-\mu_k) \cdot 2 \lrp{\c\int_{\tau_k}^{k\nu} e^{- \frac{\c^2}{3}({k\nu}-t)} \lrn{x_t - x_{\step{t}{\delta} \delta}}_2 dt}\\
            & \quad \overset{(v)}{\leq} (1-\mu_k) \cdot 2\lrp{ r_{\tau_k} + \lrp{\c\int_{\tau_k}^{k\nu} e^{- \frac{\c^2}{3}({k\nu}-t)} \lrn{x_t - x_{\step{t}{\delta} \delta}}_2 dt}} + 3\beta,
            \end{align*}
            where $(i)$ is by Eq.~\eqref{e:t:fgwkemf:1}, $(ii)$ follows by Lemma~\ref{l:muk=0_norm_contraction}, applied recursively for $i \in \lrbb{\frac{\tau_k}{\nu} ... k}$, while $(iii)$ is again by Eq.~\eqref{e:t:fgwkemf:1}. The equality in $(iv)$ can be verified as follows: By Lemma~\ref{l:over_time_synchronous} we know that $\tau_{{\tau_k}/{\nu}} = \tau_k$, which implies that $\sqrt{\lrn{z_{\tau_k}}_2^2 + \lrn{z_{\tau_k} + w_{\tau_k}}_2^2} \geq \sqrt{5}R$ based on the dynamics of $\tau_k$ in Eq.~\eqref{d:taut}. Finally $(v)$ is by definition of $r_t$ in Eq.~\eqref{d:r}.
            
            Our conclusion thus follows from the concavity of $f$ and the fact that $f(0) = 0$, so that for all $a,b,c\in \Re^+$, $f(4b) \leq 4 f(b)$ and $a\leq b + c$ implies that $f(a) \leq f(b) + c$:
            \begin{align*}
            (1-\mu_k) \cdot f(r_{k\nu}) \leq (1-\mu_k) \cdot 2\lrp{ f(r_{\tau_k}) + \c\int_{\tau_k}^{k\nu} e^{- \frac{\c^2}{3}({k\nu}-t)} \lrn{x_t - x_{\step{t}{\delta} \delta}}_2 dt} + 6\beta.
            \end{align*}
        \end{proof}
            \begin{lemma}\label{l:convenient_lower_bound}
                For all positive integer $k$, with probability one, 
                \begin{align*}
                f(r_{k\nu}) \leq 200 \L_{k\nu} + 400\xi_{k\nu} + \sigma_{k\nu} + \phi_{k\nu} + 400 \beta.
                \end{align*}
            \end{lemma}
            \begin{proof}
                From Lemma \ref{l:l_lowerbound_frk}, 
                \begin{align*}
                 {(1-\mu_k) \cdot f(r_{k\nu})}
                &\leq 2{(1-\mu_k) \cdot f(r_{\tau_k})} + 2{(1-\mu_k) \cdot \c\int_{\tau_k}^{k\nu} e^{- \frac{\c^2}{3}({k\nu}-t)} \lrn{x_t - x_{\step{t}{\delta} \delta}}_2 dt} + 6\beta\\
                &\leq 2{(1-\mu_k) \cdot f(r_{\tau_k})} + 2(1-\mu_k) {\xi_{k\nu}} + 6 \beta,
                \numberthis \label{e:t:roe}
                \end{align*}
                where the last inequality is by Eq.~\eqref{d:xi}.
                
                We can also verify from the definition of $\mu_t$ in Eq.~\eqref{d:mut} that $\mu_k = 0 \Leftrightarrow k\nu \leq \tau_k + \Ts$. Thus,
                \begin{align*}
                 (1-\mu_k) \cdot e^{-\Cm (k\nu - \tau_k)} 
                &\overset{(i)}{\geq} (1-\mu_k) \cdot e^{-\Cm \Ts}\\
                &\overset{(ii)}{\geq} (1-\mu_k) \cdot \exp\lrp{-\frac{\c^2}{3} \cdot \Ts}\\
                &= (1-\mu_k) \cdot \frac{1}{100},
                \numberthis \label{e:t:roe2}
                \end{align*}
                where $(i)$ is by Eq.~\eqref{d:cm} and $(ii)$ line is by Eq.~\eqref{d:ts}.
                
                Combining the above with the definition of $\xi_{k\nu}$ in Eq.~\eqref{d:xi} we get,
                \begin{align*}
                (1-\mu_k) \xi_{k\nu} 
                &= (1-\mu_k) e^{-\Cm (k\nu -\tau_k)} \xi_{\tau_k} + \int_{\tau_k}^{k\nu} e^{-\Cm (k\nu-s)} \c{\lrn{x_{s}-x_{\step{s}{\delta} \delta}}_2} ds\\
                &\geq (1-\mu_k) e^{-\Cm (k\nu -\tau_k)} \xi_{\tau_k}.
                \numberthis \label{e:t:roe3}
                \end{align*}
                
                Thus,
                \begin{align*}
                \L_{k\nu}
                &\overset{(i)}{=} \mu_k \lrp{f(r_{k\nu}) - \xi_{k\nu}} + (1-\mu_k) \cdot e^{-\Cm (k\nu - \tau_k)} \cdot \lrp{f(r_{\tau_k}) - \xi_{\tau_k}}  - \lrp{\sigma_{k\nu} + \phi_{k\nu}}\\
                &\overset{(ii)}{\ge} \mu_k \lrp{f(r_{k\nu}) - \xi_{k\nu}} + (1-\mu_k) \cdot e^{-\Cm (k\nu - \tau_k)} \cdot \lrp{\frac{1}{2} f(r_{k\nu}) - \xi_{k\nu} - \xi_{\tau_k} -2\beta}  - \lrp{\sigma_{k\nu} + \phi_{k\nu}}\\
                &\overset{(iii)}{\ge} \mu_k \lrp{f(r_{k\nu}) - \xi_{k\nu}} + (1-\mu_k) \cdot  \cdot \lrp{\frac{e^{-\Cm (k\nu - \tau_k)}}{2} f(r_{k\nu}) - 2 \xi_{k\nu} -2\beta}  - \lrp{\sigma_{k\nu} + \phi_{k\nu}}\\
                &\overset{(iv)}{\ge} \mu_k \lrp{f(r_{k\nu}) - \xi_{k\nu}} + (1-\mu_k) \cdot \lrp{\frac{1}{200} f(r_{k\nu}) - 2 \xi_{k\nu} -2\beta}  - \lrp{\sigma_{k\nu} + \phi_{k\nu}}\\
                &\overset{(v)}{\ge} \frac{1}{200} \lrp{\mu_k \cdot f(r_{k\nu}) + (1-\mu_k) \cdot f(r_{k\nu})} - \lrp{2\xi_{k\nu} + \sigma_{k\nu} + \phi_{k\nu}} - 2 \beta\\
                &\overset{(vi)}{=} \frac{1}{200} \lrp{f(r_{k\nu})} - \lrp{2\xi_{k\nu} + \sigma_{k\nu} + \phi_{k\nu}} - 2 \beta,
                \end{align*}
                where $(i)$ is by definition of $\L$ in Eq.~\eqref{d:L}. $(ii)$ is by Eq.~\eqref{e:t:roe}. $(iii)$ is by Eq.~\eqref{e:t:roe3} and the positivity of $f$, $\xi$, $\beta$. $(iv)$ is by Eq.~\eqref{e:t:roe2} and the fact that $f(r_t) \geq 0$ and $\xi_t \geq 0$ for all $t$. The inequalities $(v)$ and $(vi)$ are by algebraic manipulations.
                
                Rearranging terms gives
                \begin{align*}
                f(r_{k\nu}) \leq 200 \L_{k\nu} + 400 \xi_{k\nu} + \sigma_{k\nu} + \phi_{k\nu} + 400 \beta.
                \end{align*}
            \end{proof}

            \begin{lemma}\label{l:combining_4_cases}
                Assume that $e^{72LR^2} \geq 2$. With probability one, for all positive integers $k$,
                \begin{align*}
                \L_{k\nu} \leq e^{-\Cm \nu } \L_{(k-1) \nu} + (3\nu + 5) \beta.   \end{align*}
                Applying this recursively, 
                \begin{align*}
                \L_{k\nu} \leq e^{-\Cm k\nu} \L_0 + \frac{(3\nu + 5) \beta }{\Cm \nu}.
                \end{align*}
            \end{lemma}
            
            \begin{proof}
                We get the conclusion by summing the results of Lemmas \ref{l:mu01}, \ref{l:mu00}, \ref{l:mu11} and \ref{l:mu10}.
            \end{proof}

            Below, we state the lemmas which are needed to prove Lemma~\ref{l:combining_4_cases}.
            \begin{lemma}\label{l:mu01}
                Assume that $e^{72LR^2} \geq 2$. For all positive integers $k$, with probability 1,
                \begin{align*}
                 \ind{\mu_k = 1, \mu_{k-1} = 0}\cdot  \L_{k\nu} 
                \leq& \ind{\mu_k = 1, \mu_{k-1} = 0}\cdot  e^{-\Cm \nu} \L_{(k-1) \nu} + 5\beta.
                \end{align*}
            \end{lemma}
            \begin{proof}
                Given the definition of $\L_t$ in Eq.~\eqref{d:L} we find that $\ind{\mu_k = 1} \L_{k\nu} = \ind{\mu_k = 1} f(r_{k\nu})$ and $\ind{\mu_{k-1} = 0}\L_{(k-1)\nu} = \ind{\mu_{k-1} = 0}\lrp{e^{-\Cm \lrp{(k-1)\nu - \tau_{k-1}}} f(r_{\tau_{k-1}}) - \lrp{\sigma_{(k-1)\nu} + \phi_{(k-1)\nu}}}$.
                
                By the dynamics of $\mu_k$, we can verify that
                \begin{align*}
             \mu_k = 1 
                \Leftrightarrow \quad & k\nu \geq \tau_k + \Ts \\
                \Rightarrow \quad & k\nu \neq \tau_k \\
                \Rightarrow \quad & \tau_k = \tau_{k-1} \\
                \Rightarrow \quad & k\nu \geq \tau_{k-1}+ \Ts.
                \end{align*}
                We can also verify that
                \begin{align*}
                \mu_{k-1} = 0 \Rightarrow (k-1)\nu < \tau_{k-1} + \Ts.
                \end{align*}
                By our choice of $\nu$, $\Ts/\nu$ is an integer (see comment following Eq.~\eqref{d:ts}), and the inequalities above imply that $k\nu = \tau_{k-1}+ \Ts$. Thus,
                \begin{align*}
                \ind{\mu_k = 1, \mu_{k-1} = 0} =\ind{\mu_k = 1, \mu_{k-1} = 0} \cdot \ind{k\nu = \tau_{k-1} + \Ts}.
                \numberthis \label{e:t:nwsdl}
                \end{align*}
                
                To reduce clutter, let us define $\alpha := \ind{\mu_k = 1, \mu_{k-1} = 0}$ and $\alpha':= \ind{k\nu = \tau_{k-1} + \Ts}$. Hence we have,
                \begin{align*}
                 \alpha \cdot \L_{k\nu}
                &\overset{(i)}{=}\alpha \cdot \lrp{f(r_{k\nu}) - \xi_{k\nu}} - \alpha \cdot \lrp{\sigma_{k\nu} + \phi_{k\nu}}\\
                & \overset{(ii)}{=} \alpha \cdot \alpha' \lrp{f(r_{k\nu}) - \xi_{k\nu}} - \alpha \cdot \lrp{\sigma_{k\nu} + \phi_{k\nu}}\\
                &\overset{(iii)}{\le} \alpha \cdot \alpha' \cdot e^{-\Cm \Ts} \lrp{f(r_{\tau_{k-1}}) - \xi_{\tau_{k-1}}} - \alpha \cdot \lrp{\sigma_{k\nu} + \phi_{k\nu}} + 5\beta\\
                &\overset{(iv)}{=} \alpha \cdot \alpha' \cdot e^{-\Cm (k\nu - \tau_{k-1})}\lrp{f(r_{\tau_{k-1}}) - \xi_{\tau_{k-1}}} - \alpha \cdot \lrp{\sigma_{k\nu} + \phi_{k\nu}} + 5\beta\\
                &\overset{(v)}{=} \alpha \cdot \alpha' \cdot e^{-\Cm \nu} e^{-\Cm ((k-1) \nu - \tau_{k-1})}\lrp{f(r_{\tau_{k-1}}) - \xi_{\tau_{k-1}}} - \alpha \cdot \lrp{\sigma_{k\nu} + \phi_{k\nu}}+5\beta\\
                &\overset{(vi)}{=} \alpha \cdot e^{-\Cm \nu} e^{-\Cm ((k-1) \nu - \tau_{k-1})}\lrp{f(r_{\tau_{k-1}}) - \xi_{\tau_{k-1}}} - \alpha \cdot \lrp{\sigma_{k\nu} + \phi_{k\nu}} + 5\beta,
                \numberthis \label{e:t:ept2}
                \end{align*}
                where $(i)$ is by definition of $\L_{k\nu}$, $(ii)$ by Eq.~\eqref{e:t:nwsdl}, $(iii)$ is by Lemma~\ref{l:muk=0_norm_contraction_3}, $(iv)$ is by the fact that $\alpha' = \ind{\Ts = k\nu - \tau_{k-1}}$, $(v)$ is by algebra and finally $(vi)$ is again by Eq.~\eqref{e:t:nwsdl}.
                
                By definition of $\sigma_t$ in Eq.~\eqref{d:sigma},
                \begin{align*}
                \alpha \cdot \sigma_{k\nu}
                &= \alpha \int_0^{k\nu} \mu_{\step{s}{\nu}}\cdot e^{-\Cm (k\nu-s)} \cdot 4 r_s ds\\
                &\overset{(i)}{=} \alpha \int_0^{(k-1)\nu} \mu_{\step{s}{\nu}}\cdot e^{-\Cm (k\nu-s)} \cdot 4 r_s ds\\
                &= \alpha e^{-\Cm \nu} \sigma_{(k-1)\nu},
                \numberthis \label{e:t:ept3}
                \end{align*}
                where $(i)$ is because $\alpha = 1$ implies that $\mu_{\step{s}{\nu}} = \mu_{k-1} = 0$ for all $s\in[(k-1)\nu, k\nu)$.
                
                Similarly, by the definition of $\phi_t$ in Eq.~\eqref{d:phi},
                \begin{align*}
                & \alpha \cdot \phi_{k\nu}\\
                &= \alpha \int_{0}^{k\nu} \mu_{\step{s}{\nu}} \cdot e^{-\Cm (k\nu-s)} f'\lrp{r_s} q'(\lrn{z_s + w_s}_2) \lin{\frac{z_s + w_s}{\lrn{z_s + w_s}_2}, 4\sqrt{\frac{\c}{L}} \lrp{\gamma_s \gamma_s^T dB_s + \frac{1}{2} \bar{\gamma}_s \bar{\gamma}_s^T dA_s}}\\
                &\overset{(i)}{=}\alpha \int_{0}^{(k-1)\nu} \mu_{\step{s}{\nu}} \cdot e^{-\Cm (k\nu-s)} f'\lrp{r_s}q'(\lrn{z_s + w_s}_2) \lin{\frac{z_s + w_s}{\lrn{z_s + w_s}_2}, 4\sqrt{\frac{\c}{L}} \lrp{\gamma_s \gamma_s^T dB_s + \frac{1}{2} \bar{\gamma}_s \bar{\gamma}_s^T dA_s}}\\
                &= \alpha e^{-\Cm \nu} \phi_{(k-1)\nu},
                \numberthis \label{e:t:ept4}
                \end{align*}
                where $(i)$ is again because $\alpha = 1$ implies that $\mu_{\step{s}{\nu}} = \mu_{k-1} = 0$ for all $s\in[(k-1)\nu, k\nu)$.
                
                Combining these results,
                \begin{align*}
                \alpha \L_{k\nu}
                &\overset{(i)}{\leq} \alpha \cdot e^{-\Cm \nu} e^{-\Cm ((k-1) \nu - \tau_{k-1})}\lrp{f(r_{\tau_{k-1}}) - \xi_{\tau_{k-1}}}\\
                &\qquad - \alpha \cdot \lrp{\sigma_{k\nu} + \phi_{k\nu}} + 5\beta\\
                &\overset{(ii)}{=} \alpha \cdot e^{-\Cm \nu} e^{-\Cm ((k-1) \nu - \tau_{k-1})}\lrp{f(r_{\tau_{k-1}}) - \xi_{\tau_{k-1}}}\\
                &\qquad - \alpha e^{-\Cm \nu} \cdot \lrp{\sigma_{(k-1)\nu} + \phi_{(k-1)\nu}} + 5\beta\\
                &\overset{(iii)}{=} \alpha \cdot e^{-\Cm \nu} \cdot \L_{(k-1) \nu} + 5\beta,
                \end{align*}
                where $(i)$ is by Eq.~\eqref{e:t:ept2} and $(ii)$ is by Eq.~\eqref{e:t:ept3} and Eq.~\eqref{e:t:ept4}. Inequality $(iii)$ is by the definition of $\L_t$ in Eq.~\eqref{d:L}, and because $\ind{\mu_{k-1} = 0}\L_{(k-1)\nu} = \ind{\mu_{k-1} = 0}\lrp{e^{-\Cm \lrp{(k-1)\nu - \tau_{k-1}}} f(r_{\tau_{k-1}}) - \lrp{\sigma_{(k-1)\nu} + \phi_{(k-1)\nu}}}$, as noted in the beginning of the proof.
            \end{proof}
            
            \begin{lemma}\label{l:mu00}
                For all positive integers $k$, with probability one,
                \begin{align*}
                 \ind{\mu_k = 0, \mu_{k-1} = 0}\cdot  \L_{k\nu} 
                &\leq \ind{\mu_k = 0, \mu_{k-1} = 0}\cdot  e^{-\Cm \nu} \L_{(k-1) \nu} + 5\beta.
                \end{align*}
            \end{lemma}
            \begin{proof}
                Define $\alpha_1,\alpha_2$ and $\alpha_3$ to be indicators for the following events:
                \begin{align*}
                \alpha_1 := \ind{\mu_k = 0, \mu_{k-1} = 0}, \alpha_2 := \ind{k\nu = \tau_k}\ \mbox{and}\ \alpha_3 := \ind{k\nu = \tau_{k-1} + \Ts}.
                \end{align*}
                By the definition of the Lyapunov function in Eq.~\eqref{d:L} we find that 
                \begin{align*}
                & \alpha_1 \cdot \L_{k\nu} = \alpha_1 \cdot \lrp{e^{-\Cm (k\nu - \tau_k)}\lrp{f(r_{\tau_k}) - \xi_{\tau_k}} - \lrp{\sigma_{k\nu} + \phi_{k\nu}}}, \quad \text{and, }\\
                & \alpha_1 \cdot \L_{(k-1) \nu} = \alpha_1 \cdot \lrp{e^{-\Cm ((k-1)\nu - \tau_{k-1})}\lrp{f(r_{\tau_{k-1}}) - \xi_{\tau_{k-1}}} - \lrp{\sigma_{(k-1)\nu} + \phi_{(k-1)\nu}}}.
                \numberthis \label{e:t:yrlmmas:1}
                \end{align*}

                We now consider two cases: when $k\nu = \tau_k$ and when $k\nu \neq \tau_k$ and prove the result in both of these cases.
                
                \textbf{Case 1: $k \nu = \tau_k$}\\
                From the definition of $\tau_t$ in Eq.~\eqref{d:taut}, we know that $k \nu = \tau_k \Rightarrow k\nu - \tau_{k-1} \geq \Ts$. Additionally, 
                $\mu_{k-1}= 0 \Rightarrow (k-1)\nu - \tau_{k-1} < \Ts$. By our choice of $\nu$; $\Ts/\nu$ is an integer (immediately below \eqref{d:ts}). Thus it must be that $k\nu = \tau_{k-1} + \Ts$.
                Hence we have shown that
                \begin{align*}
                 \alpha_1 \cdot \alpha_2 
                =& \alpha_1 \cdot \alpha_2 \cdot \alpha_3.
                \numberthis \label{e:t:mdw}
                \end{align*}
                Thus,
                \begin{align*}
                & \alpha_1 \cdot \alpha_2 \cdot \L_{k\nu}\\
                &\overset{(i)}{=} \alpha_1 \cdot \alpha_2 \cdot \alpha_3 \cdot \lrp{e^{-\Cm (k\nu - \tau_k)}\lrp{f(r_{\tau_k}) - \xi_{\tau_k}} - \lrp{\sigma_{\tau_k} + \phi_{\tau_k}}}\\
                &\overset{(ii)}{=} \alpha_1 \cdot \alpha_2 \cdot \alpha_3 \cdot \lrp{\lrp{f(r_{k\nu}) - \xi_{k\nu}} - \lrp{\sigma_{k\nu} + \phi_{k\nu}}}\\
                &\overset{(iii)}{\le} \alpha_1 \cdot \alpha_2 \cdot \alpha_3 \cdot \lrp{e^{-\Cm (k\nu - \Ts)} \cdot \lrp{f(r_{\tau_{k-1}}) - \xi_{\tau_{k-1}}} - \lrp{\sigma_{k\nu} + \phi_{k\nu}}} + 5\beta\\
                &\overset{(iv)}{=} \alpha_1 \cdot \alpha_2 \cdot \alpha_3 \cdot \lrp{e^{-\Cm (k\nu - \Ts)} \cdot \lrp{f(r_{\tau_{k-1}}) - \xi_{\tau_{k-1}}} - e^{-\Cm \nu}\lrp{\sigma_{(k-1)\nu} + \phi_{(k-1)\nu}}} + 5\beta\\
                &\overset{(v)}{=} \alpha_1 \cdot \alpha_2 \cdot \alpha_3 \cdot \lrp{e^{-\Cm \nu} \L\lrp{\theta_{(k-1)\nu}}} + 5\beta\\
                &\overset{(vi)}{=} \alpha_1 \cdot \alpha_2 \cdot \lrp{e^{-\Cm \nu} \L\lrp{\theta_{(k-1)\nu}}} + 5\beta,
                \end{align*}
                where $(i)$ is by Eq.~\eqref{e:t:mdw}, $(ii)$ is because $\alpha_2=1$ implies $\tau_k = k\nu$, $(iii)$ is by Lemma~\ref{l:muk=0_norm_contraction_3}. Inequality $(iv)$ is because $\alpha_1 = 1$ implies $\mu_{k-1} = 0$, we can thus verify from Eq.~\eqref{d:sigma} and Eq.~\eqref{d:phi} that $\alpha_1 \cdot \lrp{\sigma_{k\nu} + \phi_{k\nu}} = \alpha_1 \cdot e^{-\Cm \nu}\lrp{\sigma_{(k-1)\nu} + \phi_{(k-1)\nu}}$ (the detailed proof is identical to proof of Eq.~\eqref{e:t:ept3} and \eqref{e:t:ept4}, and is not repeated here). $(v)$ follows by our expression for $\L_{(k-1) \nu}$ in Eq.~\eqref{e:t:yrlmmas:1} and $(vi)$ is again by Eq.~\eqref{e:t:mdw}.
                
                \textbf{Case 2: $k \nu \neq \tau_k$}\\
                In this case, by the definition of $\tau_t$ (in Eq.~\eqref{d:taut}) that $\tau_k = \tau_{k-1}$. Thus,
                \begin{align*}
                & \alpha_1 \cdot (1-\alpha_2) \cdot \L_{k\nu} \\
                &\overset{(i)}{=} \alpha_1 \cdot (1-\alpha_2) \cdot e^{-\Cm (k\nu - \tau_k)} 
                \lrp{f(r_{\tau_{k}}) - \xi_{\tau_{k}}} - \alpha_1 \cdot (1-\alpha_2) \cdot \lrp{\sigma_{k\nu} + \phi_{k\nu}}\\
                &\overset{(ii)}{=} \alpha_1 \cdot (1-\alpha_2) \cdot e^{-\Cm (k\nu - \tau_{k-1})} 
                \lrp{f(r_{\tau_{k-1}}) - \xi_{\tau_{k-1}}} - \alpha_1 \cdot (1-\alpha_2) \cdot \lrp{\sigma_{k\nu} + \phi_{k\nu}}\\
                &\overset{(iii)}{=} \alpha_1 \cdot (1-\alpha_2) \cdot e^{-\Cm (k\nu - \tau_{k-1})} 
                \lrp{f(r_{\tau_{k-1}}) - \xi_{\tau_{k-1}}} - \alpha_1 \cdot (1-\alpha_2) \cdot e^{-\Cm \nu} \lrp{\sigma_{(k-1)\nu} + \phi_{(k-1)\nu}}\\
                &\overset{(iv)}{=} \alpha_1 \cdot (1-\alpha_2) \cdot e^{-\Cm \nu} \L_{(k-1) \nu},
                \end{align*}
                where $(i)$ is by the expression for $\L_{k\nu}$ in Eq.~\eqref{e:t:yrlmmas:1}, $(ii)$ is because $\tau_k = \tau_{k-1}$. Inequality $(iii)$ is because $\alpha_1 \cdot (\sigma_{k\nu} + \phi_{k\nu}) = \alpha_1 \cdot e^{-\Cm \nu}(\sigma_{(k-1)\nu} + \phi_{(k-1)\nu})$. The proof of this fact is identical to proof of inequalities Eqs.~\eqref{e:t:ept3} and \eqref{e:t:ept4}, and is not repeated here. Finally $(iv)$ is by pulling out a factor of $e^{-\Cm \nu}$, and then using the equality in Eq.~\eqref{e:t:yrlmmas:1}.

                Therefore, summing the two cases, we get our conclusion that
                \begin{align*}
                 \ind{\mu_k = 0, \mu_{k-1} = 0}\cdot \L_{k\nu} 
                \leq \ind{\mu_k = 0, \mu_{k-1} = 0}\cdot e^{-\Cm \nu} \L_{(k-1) \nu} + 5\beta.
                \end{align*}
            \end{proof}
            
            \begin{lemma}\label{l:mu11}
                For all positive integers $k$, with probability 1,
                \begin{align*}
                 \ind{\mu_k = 1, \mu_{k-1}= 1} \cdot \L_{k\nu} 
                \leq \ind{\mu_k = 1, \mu_{k-1}= 1} \cdot e^{-\Cm \nu} \L_{(k-1) \nu} + 5\beta \nu.
                \end{align*}
            \end{lemma}
            \begin{proof}
                Let $\alpha$ denote the indicator of the following event, $\alpha := \ind{\mu_k = 1, \mu_{k-1}= 1}$.
                By the definition of our Lyapunov function  (see Eq.~\eqref{d:L}) that
                \begin{align*}
                & \alpha \cdot \L_{k\nu} = \alpha \cdot \lrp{\lrp{f(r_{k\nu}) - \xi_{k\nu}} - \lrp{\sigma_{k\nu} + \phi_{k\nu}}}, \quad \text{and,}\\
                & \alpha \cdot \L_{(k-1) \nu} = \alpha \cdot \lrp{\lrp{f(r_{(k-1)\nu}) - \xi_{(k-1)\nu}} - \lrp{\sigma_{(k-1)\nu} + \phi_{(k-1)\nu}}}.
                \numberthis \label{e:t:csegra:1}
                \end{align*}
                Thus we have,
                \begin{align*}
                \alpha \cdot \L_{k\nu}
                &\overset{(i)}{=} \alpha \cdot \lrp{\lrp{f(r_{k\nu}) - \xi_{k\nu}} - \lrp{\sigma_{k\nu} + \phi_{k\nu}}}\\
                &\overset{(ii)}{=} \alpha \cdot \lrp{\mu_k \lrp{f(r_{k\nu}) - \xi_{k\nu}} - \lrp{\sigma_{k\nu} + \phi_{k\nu}}}\\
                &\overset{(iii)}{\le} \alpha \cdot  \lrp{e^{-\Cm \nu}  \mu_k \cdot \lrp{f(r_{(k-1)\nu}) - \xi_{(k-1)\nu}} - \lrp{\sigma_{(k-1)\nu} + \phi_{(k-1)\nu}}} + 5\beta \nu\\
                &\overset{(iv)}{=} \alpha \cdot  \lrp{e^{-\Cm \nu} \cdot \lrp{f(r_{(k-1)\nu}) - \xi_{(k-1)\nu}} - \lrp{\sigma_{(k-1)\nu} + \phi_{(k-1)\nu}}} + 5\beta \nu\\
                &\overset{(v)}{=} \alpha \cdot e^{-\Cm \nu} \L_{(k-1)\nu},
                \end{align*}
                where $(i)$ is by Eq.~\eqref{e:t:csegra:1}, $(ii)$ is because $\alpha = \alpha \cdot \mu_k$, $(iii)$ is by Lemma~\ref{l:muk=1}, $(iv)$ is again because $\alpha = \alpha \cdot \mu_k$ and $(v)$ is again by Eq.~\eqref{e:t:csegra:1}.
            \end{proof}
            
            \begin{lemma}\label{l:mu10}
                For all positive integers $k$, with probability 1,
                \begin{align*}
                 \ind{\mu_k = 0, \mu_{k-1}= 1} \cdot \L_{k\nu} 
                \leq& \ind{\mu_k = 0, \mu_{k-1}= 1} \cdot e^{-\Cm \nu} \L_{(k-1) \nu} + 5\beta \nu.
                \end{align*}
            \end{lemma}
            \begin{proof}
                Let $\alpha := \ind{\mu_k = 0, \mu_{k-1}= 1}$. We can verify using the definition of the Lyapunov function in Eq.~\eqref{d:L} that:
                \begin{align*}
                 \alpha \cdot \L_{k\nu} &= \alpha \cdot \lrp{e^{-\Cm (k\nu - \tau_{k})}\lrp{f(r_{\tau_{k}}) - \xi_{\tau_{k}}} - \lrp{ \sigma_{k\nu} + \phi_{k\nu}}} \quad \text{and,}\\
                 \alpha \cdot \L_{(k-1) \nu} &= \alpha \cdot \lrp{\lrp{f(r_{(k-1)\nu}) - \xi_{(k-1)\nu}} - \lrp{\sigma_{(k-1)\nu} + \phi_{(k-1)\nu}}}.
                \numberthis \label{e:t:gfdmvn}
                \end{align*} 
                
                Additionally, we can verify from Eq.~\eqref{d:mut} that $\mu_k = 0$ implies that $k\nu - \Ts< \tau_k $ and that $\mu_{k-1} = 1$ implies thta $(k-1) \nu - \Ts \geq \tau_{k-1} $. Putting this together, we get
                \begin{align*}
                \tau_k > k\nu - \Ts > (k-1)\nu - \Ts \geq \tau_{k-1}.
                \end{align*}
                Thus $\tau_k > \tau_{k-1}$. From the definition of $\mu_t$ (in Eq.~\eqref{d:mut}), we see that $\tau_k$ is either equal to $\tau_{k-1}$ or is equal to $k\nu$, so that it must be that 
                \begin{align*}
                \tau_k = k\nu,
                \end{align*}
                when $\alpha = 1$. In particular, this implies that \begin{align*}
                \alpha \cdot \L_{k\nu} 
                &\overset{(i)}{=} \alpha \cdot \lrp{e^{-\Cm (k\nu - \tau_{k})}\lrp{f(r_{\tau_{k}}) - \xi_{\tau_k}} - \lrp{ \sigma_{k\nu} + \phi_{k\nu}}}\\
                &\overset{(ii)}{=} \alpha \cdot \lrp{\mu_k \cdot \lrp{f(r_{k\nu}) - \xi_{k\nu}} - \lrp{ \sigma_{k\nu} + \phi_{k\nu}}}\\
                &\overset{(iii)}{\le} \alpha \cdot  \lrp{e^{-\Cm \nu}  \mu_k \cdot \lrp{f(r_{(k-1)\nu}) - \xi_{(k-1)\nu}} - \lrp{\sigma_{(k-1)\nu} + \phi_{(k-1)\nu}}} + 5\beta \nu\\
                &\overset{(iv)}{\le} \alpha \cdot  \lrp{e^{-\Cm \nu} \cdot \lrp{f(r_{(k-1)\nu}) - \xi_{(k-1)\nu}} - \lrp{\sigma_{(k-1)\nu} + \phi_{(k-1)\nu}}} + 5\beta \nu\\
                &\overset{(v)}{=} \alpha \cdot  e^{-\Cm \nu}\L_{(k-1)\nu} + 5\beta \nu,
                \end{align*}
                where $(i)$ is by Eq.~\eqref{e:t:gfdmvn}, $(ii)$ is by $\alpha \cdot \mu_k = \alpha$ and because $\alpha = \alpha \cdot \ind{\tau_k = k\nu}$, $(iii)$ is by Lemma~\ref{l:muk=1}, $(iv)$ is again by $\alpha \cdot \mu_k = \alpha$ and finally $(v)$ is by Eq.~\eqref{e:t:gfdmvn}.
            \end{proof}

        \end{subsection}
    \end{section}

    \begin{section}{Properties of $f$} \label{s:fproperties}
        \label{s:onf}
        \begin{lemma} \label{lem:fpropertiesall} 
            Assume that $e^{72LR^2} \geq 2$. The function $f$ defined in \eqrefmike{d:f} has the following properties.
            \label{l:fsandwich}
            \begin{enumerate}[label=(F{\arabic*})]
                \item $f(0) = 0$, $f'(0)=1$ \label{fprop:1}.
                \item $\frac{1}{2} e^{- 2 \Cf  \Rf^2} \leq \frac{1}{2} \psi(r) \leq f'(r)\leq 1$. \label{fprop:2}
                \item $\frac{1}{2}e^{-2 \Cf  \Rf^2} r \leq \frac{1}{2} \Psi(r) \le f(r) \le \Psi(r)\le r$. \label{fprop:3}
                \item For all $0 < r \leq \Rf$, $f''(r) + \Cf  rf'(r) \leq - \frac{e^{-2 \Cf  \Rf^2}}{4 \Rf^2} f(r)$
                \label{fprop:4}
                \item For all $r> 0$, $f''$ is defined, $f''(r) \leq 0$, and $f''(r) = 0$ when
                $r>2 \Rf$. \label{fprop:5}
                \item If $2\Cf\Rf^2\ge\ln 2$, for any $0.5<s<1$, $f(sr) \leq \exp\lrp{-\frac{1-s}{4}e^{-2 \Cf  \Rf^2}}f(r)$. \label{fprop:6}
                \item For $r>0$, $\lrabs{f''(r)} \leq 4\Cf\Rf + \frac{4}{\Rf}$ \label{fprop:7}
            \end{enumerate}
        \end{lemma}
        \begin{proof}
            We refer to definitions of the functions $\psi, \Psi, g$ in Eq.~\eqref{d:psietal} and the definition of $f$ in Eq.~\eqref{d:f}.
            \begin{description}
                \item[\ref{fprop:1}] $f(0) = 0$ and $f'(0) = 1$ by the definition of $f$ and $\psi$.
                
                \item[\ref{fprop:2},\ref{fprop:3}] are
                verified from the definitions, noting that
                $\frac{1}{2} \leq g(r) \leq 1$ and
                $e^{- 2 \Cf \Rf^2} \le \psi(2\Rf)\le\psi(r)\le\psi(0)$.
                
                \item[\ref{fprop:4}] To prove this property first we observe
                that $f'(r) = \psi(r) g(r)$ so
                $$f''(r) = \psi'(r) g(r) + \psi(r)g'(r).$$
                By the definition of $\psi$, $\psi'(r) = -2\Cf  r \psi(r)$ if
                $r<\Rf$, thus
                \begin{align*}
                f''(r) + 2\Cf rf'(r) 
                &= -2\Cf  r \psi(r)g(r) + \psi(r)g'(r) + 2\Cf  rf'(r)\\
                &= \psi(r) g'(r)\\ 
                &= -\frac{1}{2} \frac{h(r) \Psi(r)}
                {\int_0^{\infty} h(s) \frac{\Psi(s)}{\psi(s)} ds}\\
                &\overset{(i)}{\leq} -\frac{1}{2} \frac{f(r)}{\int_0^{\infty}h(s) \frac{\Psi(s)}{\psi(s)} ds}\\
                &\overset{(ii)}{\leq} -\frac{e^{-2 \Cf  \Rf^2}}{4 \Rf^2} f(r),
                \end{align*}
                where $(i)$ is because $f(r) \leq \Psi(r)$ and $h(r) = 1$ for $r \leq \Rf$.
                
                
                $(ii)$ is because $f(r) \geq 0$ and 
                \begin{align*}
                \int_0^\infty h(s) \frac{\Psi(s)}{\psi(s)} ds =  \int_0^{2\Rf} h(s) \frac{\Psi(s)}{\psi(s)} ds                         
                \leq \int_0^{2\Rf} \frac{2s}{e^{- 2 \Cf \Rf^2}} ds
                \leq 4 \Rf^2 e^{2\Cf \Rf^2}.
                \end{align*}
                The first inequality above is by \ref{fprop:2}, \ref{fprop:3} and the definition of $h(s)$.
                
                \item[\ref{fprop:5}] $f''(r) \leq 0$ follows from its expression $f''(r) = \psi'(r) g(r) + \psi(r)g'(r)$, and the fact that $\psi(r) \geq 0$ from 
                \ref{fprop:2}, $g(r)\geq 1/2$, $g'(r) \leq 0$ and $\psi'(r) \leq 0$ for all $r$. For $r>2 \Rf$,
                $\psi'(r)=g'(r)=0$, so in that case
                $f''(r) = \psi'(r) g(r) + \psi(r)g'(r)=0$.
                \item[\ref{fprop:6}] 
                For any $0<c<1$,
                \begin{align*}
                f((1+c)r) 
                = & f(r) + \int_r^{(1+c)r} f'(s) ds
                \geq  f(r) + cr \cdot \frac{1}{2} e^{- 2 \Cf  \Rf^2}
                \geq  \left(1+\frac{c}{2}e^{- 2 \Cf  \Rf^2}\right) f(r),
                \end{align*}
                where the first inequality follows from \ref{fprop:2}, and
                the second inequality follows from \ref{fprop:3}.
                Under the assumption that $e^{- 2 \Cf  \Rf^2}\leq \frac{1}{2}$,
                and using the inequality $1+x \geq e^{x/2}$ for all
                $x\in[0,1/2]$, we get
                $1+(c/2)e^{-2 \Cf\Rf^2}\ge e^{(c/4)e^{-2 \Cf\Rf^2}}$.
                
                Thus, for any $s \in (1/2,1)$, let $r' := s r$, so that $r = \frac{1}{s} r' = \lrp{1 +\lrp{ \frac{1}{s} -1 }} r'$
                Applying the above with $c = \frac{1}{s} -1$, we get
                \begin{align*}
                f(s r) = f(r') 
                &\leq \frac{1}{1 + \frac{c}{2} \exp\lrp{-2 \Cf \Rf^2}} f((1+ c) r') \\
                &= \frac{1}{1 + \frac{c}{2} \exp\lrp{-2 \Cf \Rf^2}} f(r)\\
                \leq& \exp\lrp{- \frac{c}{4} e^{-2 \Cf \Rf^2}} f(r)\\
                &= \exp\lrp{- \frac{1/s - 1}{4} e^{-2 \Cf \Rf^2}} f(r)
                \leq \exp\lrp{- \frac{1-s}{4} e^{-2 \Cf \Rf^2}} f(r).
                \end{align*}
                where we use the fact that $-\frac{1-s}{s} \leq -(1-s)$.
                
                \item[\ref{fprop:7}]
                Recall that
                \begin{align*}
                f''(r) = \psi'(r) g(r) + \psi(r) g'(r)
                \end{align*}
                Thus
                \begin{align*}
                \lrabs{f''(r)}
                &\leq \lrabs{\psi'(r) g(r)} + \lrabs{\psi(r) g'(r)}\\
                &\leq 2\Cf r h(r) + \lrabs{\psi(r) g'(r)}
                \end{align*}
                From our definition of $h(r)$, we know that $rh(r) \leq 2\Rf$. In addition, since $\psi(r)$ is monotonically decreasing, $\Psi(r) = \int_0^r \psi(s) ds \geq r \psi(r)$, so that
                \begin{align*}
                \numberthis \label{e:l:psi(r)/psi(r)}
                \frac{\Psi(r)}{\psi(r)} \geq r.
                \end{align*}
                Thus $\Psi(r)/r \geq r$ for all $r$. On the other hand, using the fact that $\psi(s) \leq 1$, 
                \begin{align*}
                \numberthis \label{e:l:psi(r)_upperbound}
                \Psi(r) = \int_0^r \psi(s)ds \leq r.
                \end{align*}
                
                Combining the previous expressions,
                \begin{align*}
                \lrabs{\psi(r) \nu'(r)}
                &= \lrabs{ \frac{1}{2}\frac{h(r)\Psi(r)}{\int_0^{4\Rf}\frac{\mu(s)\Psi(s)}{\psi(s)} ds}}\\
                &\leq \lrabs{\frac{1}{2}\frac{2\Rf}{\int_0^{\Rf}\frac{\Psi(s)}{\psi(s)} ds}}\\
                &\leq \lrabs{\frac{1}{2} \frac{2\Rf}{\int_0^{\Rf} s ds}}\\
                &\leq \frac{4}{\Rf},
                \end{align*}
                where the first inequality is by the definition of $h(r)=1$ for $r\leq \Rf$ and $h(r) =0 $ for $r\geq 2\Rf$, and the second-to-last inequality is by \eqref{e:l:psi(r)/psi(r)}. 
                
                Put together, we get
                \begin{align*}
                \lrabs{f''(r)} \leq 4\Cf\Rf + \frac{4}{\Rf}.
                \end{align*}
            \end{description}
        \end{proof}
    \end{section}
    
    \begin{section}{Bounding moments}\label{s:bounding_energy}
       To bound the discretization error it is necessary to bound the moments of the random variables $x_t, u_t$ and $y_t, v_t$. The main results of this section are Lemma~\ref{l:energy_bound_all_time} (which bounds the moments of $x_t$ and $u_t$) and Lemma~\ref{l:energy_bound_all_time_y} (which bounds the moments of $y_t$ and $v_t$).
        \begin{lemma}\label{l:energy_bound_all_time}
            For $\delta \leq 2^{-10} \c$, and for all $t\ge0$,
            $$\E{\lrn{x_t}_2^{8} + \lrn{x_t + u_t}_2^{8}}\leq 2^{70} \lrp{R^2 + \frac{d}{m}}^4.$$
        \end{lemma}
        
        \begin{lemma}\label{l:energy_bound_all_time_y}
            For all $t\ge0$,
            \begin{align*}
            \E{\lrn{y_t}_2^{8} + \lrn{y_t + v_t}_2^{8}}\leq 2^{66} \lrp{R^2 + \frac{d}{m}}^4.
            \end{align*}
        \end{lemma}
        \subsection{Proof of Lemma~\ref{l:energy_bound_all_time}}
            Let us consider the Lyapunov function $l(x_t, u_t):= \lrp{\lrn{x_t}_2^2 + \lrn{x_t + u_t}_2^2 - 4R^2}_+^4$. 
            
            By calculating the derivaties of $l$ we can verify that:
            \begin{align*}
            & \nabla_x l(x_t, u_t) = 8 l(x_t,u_t)^{3/4} \lrp{x_t}\\
            & \nabla_u l(x_t, u_t) = 8 l(x_t,u_t)^{3/4} \lrp{x_t + u_t}\\
            & \nabla_u^2 l(x_t,u_t) = 8 l(x_t,u_t)^{3/4} I + 24 l(x_t,u_t)^{2/4} \lrp{x_t + u_t} \lrp{x_t + u_t}^T.
            \end{align*}
            The following are two useful inequalities which we will use in this proof:
            \begin{align*}
            & \lrn{x}_2^2 + \lrn{x + u}_2^2 \leq l(x,u)^{1/4} + 4R^2 \\
            & \lrn{x}_2^2 + \lrn{x + u}_2^2 \geq l(x,u)^{1/4}.
            \numberthis \label{e:t:jtelgk:1}
            \end{align*}
            
            Recall from the dynamics defined in Eq.~\eqref{d:xt} and Eq.~\eqref{d:ut} that
            \begin{align*}
            d x_t =& u_t dt\\
            d u_t =& -2 u_t - \frac{\c}{L}\nabla U(x_{\step{t}{\delta} \delta}) dt + 2\sqrt{\frac{\c}{L}} dB_t.
            \end{align*}
            Thus by studying the evolution of the Lyapunov function $l(x_t,u_t)$ we have:
            \begin{align*}
             \ddt \E{l(x_t,u_t)}
            &= \E{8l(x_t, u_t)^{3/4} \lrp{\lin{x_t, u_t} + \lin{x_t + u_t, - u_t - \frac{\c}{L}\nabla U(x_{\step{t}{\delta} \delta})}}}\\
            & \quad + \E{\frac{16\c}{L}\lrp{l(x_t, u_t)^{3/4}d + 3 l(x_t,u_t)^{2/4} \lrn{x_t + u_t}_2^2 } }\\
            &= \E{\underbrace{8l(x_t, u_t)^{3/4} \lrp{\lin{x_t, u_t} + \lin{x_t + u_t, - u_t - \frac{\c}{L}\nabla U(x_{t})}}}_{=:\spadesuit}}\\
            &\qquad \qquad + \E{\underbrace{8\cdot \frac{\c}{L} \cdot l(x_t, u_t)^{3/4} \lrp{\lin{x_t + u_t,\nabla U(x_{t}) - \nabla U(x_{\step{t}{\delta} \delta})}}}_{=:\heartsuit}}\\
            & \qquad \qquad + \E{\underbrace{\frac{16\c}{L}\lrp{l(x_t, u_t)^{3/4}d + 3 l(x_t,u_t)^{2/4} \lrn{x_t + u_t}_2^2 }}_{=:\clubsuit}}.
            \end{align*}
            We will bound the three terms separately. We begin by bounding $\spadesuit$:
            \begin{align*}
            \spadesuit &= 8l(x_t, u_t)^{3/4} \lrp{\lin{x_t, u_t} + \lin{x_t + u_t, - u_t - \frac{\c}{L}\nabla U(x_{t})}}\\
            &\overset{(i)}{\leq} - \c^2 l(x_t, u_t)^{3/4} \lrp{\lrn{x_t}_2^2 + \lrn{x_t + u_t}_2^2}\\
            &\overset{(ii)}{\leq} - \c^2 l(x_t, u_t),
            \end{align*}
            where $(i)$ is by invoking Lemma~\ref{l:xu}, and $(ii)$ is by Eq.~\eqref{e:t:jtelgk:1}. Next consider the term $\heartsuit$:
            \begin{align*}
            \heartsuit &= 8\cdot \frac{\c}{L} \cdot l(x_t, u_t)^{3/4} \lrp{\lin{x_t + u_t,\nabla U(x_{t}) - \nabla U(x_{\step{t}{\delta} \delta})}}\\
            &\overset{(i)}{\leq} 8\c l(x_t, u_t)^{3/4} \lrn{x_t + u_t}_2 \lrn{x_t - x_{\step{t}{\delta} \delta}}_2\\
            &\overset{(ii)}{\leq}  8\c l(x_t, u_t)^{3/4} \lrp{l(x_t,u_t)^{1/8} + 2R} \lrn{x_t - x_{\step{t}{\delta} \delta}}_2\\
            &\overset{(iii)}{\leq} 8\c l(x_t, u_t)^{7/8} \lrn{x_t - x_{\step{t}{\delta} \delta}}_2 + 16\c l(x_t, u_t)^{3/4} R \lrn{x_t - x_{\step{t}{\delta} \delta}}_2\\
            &\overset{(iv)}{\leq} \frac{\c^2}{8} l(x_t, u_t) + \frac{2^{32}}{\c^6} \lrp{\lrn{x_t - x_{\step{t}{\delta} \delta}}_2^8} + \frac{\c^2}{8}l(x_t, u_t) + \frac{2^{28}}{\c^2}\lrp{R^4 \lrn{x_t - x_{\step{t}{\delta} \delta}}_2^4}\\
            &\overset{(v)}{\leq} \frac{\c^2}{4} l(x_t, u_t) + \frac{2^{32}}{\c^6} \lrp{\lrn{x_t - x_{\step{t}{\delta} \delta}}_2^8}+ {2^{28}\c^2}{R^8} +  \frac{2^{28}}{\c^6}\lrn{x_t - x_{\step{t}{\delta} \delta}}_2^8\\
            &\overset{(vi)}{\leq} \frac{\c^2}{4} l(x_t, u_t) + {2^{28}\c^2}{R^8} +  \frac{2^{33}}{\c^6}\lrn{x_t - x_{\step{t}{\delta} \delta}}_2^8\\
            &\overset{(vii)}{\leq} \frac{\c^2}{4} l(x_t, u_t) + {2^{28}\c^2}{R^8} +  \frac{2^{33}}{\c^6}\lrn{\int_{\step{t}{\delta} \delta}^t u_s ds}_2^8\\
            &\overset{(viii)}{\leq} \frac{\c^2}{4} l(x_t, u_t) + {2^{28}\c^2}{R^8} +  \frac{2^{33}}{\c^6} \lrp{ \lrp{t - \step{t}{\delta} \delta}^7 \int_{\step{t}{\delta} \delta}^t\lrn{ u_s }_2^8 ds}\\
            &\overset{(ix)}{\le} \frac{\c^2}{4} l(x_t, u_t) + {2^{28}\c^2}{R^8} +  \frac{2^{33}}{\c^6}\lrp{\delta^7 \int_{\step{t}{\delta} \delta}^t\lrn{ u_s }_2^8 ds},
            \end{align*}
            where $(i)$ is by Cauchy-Schwarz and Assumption~\ref{ass:smoothness}, $(ii)$ is by Eq.~\eqref{e:t:jtelgk:1}, $(iii)$ is again by Eq.~\eqref{e:t:jtelgk:1}, $(iv)$ is by Young's inequality, $(v)$ is again by Young's inequality, $(vi)$ follows by an algebraic manipulation, $(vii)$ is by the dynamics defined in Eq.~\eqref{d:xt}, $(viii)$ is by Jensen's inequality and finally $(ix)$ is because $t - \step{t}{\delta}\delta \leq \delta$.
            Also:
            \begin{align*}
            \clubsuit &= \frac{16\c}{L}\lrp{l(x_t, u_t)^{3/4}d + 3 l(x_t,u_t)^{2/4} \lrn{x_t + u_t}_2^2 }\\
            &\overset{(i)}{\le} \frac{16\c}{L}\lrp{l(x_t, u_t)^{3/4}d + 3 l(x_t,u_t)^{3/4} + 12 l(x_t,u_t)^{2/4} R^2}\\
            &\overset{(ii)}{\le} \frac{\c^2}{16} l(x_t, u_t) + \frac{2^{28}}{\c^2 L^4} d^4 + \frac{\c^2}{16}l(x_t, u_t) + \frac{2^{36}}{\c^2 L^4} + \frac{\c^2}{16} l(x_t, u_t) + \frac{2^{16} \c^2 R^4}{L^2}\\
            &\overset{(iii)}{\le} \frac{\c^2}{4}l(x_t, u_t)  + 2^{29} \c^2 \lrp{\frac{d^4}{m^4} + \frac{R^4}{L^2}}\\
            &\overset{(iv)}{\le} \frac{\c^2}{4}l(x_t, u_t)  + 2^{30} \c^2 \lrp{\frac{d^4}{m^4} + R^8},
            \end{align*}
where $(i)$ is by Eq.~\eqref{e:t:jtelgk:1}, $(ii)$ is by Young's inequality, $(iii)$ follows by definition of $\c$ in Eq.~\eqref{d:c} and  $(iv)$ is by Young's inequality, and because $m\leq L$.
            
Putting together the upper bounds on $\spadesuit, \heartsuit, \clubsuit$:
            \begin{align*}
             &\frac{d}{dt} \E{l (x_t, u_t)} 
            = \spadesuit + \heartsuit + \clubsuit\\
            & \leq  \E{- \c^2 l(x_t, u_t) + {2^{28}\c^2}{R^8} +  \frac{2^{33}}{\c^6}\lrp{\delta^7 \int_{\step{t}{\delta} \delta}^t\lrn{ u_s }_2^8 ds} + \frac{\c^2}{4}l(x_t, u_t)  + 2^{30} \c^2 \lrp{\frac{d^4}{m^4} + R^8}}\\
            &\leq - \frac{\c^2}{2} \E{l(x_t, u_t)} + {2^{33}}{\c^2} \lrp{\frac{d^4}{m^4} + R^8} + \frac{2^{33}}{\c^6} \delta^7 \int_{\step{t}{\delta} \delta}^t\E{\lrn{ u_s }_2^8} ds\\
            & \overset{(i)}{\leq} - \frac{\c^2}{2} \E{l(x_t, u_t)} + {2^{33}}{\c^2} \lrp{\frac{d^4}{m^4} + R^8} + \frac{2^{33}}{\c^6} \delta^8 \lrp{1.1 \E{\lrp{\lrn{x_{\step{t}{\delta} \delta}}_2^8 + \lrn{u_{\step{t}{\delta} \delta}}_2^8}} + 2\lrp{ \frac{d}{m}}^4}\\
            &\overset{(ii)}{\leq} - \frac{\c^2}{2} \E{l(x_t, u_t)} + {2^{33}}{\c^2} \lrp{\frac{d^4}{m^4} + R^8} + \frac{\c^2}{8} \lrp{\E{l\lrp{x_{\step{t}{\delta} \delta}, u_{\step{t}{\delta} \delta}}} + R^8 + \lrp{ \frac{d}{m}}^4}\\
            &\leq - \frac{\c^2}{2} \E{l(x_t, u_t)} + {2^{34}}{\c^2} \lrp{\frac{d^4}{m^4} + R^8} + \frac{\c^2}{8} \E{l\lrp{x_{\step{t}{\delta} \delta}, u_{\step{t}{\delta} \delta}}},
            \numberthis \label{e:t:nmvkfd:1}
            \end{align*}
            where $(i)$ is by Lemma~\ref{l:energy_bound_uniform}, and $(ii)$ is by Eq.~\eqref{e:t:jtelgk:1} and Eq.~\eqref{d:c} along with some algebra.
            
            Consider an arbitrary positive interger $k$. By Gr\"onwall's Lemma applied over $s\in [k\delta, (k+1)\delta)$,
            \begin{align*}
            &\E{l(x_{(k+1)\delta}, u_{(k+1)\delta}} \\
            &\qquad \leq e^{-\frac{\c^2}{2} \delta} \E{l(x_{k\delta}, u_{k\delta})}+ \delta \cdot \lrp{{2^{34}}{\c^2} \lrp{\frac{d^4}{m^4} + R^8} + \frac{\c^2}{8} \E{l\lrp{x_{\step{t}{\delta} \delta}, u_{\step{t}{\delta} \delta}}}}\\
            & \qquad\overset{(i)}{\leq} \lrp{1- \frac{\c^2\delta}{4}} \E{l(x_{k\delta}, u_{k\delta})}+ \delta \cdot \lrp{{2^{34}}{\c^2} \lrp{\frac{d^4}{m^4} + R^8} + \frac{\c^2}{8} \E{l\lrp{x_{\step{t}{\delta} \delta}, u_{\step{t}{\delta} \delta}}}}\\
            & \qquad\overset{(ii)}{\leq} e^{-\frac{\c^2 \delta}{8}}\E{l(x_{k\delta}, u_{k\delta})} + 2^{34} \c^2\delta \lrp{2^{34} \lrp{\frac{d^4}{m^4} + R^8}},
            \end{align*}
            where $(i)$ and $(ii)$ use the fact that $\c^2 \delta \leq \frac{1}{10}$, along with $1-a \leq e^{-a} \leq 1 - \frac{a}{2}$ for $\lrabs{a} \leq \frac{1}{10}$.
            
            Applying the above recursively, using the geometric sum, and Eq.~\eqref{d:x0}, we show that for all positive integers $k$,
            \begin{align*}
            \E{l(x_{k\delta}, u_{k\delta})} \leq 2^{38} \lrp{\frac{d^4}{m^4} + R^8}.
            \end{align*}
            For an arbitrary $t\ge 0 $, we can similarly verify using the above result, Eq.~\eqref{e:t:nmvkfd:1}, and Gr\"onwall's Lemma that
            \begin{align*}
            \E{l(x_t, u_t)} \leq 2^{39} \lrp{\frac{d^4}{m^4} + R^8}.
            \end{align*}
            
            This completes the proof of the lemma.

            We now state and prove some auxillary lemmas that were useful in the proof above.

        \begin{lemma}\label{l:energy_bound_uniform}
            Assume that $\delta \leq \frac{1}{1000}$. Then for all $t\ge 0$,
            \begin{align*}
            \E{\lrn{x_{t}}_2^8 + \lrn{u_{t}}_2^8} \leq 1.1 \E{\lrp{\lrn{x_{\step{t}{\delta} \delta}}_2^8 + \lrn{u_{\step{t}{\delta} \delta}}_2^8}} + 2\lrp{ \frac{d}{m}}^4.
            \end{align*}
        \end{lemma}
        
        \begin{proof}
            From the stochastic dynamics defined in  Eq.~\eqref{d:xt}, Eq.~\eqref{d:ut}, Eq.~\eqref{d:yt} and Eq.~\eqref{d:vt}, we can verify that 
            \begin{align*}
             \ddt \E{ \lrp{\lrn{x_t}_2^8 + \lrn{u_t}_2^8}}
            \overset{(i)}{=}& \E{8\lrn{x_t}_2^6\lin{x_t, u_t} + 8\lrn{u_t}_2^6\lin{u_t, -2 u_t - \frac{\c}{L} \nabla U(x_{\step{t}{\delta} \delta})} }\\
            &\qquad\qquad + \E{\frac{8\c d}{L}\lrn{u_t}_2^6  + \frac{48 \c}{L}\lrn{u_t}_2^6 }\\
            \overset{(ii)}{\le}& 8 \E{\lrn{x_t}_2^8 + \lrn{u_t}_2^8 + \lrn{u_t}_2^8 + \c^8 \lrn{x_{\step{t}{\delta} \delta}}_2^8} + \E{\frac{d}{m} \lrn{u_t}_2^6}\\
            \overset{(iii)}{\le}& 64 \E{\lrn{x_t}_2^8 + \lrn{u_t}_2^8} +  \E{\lrn{x_{\step{t}{\delta} \delta}}_2^8} + \lrp{\frac{d}{m}}^4,
            \end{align*}
            where $(i)$ is by $\Ito$'s Lemma, $(ii)$ is by Assumption~\ref{ass:smoothness}, Young's inequality and by the definition of $\c$ in Eq.~\eqref{d:c}, and $(iii)$ is again by Young's inequality and definition of $\c$.
            
            Consider an arbitrary $t\ge 0$, and let $k:= \step{t}{\delta}$. Then for all $s \in [k\delta, (k+1)\delta)$, we have:
            \begin{align*}
             &\E{ \lrp{\lrn{x_t}_2^8 + \lrn{u_t}_2^8}}\\
            &\qquad \leq e^{64 (s-k\delta)}  \E{ \lrp{\lrn{x_{\step{t}{\delta} \delta}}_2^8 + \lrn{u_{\step{t}{\delta} \delta}}_2^8}} + \lrp{e^{64 (s-k\delta)}-1}\lrp{ \E{\lrn{x_{\step{t}{\delta} \delta}}_2^8} + \lrp{\frac{d}{m}}^4}\\
             &\qquad \leq (1+128 \delta) \E{ \lrp{\lrn{x_{\step{t}{\delta} \delta}}_2^8 + \lrn{u_{\step{t}{\delta} \delta}}_2^8}} + 128\delta  \lrp{\frac{d}{m}}^4\\
             &\qquad \leq 1.1 \E{ \lrp{\lrn{x_{\step{t}{\delta} \delta}}_2^8 + \lrn{u_{\step{t}{\delta} \delta}}_2^8}} + 2 \frac{d}{m},
            \end{align*}
            where the final two inequalities are both by our assumption that $\delta \leq \frac{1}{1000}$.
        \end{proof}
        
        \begin{lemma} \label{l:xu}
            For $(x_t, u_t)$ satisfying $\lrn{x_t}_2^2 + \lrn{x_t + u_t}_2^2 \geq 4 R^2$,     
            \begin{align*}
            \lin{x_t, u_t} + \lin{x_t + u_t, - u_t - \frac{\c}{L}\nabla U(x_{t})} 
            \leq&  - \frac{\c^2}{6} \lrp{\lrn{x_t}_2^2 + \lrn{x_t + u_t}_2^2} .
            \end{align*}
        \end{lemma}
        \begin{proof}
            We first verify that 
            \begin{align*}
            &\lin{x_t, u_t} + \lin{x_t + u_t, - u_t - \frac{\c}{L} \nabla U(x_{i\nu})}\\
            &\qquad = - \|u_t\|_2^2 - \frac{\c}{L}\lin{x_t,  \nabla U(x_{t})} - \lin{u_t,  \frac{\c}{L}\nabla U(x_{t})} \\
            &\qquad = - \|u_t\|_2^2 - \frac{\c}{L}\lin{x_t,  \nabla U(x_{t})} - \frac{\c}{L}\lin{u_t, \nabla U(x_{t})}\\
            &\qquad = -\lrn{u_t}_2^2 - \frac{\c}{L}\lin{x_t, \nabla U(x_t)} + \frac{1}{2} \lrn{u_t}_2^2 + \frac{\c^2}{2L^2} \lrn{\nabla U(x_t)}_2^2 - \frac{1}{2} \lrn{u_t + \frac{\c}{L} \nabla U(x_t)}_2^2\\
            &\qquad  \leq - \frac{1}{2} \lrn{u_t}_2^2 - \frac{\c}{L}\lin{x_t, \nabla U(x_t)} + \frac{\c^2}{2} \lrn{x_t}_2^2 =: \spadesuit
            \numberthis \label{e:t:oqi}
            \end{align*}
            
            Now consider two cases:
            
            \textbf{Case 1: ($\lrn{x_t}_2 \leq R$)}
            By Young's inequality we get that,
            \begin{align*}
            \lrn{x_t + u_t}_2^2 \leq 11 \lrn{u_t}_2^2 + 1.1 \lrn{x_t}_2^2.
            \end{align*}
            Furthermore, by our assumption that $\lrn{x_t}_2^2 + \lrn{x_t + u_t}_2^2 \geq 4R^2$, 
            \begin{align*}
            11\lrn{u_t}_2^2
            \geq & \lrn{x_t + u_t}_2^2 - 1.1 \lrn{x_t}_2^2\\
            \geq & 4 R^2 - 2.1 R^2\\
            \geq & 1.9 R^2\\
            \geq & 1.9 \lrn{x_t}_2^2.
            \numberthis \label{e:t:mnwen}
            \end{align*}
            Thus in this case $\lrn{u_t}_2^2 \geq \frac{1}{10} R^2$, and $\spadesuit$ can be upper bounded by
            \begin{align*}
            \spadesuit = & - \frac{1}{2} \lrn{u_t}_2^2 - \lin{x_t, \frac{\c}{L}\nabla U(x_t)} + \frac{c^2}{2} \lrn{x_t}_2^2\\
            \overset{(i)}{\leq}& - \frac{1}{2} \lrn{u_t}_2^2 + \c \lrn{x_t}_2^2 + \frac{\c^2}{2} \lrn{x_t}_2^2\\
            \overset{(ii)}{\leq}& - \frac{1}{2} \lrn{u_t}_2^2 + 2\c \lrn{x_t}_2^2\\
            \overset{(iii)}{\leq}& - \frac{1}{4} \lrn{u_t}_2^2\\
            \overset{(iv)}{\leq}& - \frac{1}{160} \lrp{\lrn{x_t}_2^2 + \lrn{x_t + u_t}_2^2},
            \end{align*}
            where $(i)$ is by $L$-Lipschitz of $\nabla U$ and Cauchy-Schwarz, $(ii)$ and $(iii)$ are because $\c := \frac{1}{1000 \kappa}  \leq \frac{1}{1000}$ and by Eq.~\eqref{e:t:mnwen}, the $(iv)$ is because
            \begin{align*}
         \|x_t\|_2^2 + \|x_t + u_t\|_2^2 
            \leq & 3\|x_t\|_2^2 + 2\|u_t\|_2^2 \\
            \leq & 30 \|u_t\|_2 + 2\|u_t\|_2 \\
            \leq & 40 \|u_t\|_2^2,
            \end{align*}
            where the second inequality is by again by Eq.~\eqref{e:t:mnwen}.
            
            \textbf{Case 2: ($\lrn{x_t}_2 \geq R$})
            
            By Assumption \ref{ass:strongconvexity}, $- \frac{\c}{L} \lin{x_t, \nabla_t} \leq - \frac{\c}{\kappa} \lrn{x_t}_2^2$. Thus we can upper bound $\spadesuit$ as follows:
            \begin{align*}
            \spadesuit
            = & - \frac{1}{2} \lrn{u_t}_2^2 - \frac{\c}{L}\lin{x_t, \nabla_t} + \frac{c^2}{2} \lrn{x_t}_2^2\\
            \le& - \frac{1}{2} \lrn{u_t}_2^2 - \c^2 \lrn{x_t}_2^2 + \frac{c^2}{2} \lrn{x_t}_2^2\\
           \le & - \lrn{u_t}_2^2 - \frac{\c^2}{2} \lrn{x_t}_2^2\\
            \leq & - \frac{\c^2}{3} \lrp{\lrn{x_t}_2^2 + \lrn{x_t + u_t}_2^2}.
            \end{align*}
            
            Putting the previous two results together, and using Young's inequality:
            \begin{align*}
            \spadesuit \leq& - \frac{\c^2}{3} \lrp{\lrn{x_t}_2^2 + \lrn{x_t + u_t}_2^2} \\
            \leq& - \frac{\c^2}{3} \lrp{\lrn{x_t}_2^2 + \lrn{x_t + u_t}_2^2}\\
            \leq& - \frac{\c^2}{6} \lrp{\lrn{x_t}_2^2 + \lrn{x_t + u_t}_2^2}.
            \end{align*}
        \end{proof}

        \subsection{Proof of Lemma~\ref{l:energy_bound_all_time_y}}
            Let us consider the Lyapunov function $l(y_t, v_t):= \lrp{\lrn{y_t}_2^2 + \lrn{y_t + v_t}_2^2 - 4R^2}_+^4$.
            
           By calculating its derivatives we can verify that 
            \begin{align*}
            & \nabla_x l(y_t, v_t) = 8 l(y_t,v_t)^{3/4} \lrp{y_t}\\
            & \nabla_u l(y_t, v_t) = 8 l(y_t,v_t)^{3/4} \lrp{y_t + v_t}\\
            & \nabla_u^2 l(y_t,v_t) = 8 l(y_t,v_t)^{3/4} I + 24 l(y_t,v_t)^{2/4} \lrp{y_t + v_t} \lrp{y_t + v_t}^T.
            \end{align*}
            Recall the dynamics of the variables $y_t$ and $v_t$,
            \begin{align*}
            d y_t =& v_t dt\\
            d v_t =& -2 v_t - \frac{\c}{L}\nabla U(x_{y_t}) dt + 2\sqrt{\frac{\c}{L}} dB_t.
            \end{align*}
            By $\Ito$'s lemma we can study the time evolution of this Lyapunov function:
            \begin{align*}
             d l(y_t,v_t) 
            &= 8l(y_t, v_t)^{3/4} \lrp{\lin{y_t, v_t} + \lin{y_t + v_t, - v_t - \frac{\c}{L}\nabla U(y_{t})}} dt\\
            & \qquad + \frac{16\c}{L}\lrp{l(y_t, v_t)^{3/4}d + l(y_t,v_t)^{2/4} \lrn{y_t + v_t}_2^2 } dt\\
            & \qquad\qquad + 16\sqrt{\frac{\c}{L}}l(y_t, v_t)^{3/4} \lrp{\lin{y_t, v_t} + \lin{y_t + v_t, d B_t}}\\
            &\overset{(i)}{\le} 8l(y_t, v_t)^{3/4} \lrp{- \frac{\c^2}{6} \lrp{\lrn{y_t}_2^2 + \lrn{y_t + v_t}_2^2}} dt\\
            & \qquad + \frac{16\c}{L}\lrp{l(y_t, v_t)^{3/4}d + l(y_t,v_t)^{2/4} \lrn{y_t + v_t}_2^2 } dt\\
            & \qquad\qquad + 16\sqrt{\frac{\c}{L}}l(y_t, v_t)^{3/4} \lrp{\lin{y_t, v_t} + \lin{y_t + v_t, d B_t}}\\
            &\overset{(ii)}{\le} -\c^2 l(y_t, v_t) dt\\
            & \qquad + \frac{32\c}{L}\lrp{l(y_t, v_t)^{3/4}d} dt + \frac{64\c}{L}\lrp{l(y_t,v_t)^{2/4} R^2 } dt\\
            & \qquad\qquad + 16\sqrt{\frac{\c}{L}}l(y_t, v_t)^{3/4} \lrp{\lin{y_t, v_t} + \lin{y_t + v_t, d B_t}}\\
            &\le -\c^2 l(y_t, v_t) dt\\
            &\qquad + \frac{\c^2}{8}l(y_t,v_t) dt + \frac{2^{25} d^4}{\c^2 L^4} dt + \frac{\c^2}{8} l(y_t,v_t) dt + \frac{2^{16} R^4}{L^2}\\
            & \qquad\qquad + 16\sqrt{\frac{\c}{L}}l(y_t, v_t)^{3/4} \lrp{\lin{y_t, v_t} + \lin{y_t + v_t, d B_t}}\\
            &\leq -\frac{\c^2}{2} l(y_t, v_t) dt + 2^{26} \lrp{\frac{d^4}{\c^2 L^4} + c^2 R^8}dt\\
            & \qquad+ 16\sqrt{\frac{\c}{L}}l(y_t, v_t)^{3/4} \lrp{\lin{y_t, v_t} + \lin{y_t + v_t, d B_t}},
            \end{align*}
            where $(i)$ can be proved by an argument similar to the proof of Lemma~\ref{l:xu}, and is omitted, while $(ii)$ follows because
            \begin{align*}
             \lrn{y + v}_2^2 \leq l(y,v)^{1/4} + 4R^2, \quad \text{and, } \quad 
             \lrn{y}_2^2 + \lrn{x + u}_2^2 \geq l(y,v)^{1/4}
            \end{align*}
            by the definition of $l(x,u)$.
            Taking expectations on both sides, the term involving the Brownian motion, $dB_t$, goes to zero. Note also that $(y_t,v_t)$ is distributed according to the invariant distribution $p^*$ for all $t\ge 0$, therefore,
            \begin{align*}
            0 = \ddt \E{l(y_t, v_t)}
            \leq& -\frac{\c^2}{2} \E{l(y_t, v_t)} + 2^{26} \lrp{\frac{d^4}{\c^2 L^4} + c^2 R^8}\\
            \leq& -\frac{\c^2}{2} \E{l(y_t, v_t)} + 2^{26} \c^2 \lrp{10^{12} \frac{d^4}{m^4} + R^8}
            \end{align*}
            Thus
            \begin{align*}
            \E{l(y_t,v_t)} 
            \leq 2^{27} \lrp{10^{12}\frac{d^4}{m^4} + R^8}
            \leq 2^{66} \lrp{\frac{d}{m} + R^2}^4.
            \end{align*}
    This completes the proof of the lemma.
            
    We now state and prove some auxillary lemmas that were useful in the proof above.

        \begin{lemma}\label{l:energy_bound_pt}
            Let $x_t$ be evolved according to the dynamics in Eq.~\eqref{d:o:xt}.
            Then for all $t\ge 0$, 
            \begin{align*}
            \E{\lrn{x_t}_2^2} \leq 2 \lrp{R^2 + \frac{d}{m}}.
            \end{align*}
        \end{lemma}
        \begin{proof}
            Let $\theta_k \sim \N(0, I)$ then we have,
            \begin{align*}
             \ind{\lrn{x_{k\delta}}_2 \leq R} \cdot \lrn{x_{(k+1)\delta}}_2^2
            &= \ind{\lrn{x_{k\delta}}_2 \leq R} \cdot \lrn{x_{k\delta} - \delta \nabla U(x_k) + \sqrt{2\delta} \theta_k}_2^2\\
            &\leq \ind{\lrn{x_{k\delta}}_2 \leq R} \cdot \lrn{x_{k\delta} - \delta \nabla U(x_k)}_2^2 \\
            &\quad + \ind{\lrn{x_{k\delta}}_2 \leq R} \cdot \lrn{\sqrt{2\delta} \theta_k}_2^2\\
            &\quad + \ind{\lrn{x_{k\delta}}_2 \leq R} \cdot 2\lin{x_{k\delta} - \delta \nabla U(x_k), \sqrt{2\delta} \theta_k }.
            \end{align*}
            Consider two cases:
            
            If $\lrn{x_{k\delta}}_2 \geq R$, then 
            \begin{align*}
            \lrn{x_{k\delta} - \delta \nabla U(x_k)}_2^2
            &= \lrn{x_{k\delta}}_2^2 - \lin{x_{k\delta}, 2\delta \nabla U(x_k)} + \delta^2 \lrn{\nabla U(x_k)}_2^2\\
            &\overset{(i)}{\leq} \lrp{1-2\delta m} \lrn{x_{k\delta}}_2^2 + \delta^2 \lrn{\nabla U(x_k)}_2^2\\
            &\overset{(ii)}{\leq} \lrp{1-2\delta m + \delta^2 L^2} \lrn{x_{k\delta}}_2^2\\
            &\overset{(iii)}{\leq} \lrp{1-\delta m} \lrn{x_{k\delta}}_2^2,
            \end{align*}
            where $(i)$ is by Assumption~\ref{ass:strongconvexity}, $(ii)$ is by Assumption~\ref{ass:smoothness}, and $(iii)$ is by our assumption that $\delta \leq \frac{1}{\kappa L}$.
            
            While I=if $\lrn{x_{k\delta}}_2 \leq R$, then
            \begin{align*}
             \lrn{x_{k\delta} - \delta \nabla U(x_k)}_2^2
            &= \lrn{x_{k\delta}}_2^2 - \lin{x_{k\delta}, 2\delta \nabla U(x_k)} + \delta^2 \lrn{\nabla U(x_k)}_2^2\\
            &\overset{(i)}{\leq} \lrp{1 + 2\delta L + \delta^2 L^2} \lrn{x_k}_2^2\\
            &\overset{(ii)}{\leq} \lrp{1 + 3\delta L} \lrn{x_k}_2^2,
            \end{align*}
            where $(i)$ is by Assumption~\ref{ass:smoothness}, and $(ii)$ is by our assumption that $\delta \leq \frac{1}{\kappa L}$.
            
            Thus for both cases we have,
            \begin{align*}
             \lrn{x_{k\delta} - \delta \nabla U(x_k)}_2^2
            \leq& \ind{\lrn{x_{k\delta}}_2 \geq R} \lrp{1-\delta m} \lrn{x_{k\delta}}_2^2+ \ind{\lrn{x_{k\delta}}_2 \leq R}\lrp{1 + 3\delta L} \lrn{x_k}_2^2\\
            \leq& \lrn{x_{k\delta}}_2^2 - \delta m \lrn{x_{k\delta}}_2^2 + \ind{\lrn{x_{k\delta}}_2 \leq R} \cdot (3\delta L + \delta m) \lrn{x_{k\delta}}_2^2.
            \end{align*}
            Thus we have:
            \begin{align*}
             \ind{\lrn{x_{k\delta}}_2 \leq R} \lrn{x_{k\delta} - \delta \nabla U(x_k)}_2^2
            \leq& \ind{\lrn{x_{k\delta}}_2 \leq R} (1-\delta m) \lrn{x_{k\delta}}_2^2.
            \end{align*}
            By taking expectations with respect to the Brownian motion we get,
            \begin{align*}
             \E{\ind{\lrn{x_{k\delta}}_2 \leq R} \lrn{x_{(k+1)\delta}}_2^2} 
            &\leq (1-\delta m) \E{ \ind{\lrn{x_{k\delta}}_2 \leq R}\lrn{x_{k\delta}}_2^2}\\
            & \qquad \qquad + \E{ \ind{\lrn{x_{k\delta}}_2 \leq R} \lrn{\sqrt{2\delta} \theta_k}_2^2}\\
            &\leq (1-\delta m) \E{ \ind{\lrn{x_{k\delta}}_2 \leq R}\lrn{x_{k\delta}}_2^2} + 2\delta d.
            \end{align*}
            Applying this inequality recursively over $k$ steps we arrive at,
            \begin{align*}
            \E{\ind{\lrn{x_{k\delta}}_2 \leq R} \lrn{x_{(k+1)\delta}}_2^2}
            \leq& e^{-\delta m} \E{\ind{\lrn{x_{k\delta}}_2 \leq R} \lrn{x_{(k+1)\delta}}_2^2} + \frac{2\delta d}{\delta m}\\
            \leq& \frac{2d}{m}.
            \end{align*}
            Thus we get that,
            \begin{align*}
            \E{\lrn{x_{(k+1)\delta}}_2^2} \leq 2 \lrp{R^2 + \frac{d}{m}}.
            \end{align*}
        \end{proof}
        
        \begin{lemma}\label{l:energy_bound_pstar}
            Let $y\sim p^*(y) \propto e^{-U(y)}$.
            Then 
            \begin{align*}
            \E{\lrn{y}_2^8} \leq 2^{20} \lrp{\frac{d^4}{m^4} + R^8}.
            \end{align*}
        \end{lemma}
        \begin{proof}
            Let $l(y) := \lrp{\lrn{y}_2^2 - R^2}_+^4$.
            We  calculate derivatives and verify that
            \begin{align*}
            & \nabla l(y) = 8 l(y)^{3/4} \cdot y\\
            & \nabla^2 l(y) = 48 l(y)^{2/4} \cdot yy^T + 8 l(y)^{3/4} I,
            \end{align*}
            where $I$ is the identity matrix.
            By $\Ito$'s Lemma:
            \begin{align}\label{e:itoappliedtol}
            d l(y_t) 
            = \lin{\nabla l(y_t), -\nabla U(y_t)} dt + \lin{\nabla l(y_t), \sqrt{2} dB_t} + \frac{1}{2} \tr\lrp{\nabla^2 l(y_t)}.
            \end{align}
            We start by analyzing the first term,
            \begin{align*}
             \lin{\nabla l(y_t), - \nabla U(y_t)}
            &= \lin{8 l(y_t)^{3/4} \cdot y_t, - \nabla U(y_t)}\\
            &\overset{(i)}{\leq} \ind{\lrn{y_t}_2 \geq R} \lrp{8l(y_t)^{3/4}} \lrp{-m \lrn{y_t}_2^2}\\ &\qquad \qquad  + \ind{\lrn{y_t}_2 < R} \lin{8l(y_t)^{3/4} y, -\nabla U(y_t)}\\
            &\overset{(ii)}{=} \lrp{8l(y_t)^{3/4}} \lrp{-m \lrn{y_t}_2^2}\\
            &\leq -8m l(y_t),
            \end{align*}
            where $(i)$ is by Assumption~\ref{ass:strongconvexity},  and, $(ii)$ is because $\ind{\lrn{y_t}_2 < R} \cdot l(y_t) = 0$ and $\ind{\lrn{y_t}_2 \geq R} \cdot l(y_t) = l(y_t)$ by definition of $l(y_t)$.
            
            Consider the other term on the right-hand side of Eq.~\eqref{e:itoappliedtol}:
            \begin{align*}
             \tr\lrp{\nabla^2 l(y_t)}
            &= 48 l(y_t)^{2/4} \lrn{y}_2 + 8 l(y_t)^{3/4} d\\
            &\overset{(i)}{\leq} 48 l(y_t)^{3/4} + 8 l(y_t)^{3/4} d + 48 l(y_t)^{2/4}R^2\\
            &\leq 64 l(y_t)^{3/4} d + 48 l(y_t)^{2/4}R^2\\
            &\overset{(ii)}{\leq} 2m l(y_t) + 2^{21} \frac{d^4}{m^3} + 2m l(y_t) + 2^{11} \frac{R^4}{m}\\
            &\overset{(iii)}{\leq} 4m l(y_t) + 2^{22} \lrp{\frac{d^4}{m^3} + mR^8},
            \end{align*}
            where $(i)$ is by definition of $l(y)$, while $(ii)$ and $(iii)$ are by Young's inequality.
            
            Put together into Eq.~\eqref{e:itoappliedtol} and taking expectations,
            \begin{align*}
            \ddt \E{l(y_t)} 
            \leq& - 8m\E{l(y_t)}+ 4m l(y_t) + 2^{22} \lrp{\frac{d^4}{m^3} + mR^8}\\
            \leq& -4m \E{l(y_t)} + 2^{22} \lrp{\frac{d^4}{m^3} + mR^8}.
            \end{align*}
            Since $y_t \sim p^*$ for all $t$, $\ddt \E{l(y_t)} = 0$, thus we get,
            \begin{align*}
            \E{l(y_t)} \leq 2^{20} \lrp{\frac{d^4}{m^4} + R^8}.
            \end{align*}
        \end{proof}
    \end{section}
    
    \begin{section}{Existence of Coupling}
    \label{ass:existence_coupling}

        \begin{proof}[Proof of Lemma \ref{l:existence_of_process}]
            We prove the existence of a unique strong solution for $(x_t, u_t, y_t, v_t, \tau_{\step{t}{\nu}})$ inductively: Let $k$ be an arbitrary nonnegative integer, and suppose that the lemma statement holds for all $s \in [0, k\nu]$.  We show that the lemma statement holds for all $s\in [0, (k+1)\nu]$.

            First, we can verify that for $t\in [k\nu, (k+1)\nu)$,
            \begin{align*}
            \tau_{\step{t}{\nu}} = \tau_{k},
            \end{align*}
            that is, $\tau_{\step{t}{\nu}}$ is a constant, and so $\mu_{\step{t}{\nu}} = \mu_k$ is also a constant.
            
            Next, we find that for $t\in [k\nu, (k+1)\nu)$, the following is algebraically equivalent to dynamics described by Eqs.\eqref{d:xt}--\eqref{d:vt}:
            \begin{align*}
            & d x_t = u_t dt,\\
            & d u_t = -2 u_t dt - \frac{\c}{L}\nabla U\lrp{x_{\step{t}{\delta} \delta}} dt + 2\sqrt{\frac{\c}{L}} dB_t,\\
            & d y_t = v_t dt,\\
            & d v_t = -2 v_t - \frac{\c}{L}\nabla U(y_t) dt + 2\sqrt{\frac{\c}{L}} dB_t - \mu_{k} \cdot \lrp{4\sqrt{\frac{\c}{L}}\gamma_t \gamma_t^T dB_t + 2 \sqrt{\frac{\c}{L}} \bar{\gamma}_t \bar{\gamma}_t^T dA_t},
            \end{align*}
            where we use the fact that $\mu_t$ takes on a constant value over $t\in[k\nu, (k+1)\nu)$. 
            
            We proceed by applying Theorem~5.2.1 of \cite{oksendal2013stochastic}, which states that if the following holds:
            \begin{enumerate}
                \item $\E{\lrn{x_{k\nu}}_2^2+\lrn{y_{k\nu}}_2^2+\lrn{u_{k\nu}}_2^2+\lrn{v_{k\nu}}_2^2} \leq \infty$.
                \item For all $x,y \in \mathbb{R}^d$, $\lrn{\nabla U(x) - \nabla U(y)}_2 \leq D\lrn{x-y}_2$ for some constant $D > 0$.
                \item For all $(x,y,u,v), (x',y',u',v')$,
                \begin{align*}
                \lrn{{\gamma} \gamma^T - {\gamma}'\gamma'^T}_2 + \lrn{\bar{\gamma}\bar{\gamma}_t'^T - \bar{\gamma}_t'\bar{\gamma}'^T}_2 \leq D \lrp{\lrn{x-x'}_2+\lrn{y-y'}_2 + \lrn{u-u'}_2 + \lrn{v-v'}_2},
                \end{align*}
                for some constant $D$ (where $\gamma$ and $\bar{\gamma}$ are functions of $(x,y,u,v)$, as defined in Eq.~\eqref{d:gamma}, similarly for $\gamma'$, $\bar{\gamma}'$ and $(x',y',u',v')$),
            \end{enumerate}
        then there is a solution $(x_t,y_t,u_t,v_t)$ for $t\in[k\nu, (k+1)\nu]$ with the properties:
            \begin{enumerate}[label=(\alph*)]
                \item $(x_t,y_t,u_t,v_t)$ is unique and $t$-continuous with probability one.
                \item $(x_t,y_t,u_t,v_t)$ is adapted to the filtration $\F_t$ generated by $(x_{k\nu},y_{k\nu},u_{k\nu},v_{k\nu})$ and $dB_t$ and $dA_t$ for $t\in[k\nu, (k+1)\nu)$.
                \item $\int_0^T \E{ \lrn{x_t}_2^2 + \lrn{y_t}_2^2 + \lrn{u_t}_2^2 + \lrn{v_t}_2^2} dt < \infty$.
            \end{enumerate}
            
            We can verify the first condition holds by using Lemma~\ref{l:energy_bound_all_time} and Lemma~\ref{l:energy_bound_all_time_y}. Condition 2 holds due to our smoothness assumption, Assumption~\ref{ass:smoothness}.
            
            We can verify that Condition 3 also holds using the argument below:
            
            From the definition of $\mathcal{M}$ in Eq.~\eqref{d:gamma}, we know that $\lrabs{\mathcal{M}(r)'} \leq \frac{1}{2} \lrabs{\sin\lrp{r \cdot 2\pi/\beta}}\cdot \frac{2\pi}{\beta} \leq \frac{\pi}{\beta}$.

           By definition of $\gamma_t$ in Eq.~\eqref{d:gamma},
            \begin{align*}
            &\gamma_t \gamma_t^T := \mathcal{M}(\lrn{z_t + w_t}_2) \cdot \frac{\lrp{z_t + w_t} \lrp{z_t + w_t}^T}{\lrn{z_t + w_t}_2^2}.
            \end{align*}
            
            To simplify notation, consider an arbitrary $x,y\in \Re^d$, and assume wlog that $\lrn{x}_2 \leq \lrn{y}_2$. We will bound 
            \begin{align*}
            \lrn{\mathcal{M}(\lrn{x}_2) \frac{xx^T}{\lrn{x}_2^2} - \mathcal{M}(\lrn{y}_2) \frac{yy^T}{\lrn{y}_2^2}}_2 \leq D\lrn{x-y}_2,
            \end{align*}
            for some $D$, which implies condition 3.
            
            By the triangle inequality, 
            \begin{align*}
             \lefteqn{\lrn{\mathcal{M}(\lrn{x}_2) \frac{xx^T}{\lrn{x}_2^2} - \mathcal{M}(\lrn{y}_2) \frac{yy^T}{\lrn{y}_2^2}}_2} \\
            &\leq \mathcal{M}(\lrn{x}_2) \lrn{\frac{xx^T}{\lrn{x}_2^2} - \frac{yy^T}{\lrn{y}_2^2}}_2 + \lrn{\frac{yy^T}{\lrn{y}_2^2}}_2 \lrabs{\mathcal{M}(\lrn{x}_2) - \mathcal{M}(\lrn{y}_2) }\\
            &\leq \mathcal{M}(\lrn{x}_2) \lrn{\frac{xx^T}{\lrn{x}_2^2} - \frac{yy^T}{\lrn{y}_2^2}}_2 + \lrabs{\mathcal{M}(\lrn{x}_2) - \mathcal{M}(\lrn{y}_2) }.
            \numberthis \label{e:t:sdfnm}
            \end{align*}
            The second term can be bounded as
            \begin{align*}
            & \lrabs{\mathcal{M}(\lrn{x}_2) - \mathcal{M}(\lrn{y}_2) } 
            \leq \frac{\pi}{\beta} \lrabs{\lrn{x}_2 - \lrn{y}_2}_2 \leq \frac{\pi}{\beta} \lrn{x-y}_2,
            \end{align*}
            where we use the upper bound we established on $\lrabs{\mathcal{M}'(r)}$.
            
            To bound the first term, we consider two cases:
            
            If $\lrn{x}_2 \leq \beta/2$, $\mathcal{M}(\lrn{x}_2) = 0$ and we are done.

            If $\lrn{x}_2 \geq \beta/2$, we verify that the transformation $T(x) = \frac{x}{\lrn{x}_2}$ has Jacobian $\nabla T(x) = \frac{1}{\lrn{x}_2} \lrp{I - \frac{xx^T}{\lrn{x}_2}}$, so that $\lrn{\nabla T(x)}_2 \leq \frac{1}{\lrn{x}_2}$. By our earlier assumption that $\lrn{x}_2 \leq \lrn{y}_2$, we know that $\lrn{ax + (1-a)y}_2 \geq \beta/2$ for all $a\in[0,1]$. Therefore,
            \begin{align*}
            \lrn{\frac{x}{\lrn{x}_2} - \frac{y}{\lrn{y}_2}}_2 = \lrn{T(x) - T(y)}_2 \leq \frac{1}{\lrn{x}_2} \lrn{x-y}_2 \leq \frac{2}{\beta}.
            \end{align*}
            By the triangle inequality and some algebra, we obtain:
            \begin{align*}
            \lefteqn{ \lrn{\frac{xx^T}{\lrn{x}_2^2}-\frac{yy^T}{\lrn{y}_2^2}}_2}\\
            &\leq \lrn{\frac{x}{\lrn{x}_2} + \frac{y}{\lrn{y}_2}}_2\lrn{\frac{x}{\lrn{x}_2}-\frac{y}{\lrn{y}_2}}_2\\
            &\leq 2\lrn{\frac{x}{\lrn{x}_2}-\frac{y}{\lrn{y}_2}}_2\\
            &\leq \frac{4}{\beta}\lrn{x-y}_2,
            \end{align*}
            where the first two inequalities are due to the triangle inequality. Combined with the fact that $\mathcal{M}(r) \leq 1$ for all $r$, we can bound Eq.~\eqref{e:t:sdfnm} by $\frac{8}{\beta} \lrn{x-y}_2$.
            
            A similar argument can be used to show that $\bar{\gamma}_t$ is Lipschitz. Let $\mathcal{N}(x) := \lrp{1-\lrp{1-2\mathcal{M}\lrp{\lrn{x}_2}}^2}^{1/2}$. Then we verify that
            \begin{align*}
            \mathcal{N}(r) := &
            \threecase{1}{r \in [\beta, \infty)}{\sin \lrp{r \cdot \frac{2\pi}{\beta}}}{r \in[\beta/2, \beta]}{0}{r \in[0,\beta/2]}\\
            \bar{\gamma}_t :=& \lrp{\mathcal{N}\lrp{\lrn{z_t + w_t}_2}}^{1/2}\frac{z_t + w_t}{\lrn{z_t + w_t}_2}.
            \end{align*}
            
            The proof is almost identical to the proof of \eqref{e:t:sdfnm}, so we omit it, but highlight two crucial facts:
            \begin{align*}
            &1.\ \mathcal{N}(r) \in [0,1] \qquad \text{for all $r$}\\
            &2.\ \lrabs{\mathcal{N}'(r)} \leq \frac{2\pi}{\beta} \qquad \text{for all $r$}.
            \end{align*}                   
            Thus we find that Condition 3 is satisfied with $D = \frac{16}{\beta}$, and in turn show that (a)-(c) hold for $t\in[k\nu, (k+1)\nu]$. From Eq.~\eqref{d:taut} we know that $\tau_{{(k+1)\nu}}$ is a function of $(x_{(k+1)\nu}, u_{(k+1)\nu}, y_{(k+1)\nu}, v_{(k+1)\nu}, \tau_{k})$. Thus we have shown the existence of a unique solution $(x_t,y_t,u_t,v_t,\tau_{\step{t}{\nu}})$ for $t\in[k\nu, (k+1) \nu]$, where $(x_t,y_t,u_t,v_t)$ is $t$-continuous.
            
            The proof of the lemma now follows by induction over $k$.
        \end{proof}
        \begin{lemma}\label{l:existence_of_phi}
            Let $B_t$ and $A_t$ be two independent Brownian motions, and let $\F_t$ be the $\sigma$-algebra generated by $B_s$, $A_s$; $s\leq t$, and $(x_0,u_0,y_0,v_0)$.
            
            For all $t\ge 0$, the stochastic process $\phi_t$ defined in Eqs.~\eqref{d:xi} has a unique solution such that $\phi_t$ is $t$-continuous with probability one, and satisfies the following, for all $s\ge 0$:
            \begin{enumerate}
                \item $\phi_t$ is adapted to the filtration $\F_s$.
                \item $\E{\lrn{\phi_t}_2^2} \leq \infty$.
            \end{enumerate}
        \end{lemma}
        \begin{proof}
            The proof is almost identical to that of Lemma \ref{l:existence_of_process}. The main additional requirement is showing that there exists a constant $D$ such that for any $(x,y,u,v)$ and $(x',y',u',v')$,
            \begin{align*}
            \lrn{\nabla_{w} f(r) \gamma \gamma^T - \nabla_{w'} f(r') \gamma' \gamma'^T}_2,
            \numberthis
            \label{e:l:existence_of_phi:objective}
            \end{align*}
            with $\gamma$ (resp $\gamma'$) being a function of $(x,y,u,v)$ (resp $\gamma'$) as defined in \eqref{d:gamma}. and $r$ being a function of $(x,y,u,v)$ as defined in \eqref{d:mut}. In the proof of Lemma \ref{l:existence_of_process}, we already showed that $\gamma \gamma^T$ and $\gamma' \gamma'^T$ are uniformly bounded and lipschitz, thus it is sufficient to show that
            \begin{align*}
            &1.\ \lrn{\nabla_{w} f(r) - \nabla_{w'} f(r')}_2 \leq D \lrn{x-x'}_2 + \lrn{y-y'}_2 + \lrn{u-u'}_2 + \lrn{v-v'}_2\\
            &2.\ \lrn{\nabla_{w} f(r)}_2 \leq D.
            \numberthis
            \label{e:l:existence_of_phi:objective_2}
            \end{align*}
            The second point is easy to verify:
            \begin{align*}
            \nabla_w f(r) = f'(r) \nabla_w r = f'(r) \lrp{\nabla_w\l(z + w)}
            \end{align*}
            Thus $\lrn{\nabla_w f(r)}_2 \leq 1$ using item (F2) of Lemma \ref{lem:fpropertiesall} and item 2 of Lemma \ref{l:def_l}.
            
            To prove the first point, we verify that
            \begin{align*}
            \nabla^2_w f(r) = f''(r) \lrp{\nabla_w\l(z + w)} + f'(r)\lrp{\nabla^2_w\l(z + w)}.
            \end{align*}
            Using item (F7) of Lemma \ref{lem:fpropertiesall} and item 
            \begin{align*}
            \lrn{\nabla^2_w f(r)}_2 \leq 4\Cf\Rf + \frac{4}{\Rf} + \frac{8}{\beta};
            \end{align*}
            this implies \ref{e:l:existence_of_phi:objective_2} which in turn implies \eqref{e:l:existence_of_phi:objective}. Note that $\lrn{w-w'}_2 \leq \lrn{u-u'}_2 + \lrn{v-v'}$.

        \end{proof}
    \end{section}
    
    \begin{section}{Coupling and Discretization}\label{ass:marginal_correctness}
        
        \begin{proof}[Proof of Lemma \ref{l:marginal_yt}]
            Let us define
            \begin{align*}
            \bar{B}_t := \int_0^t  dB_t - \ind{k\nu \geq \tau_{\step{t}{\nu}} + \Ts} \cdot \lrp{2\gamma_t \gamma_t^T dB_t + \bar{\gamma}_t \bar{\gamma}_t^T dA_t}.
            \end{align*}
            We will show that $\bar{B}_t$ is a Brownian motion by using Levy's characterization. The conclusion then follows immediately from the dynamics defined in Eq.~\eqref{e:exactunderdampedlangevindiffusion}.
            
            Since $B_t$ and $A_t$ are Brownian motions, $\bar{B}_t$ is also a continuous martingale with respect to the filtration $\F_t$. Further the quadratic variation of $\bar{B}_t$ over an interval $[s,s']$ is
            \begin{align*}
            & \int_s^{s'} \underbrace{\lrp{I - 2 \ind{k\nu \geq \tau_{\step{t}{\nu} }+\Ts }\gamma_t \gamma_t^T}^2 + \ind{k\nu \geq \tau_{\step{t}{\nu} }+\Ts } \lrp{\bar{\gamma}_t \bar{\gamma}_t^T}^2}_{=:\spadesuit} dt.
            \end{align*}
            If $\ind{k\nu \geq \tau_{\step{t}{\nu} }+\Ts } = 0$, then the above is clearly the identity matrix -- $I$. 
            
            If, on the other hand, $\ind{k\nu \geq \tau_{\step{t}{\nu} }+\Ts } = 1$, define $c_t := z_t + w_t$; then by the definition of $\gamma_t$ and $\bar{\gamma}_t$ in Eq.~\eqref{d:gamma}, we find that
            \begin{align*}
            \spadesuit
            &= \lrp{I - 2 \mathcal{M}(\lrn{c_t}_2) \frac{c_t c_t^T}{\lrn{c_t}_2^2}}^2 + \lrp{1 - \lrp{1 - 2\mathcal{M}(\lrn{c_t}_2)}^2} \frac{c_t c_t^T}{\lrn{c_t}_2^2}\\
            &\overset{(i)}{=} I - \frac{c_t c_t^T}{\lrn{c_t}_2^2} + \lrp{1-2\mathcal{M}(\lrn{c_t}_2)}^2  \frac{c_t c_t^T}{\lrn{c_t}_2^2} + \lrp{1 - \lrp{1 - 2\mathcal{M}(\lrn{c_t}_2)}^2} \frac{c_t c_t^T}{\lrn{c_t}_2^2}\\
            &= I,
            \end{align*}
            where $(i)$ follows by the eigenvalue decomposition of the matrix $\lrp{I - 2 \mathcal{M}(\lrn{c_t}_2) \frac{c_t c_t^T}{\lrn{c_t}_2^2}}^2$.
            
            Thus the quadratic variation of $\bar{B}_t$ over the interval $[s,s]$ is $(s'-s)I$, thus satisfying Levy's characterization of a Brownian motion.
            
        \end{proof}
        \begin{proof}[Proof of Lemma \ref{l:o:marginal_yt}]
                Using similar steps as Lemma~\ref{l:marginal_yt}, we can verify that 
                \begin{align*}
                    \bar{B}_t := \int_0^t \sqrt{2} dB_t - 2\sqrt{2} \gamma_t \gamma_t^T dB_t + \sqrt{2} \bar{\gamma}_t \bar{\gamma}_t^T dA_t.
                \end{align*}
                is a Brownian motion. The proof follows immediately.
            \end{proof}
        \begin{lemma}\label{l:explicitform}
            Given $(x_{k\delta}, u_{k\delta})$, the solution $(x_t,u_t)$, for $t\in (k\delta, (k+1)\delta]$, of the discrete underdamped Langevin diffusion defined by the dynamics in Eq.~\eqref{e:discreteunderdampedlangevindiffusion} is
            \begin{align*}
            \numberthis \label{e:vtildedynamics}
            u_t &= u_{k\delta} e^{-2 (t-k\delta)} - \frac{\c}{L} \left(\int_{k\delta}^t e^{-2(t-s)} \nabla U(x_{k\delta}) ds \right) + \sqrt{\frac{4\c}{L}} \int_{k\delta}^t e^{-2 (t-s)} dB_s\\
            x_t &= x_{k\delta} + \int_{k\delta}^t u_s ds.
            \end{align*}
        \end{lemma}
        \begin{proof}
            It can be easily verified that the above expressions have the correct initial values $(x_{k\delta},u_{k\delta})$. By taking derivatives, one can also verify that they satisfy the stochastic differential equations in \eqrefmike{e:discreteunderdampedlangevindiffusion}.
        \end{proof}
        
        \begin{lemma}\label{l:gaussianexpressionforsamplingxtut}
            Conditioned on $(x_{k\delta}, u_{k\delta})$, the solution $(x_{(k+1)\delta},u_{(k+1)\delta})$ of Eq.~\eqref{e:discreteunderdampedlangevindiffusion} is a Gaussian with  mean,
            \begin{align*}
            \E{u_{(k+1)\delta} } & = u_{k\delta} e^{-2 \delta} - \frac{\c}{2L}(1-e^{-2 \delta}) \nabla f(x_{k\delta})\\
            \E{x_{(k+1)\delta} } & = x_{k\delta} + \frac{1}{2}(1-e^{-2 \delta})u_{k\delta} - \frac{\c}{2L} \left( \delta - \frac{1}{2}\left(1-e^{-2 \delta}\right) \right) \nabla U(x_{k\delta}),
            \end{align*}
            and covariance,
            \begin{align*}
            \E{\left(x_{(k+1)\delta} - \E{x_{(k+1)\delta}}\right) \left(x_{(k+1)\delta} - \E{x_{(k+1)\delta}}\right)^{\top}}&= \frac{\c}{L } \left[\delta-\frac{1}{4}e^{-4\delta}-\frac{3}{4}+e^{-2\delta}\right] \cdot I_{d\times d}\\
            \E{\left(u_{(k+1)\delta} - \E{u_{(k+1)\delta}}\right) \left(u_{(k+1)\delta} - \E{u_{(k+1)\delta}}\right)^{\top} } &= \frac{\c}{L}(1-e^{-4 \delta})\cdot I_{d\times d}\\
            \E{\left(x_{(k+1)\delta} - \E{x_{(k+1)\delta}}\right) \left(u_{(k+1)\delta} - \E{u_{(k+1)\delta}}\right)^{\top} }&= \frac{\c}{2L} \left[1+e^{-4\delta}-2e^{-2\delta}\right] \cdot I_{d \times d}.
            \end{align*}
        \end{lemma}
        \begin{proof}
            Consider some $t\in[k\delta, (k+1)\delta)$.
            
            It follows from the definition of Brownian motion that the distribution of $(x_t,u_t)$ is a $2d$-dimensional Gaussian distribution. We will compute its moments below, using the expression in Lemma~\ref{l:explicitform}. Computation of the conditional means is straightforward, as we can simply ignore the zero-mean Brownian motion terms:
            \begin{align}
            \E{u_t } & = u_{k\delta} e^{-2 (t-k\delta)} - \frac{\c}{2 L}(1-e^{-2 (t-k\delta)}) \nabla U(x_{k\delta})\\
            \E{x_t } & = x_{k\delta} + \frac{1}{2}(1-e^{-2 (t-k\delta)})u_{k\delta} - \frac{\c}{2 L} \left( t - k\delta - \frac{1}{2}\left(1-e^{-2 (t-k\delta)}\right) \right) \nabla U(x_{k\delta}).
            \end{align}
            The conditional variance for $u_t$ only involves the Brownian motion term:
            \begin{align*}
            \E{\left(u_t - \E{u_t}\right) \left(u_t - \E{u_t}\right)^{\top} }
            &= \frac{4\c}{L}\E{\left(\int_{k\delta}^t e^{-2 (t-s)} dB_s\right)\left(\int_{k\delta}^t e^{-2(s-t)} dB_s\right)^{\top}}\\
            &= \frac{4\c}{L} \left(\int_{k\delta}^t e^{-4(t-s)} ds \right) \cdot I_{d\times d}\\
            &= \frac{\c}{L}(1-e^{-4 (t-k\delta)})\cdot I_{d\times d}.
            \end{align*}
            The Brownian motion term for $x_t$ is given by
            \begin{align*}
            \sqrt{\frac{4\c}{L}} \int_{k\delta}^t  \left( \int_{k\delta}^r e^{-2 (r-s)} dB_s \right)dr
            &= \sqrt{\frac{4\c}{L}} \int_{k\delta}^t e^{2s}\left( \int_s^t e^{-2r} dr \right) dB_s\\
            &= \sqrt{\frac{\c}{L}} \int_{k\delta}^t \left(1-e^{-2(t-s)} \right) dB_s.
            \end{align*}
            Here the second equality follows by Fubini's theorem. The conditional covariance  for $x_t$ now follows as 
            \begin{align*}
            \E{\left(x_t - \E{x_t}\right) \left(x_t - \E{x_t}\right)^{\top} }
            &= \frac{\c}{L} \E{\left( \int_{k\delta}^t \left(1- e^{-2(t-s)}  \right)dB_s\right)\left( \int_{k\delta}^t \left(1-e^{-2(t-s)} \right)dB_s\right)^{\top}}\\
            &= \frac{\c}{L} \left[ \int_{k\delta}^t \left(1-e^{-2(t-s)}\right )^2 ds \right]\cdot I_{d \times d}\\
            &= \frac{\c}{L} \left[t-k\delta-\frac{1}{4}e^{-4(t-k\delta)}-\frac{3}{4}+e^{-2(t-k\delta)}\right] \cdot I_{d\times d}.
            \end{align*}
            Finally we compute the cross-covariance between $x_t$ and $u_t$,
            \begin{align*}
            \E{\left(x_t - \E{x_t}\right) \left(u_t - \E{u_t}\right)^{\top}}
            &= \frac{2 \c}{L} \E{\left( \int_{k\delta}^t\left(1-e^{-2 (t-s)} \right)dB_s\right)\left( \int_{k\delta}^t e^{-2 (t-s)} dB_s\right)^{\top}}\\
            &= \frac{2 \c}{L} \left[ \int_{k\delta}^t(1-e^{-2 (t-s)})(e^{-2 (t-s)}) ds \right]\cdot I_{d \times d}\\
            &= \frac{\c}{2 L} \left[1+e^{-4(t-k\delta)}-2e^{-2(t-k\delta)}\right] \cdot I_{d \times d}.
            \end{align*}           
            We thus have an explicitly defined Gaussian. Notice that we can sample from this distribution in time linear
            in $d$, since all $d$ coordinates are independent.
        \end{proof}    
        
   \end{section}

    \newpage

\end{document}